\documentclass[10pt,twocolumn,letterpaper]{article}

\usepackage{iccv}
\usepackage{times}
\usepackage{epsfig}
\usepackage{graphicx}
\usepackage{amsmath}
\usepackage{amssymb}

\usepackage{amsmath, amsthm, amssymb}
\usepackage{textcomp}
\usepackage{stmaryrd}
\usepackage{upgreek}
\usepackage{bm}
\usepackage{cases}
\usepackage{mathtools}
\usepackage{arydshln}
\usepackage{multirow}
\usepackage{bbm}
\usepackage{etoc}       

\usepackage{multirow}
\usepackage{rotating}
\usepackage{booktabs}
\usepackage{tabularx}
\usepackage{tabulary}
\usepackage[table]{xcolor}

\usepackage{multirow}
\usepackage{array}
\usepackage{booktabs}
\usepackage{wrapfig}
\usepackage{duckuments}
\usepackage{arydshln}

\usepackage{inconsolata}

\usepackage{pifont}%

\usepackage[english]{babel}
\usepackage{blindtext}

\usepackage{xcolor,colortbl}

\definecolor{Gray}{gray}{0.85}
\usepackage{subfigure}

\newlength\myheight
\newlength\mydepth
\settototalheight\myheight{Xygp}
\newcommand*\inlinegraphics[1]{%
  \settototalheight\myheight{Xygp}%
  \settodepth\mydepth{Xygp}%
  \raisebox{-\mydepth}{\includegraphics[height=\myheight]{#1}}%
}

\newcommand{\band}{\rowcolor{gray!20}}
\newcommand{\yband}{\rowcolor{yellow!20}}
\newcommand{\bband}{\rowcolor{blue!20}}
\newcommand{\gband}{\rowcolor{green!20}}
\newcommand{\rband}{\rowcolor{red!20}}

\makeatletter
\@namedef{ver@everyshi.sty}{}
\makeatother
\usepackage{hhline} %
\usepackage{highlight}

\usepackage{amsmath,amsfonts,bm}

\def\eqref#1{equation~\ref{#1}}

\def\1{\bm{1}}

\DeclareMathAlphabet{\mathsfit}{\encodingdefault}{\sfdefault}{m}{sl}
\SetMathAlphabet{\mathsfit}{bold}{\encodingdefault}{\sfdefault}{bx}{n}

\belowdisplayskip \abovedisplayskip

\newlength{\sectionReduceTop}
\newlength{\sectionReduceBot}
\newlength{\subsectionReduceTop}
\newlength{\subsectionReduceBot}
\newlength{\abstractReduceTop}
\newlength{\abstractReduceBot}
\newlength{\captionReduceTop}
\newlength{\captionReduceBot}
\newlength{\subsubsectionReduceTop}
\newlength{\subsubsectionReduceBot}

\newlength{\eqnReduceTop}
\newlength{\eqnReduceBot}

\newlength{\horSkip}
\newlength{\verSkip}

\newlength{\figureHeight}
\newcommand{\prithvi}{\textcolor{black}}

\newcommand{\pnav}{\textsc{PointNav}\xspace}
\newcommand{\onav}{\textsc{ObjectNav}\xspace}
\newcommand{\rnav}{\textsc{RobustNav}\xspace}
\newcommand{\rthor}{\textsc{RoboTHOR}\xspace}

\newcommand{\degr}{$^{\circ}$\xspace}
\newcommand{\vcr}{{\texttt{vis}\xspace}}
\newcommand{\dcr}{{\texttt{dyn}\xspace}}
\newcommand{\vdcr}{{\texttt{vis}+\texttt{dyn}\xspace}}

\newcommand{\mv}{\texttt{\textbf{move\_ahead}}\xspace}
\newcommand{\rlf}{\texttt{\textbf{rotate\_left}}\xspace}
\newcommand{\rrr}{\texttt{\textbf{rotate\_right}}\xspace}
\newcommand{\lu}{\texttt{\textbf{look\_up}}\xspace}
\newcommand{\ld}{\texttt{\textbf{look\_down}}\xspace}
\newcommand{\ed}{\texttt{\textbf{end}}\xspace}

\makeatletter
\DeclareRobustCommand\onedot{\futurelet\@let@token\@onedot}
\def\@onedot{\ifx\@let@token.\else.\null\fi\xspace}

\makeatother

\newcommand{\eat}[1]{}
\newcommand{\replace}[2]{} %

\usepackage[pagebackref=true,breaklinks=true,letterpaper=true,colorlinks,bookmarks=false]{hyperref}

\iccvfinalcopy %

\def\iccvPaperID{3176} %

\begin{document}

\title{\textsc{RobustNav}: Towards Benchmarking Robustness in Embodied Navigation}

\author{
  Prithvijit Chattopadhyay$^{1,2}$\thanks{Part of the work done when PC was a research intern at AI2.} \qquad Judy Hoffman$^1$ \qquad Roozbeh Mottaghi$^{2,3}$ \qquad Aniruddha Kembhavi$^{2,3}$\\[0.02in]
  \\
  \normalsize $^1$Georgia Tech \qquad $^2$PRIOR @ Allen Institute of AI \qquad $^3$University of Washington\\[0.05in]
{{\small\tt \{prithvijit3,judy\}@gatech.edu}} \qquad {{\small \tt \{roozbehm,anik\}@allenai.org}}\\
\small\tt \url{prior.allenai.org/projects/robustnav}
}

\maketitle
\ificcvfinal\thispagestyle{empty}\fi

\begin{abstract}
\vspace{\abstractReduceTop}

As an attempt towards assessing the
robustness of embodied navigation agents,
we propose \textbf{\rnav}, a framework %
to quantify 
the performance 
of embodied navigation agents 
when exposed
to a wide variety
of visual -- affecting RGB inputs -- and dynamics -- affecting transition dynamics
-- corruptions. 
Most recent efforts in visual navigation
have 
typically
focused on generalizing to novel target environments 
with
similar appearance and dynamics characteristics.
With \rnav, we find that 
some
standard embodied navigation agents significantly underperform (or fail)
in the presence of visual or dynamics corruptions. We 
systematically
analyze the kind of idiosyncrasies that emerge in the behavior of such agents
when operating under corruptions. 
Finally, for visual corruptions in \rnav, we show that while standard techniques to
improve robustness such as data-augmentation and self-supervised
adaptation offer some zero-shot resistance and improvements in navigation performance, there
is still a long way to go in terms of recovering lost performance relative to
clean ``non-corrupt'' settings, warranting more research in this direction.
Our code is available at
\url{https://github.com/allenai/robustnav}.

\vspace{\abstractReduceBot}
\end{abstract}

\section{Introduction}
\label{sec:intro}

\begin{figure}[ht!]
\centering
\includegraphics[width=\linewidth]{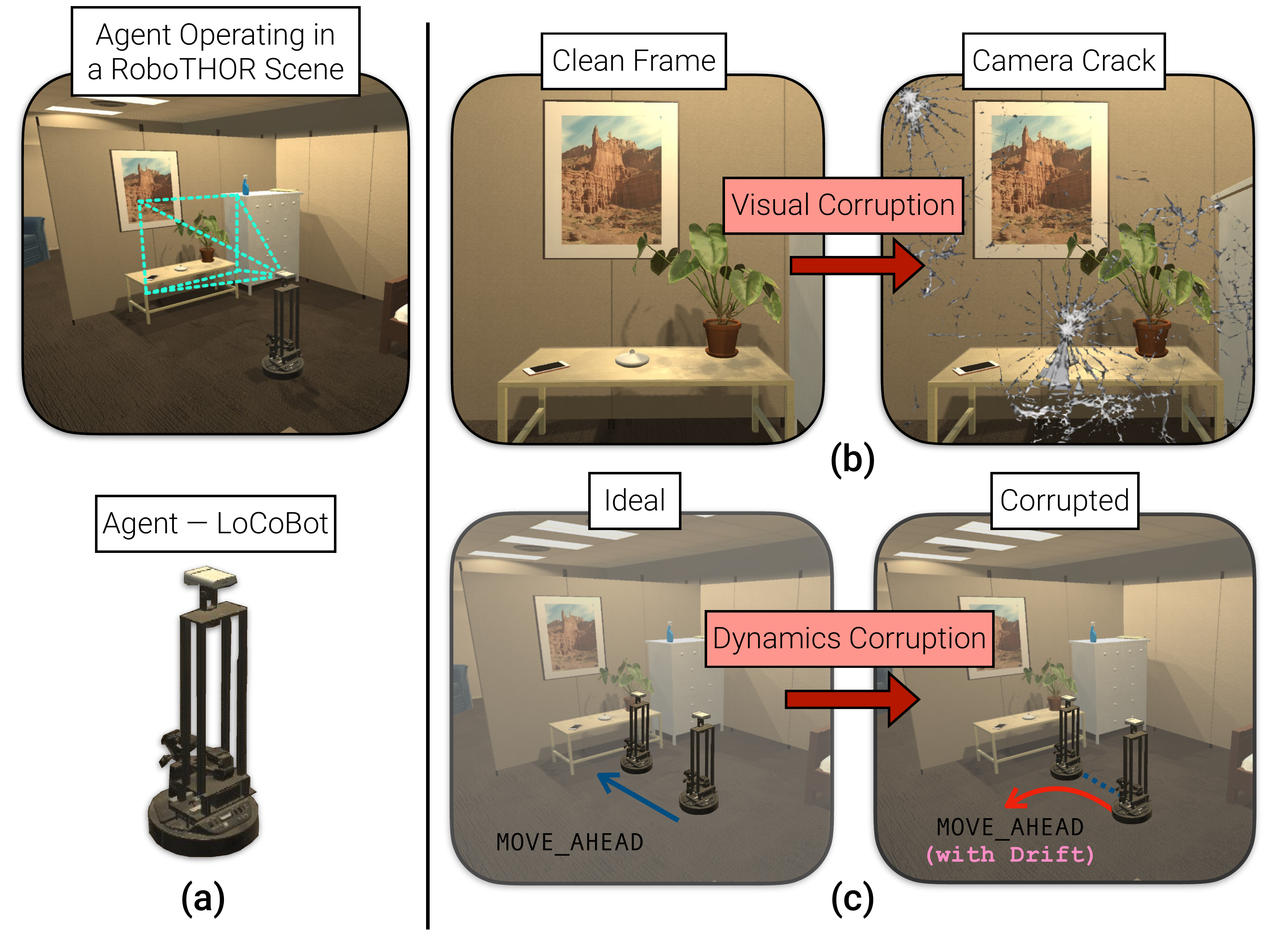}
\caption{\textbf{\rnav.} (a) A navigation agent pretrained in clean environments is
asked to navigate to targets in unseen environments in the presence of (b) \textit{visual} and
(c) \textit{dynamics} based corruptions. 
Visual corruptions (ex. camera crack) affect the agent's egocentric RGB observations while
Dynamics corruptions
(ex. drift in translation)
affect 
\prithvi{transition dynamics in the unseen environment.}
}
\label{fig:teaser_figure}
\end{figure}

A longstanding goal of the artificial intelligence community has been to develop
algorithms for embodied agents that are capable of reasoning about rich perceptual
information and thereby accomplishing tasks by navigating in and interacting
with their environments.
In addition to being able to exhibit these capabilities, it is equally important
that such embodied agents are able to do so in a robust and generalizable manner.

A major challenge in Embodied AI is to ensure that agents can
generalize to environments with different \textit{appearance statistics} and \textit{motion dynamics} than the environment used for training those agents. 
For instance, an agent that is trained to navigate in 
``sunny'' weather 
should continue to operate in rain 
despite the
drastic changes in the appearance, 
and
an agent that is trained to move on carpet 
should decidedly navigate when on
a hardwood floor 
despite
the discrepancy in friction. 
While a potential solution may be to calibrate the agent for a specific target
environment, it is not a scalable one since there can be enormous varieties of unseen environments and situations. A more robust, efficient and scalable solution is 
to equip agents with the ability to autonomously adapt to new situations by
interaction without having to train for every possible target scenario.
Despite the remarkable progress in Embodied AI, especially in embodied navigation \cite{zhu17,savinov18,savva2019habitat,wortsman2019learning,chaplot2020object}, most efforts focus on  
generalizing trained agents
to unseen environments, but critically assume similar appearance and dynamics attributes across train and test environments.

As a first step towards assessing general purpose robustness
of embodied agents, we propose \rnav, a framework to quantify the performance 
of embodied navigation agents when exposed to a wide variety of common visual (\vcr)  and dynamics (\dcr) corruptions -- artifacts that affect the egocentric RGB observations and transition dynamics, respectively. We envision \rnav as a testbed for adapting agent behavior across different perception and actuation properties. While assessing robustness to changes (stochastic or otherwise) in environments has been 
investigated in the robotics community~\cite{hofer2020perspectives,del2015addressing,del2016robustness,giftsun2017robustness}, the simulated nature of \rnav enables practitioners to explore robustness against a rich and very diverse set of changes, while inheriting the advantages of working in simulation -- speed, safety, low cost and reproducibility.

\rnav consists of two widely studied embodied navigation tasks, Point-Goal Navigation (\textsc{PointNav})~\cite{anderson2018evaluation} and Object-Goal Navigation (\textsc{ObjectNav})~\cite{batra2020objectnav} -- the tasks of navigating to a goal-coordinate in a global reference frame or an instance of a specified object, respectively. Following the standard protocol, agents learn using a set of training scenes and are evaluated within a set of held out test scenes, but differently, \rnav test scenes are subject to a variety of realistic \textit{visual} and \textit{dynamics} corruptions. 
These corruptions can emulate real world scenarios such as a malfunctioning camera or 
drift
(see Fig.\ref{fig:teaser_figure}).

As zero shot adaptation to test time corruptions may be out of reach for our current algorithms, we provide agents with a fixed ``calibration budget" (number of interactions) within the target world for unsupervised adaptation. This mimics a real-world analog where a shipped robot is allowed to adapt to changes in the environment by executing a reasonable number of unsupervised interactions. Post calibration, agents are evaluated on the two tasks in the corrupted test environments using standard navigation metrics.

Our extensive analysis reveals that both \pnav and \onav agents experience significant degradation in performance across the range of corruptions, particularly when multiple corruptions are applied together. We show that this degradation reduces in the presence of a clean depth sensor suggesting the advantages of incorporating multiple sensing modalities, to improve robustness. 
We find that data augmentation and self-supervised adaptation
strategies offer some zero-shot resistance and improvement over degraded performance,
but are unable to fully recover this gap in performance.
Interestingly, we also note that visual corruptions affect embodied tasks differently from static tasks like object recognition -- suggesting that visual robustness should be explored within an embodied task. %
Finally, we analyze several interesting behaviors
our agents 
exhibit in the presence of corruptions --
such as increase in the number of collisions and inability to terminate episodes successfully.

In summary, our contributions include: 
(1) We present \rnav -- a framework for benchmarking and assessing the robustness of embodied navigation agents to visual and dynamics corruptions. 
(2) Our findings show that present day navigation agents trained in simulation underperform severely when evaluated in corrupt target environments. 
(3) We systematically analyze the kinds of mistakes embodied navigation agents make when operating under such corruptions.
(4) We find that although standard data-augmentation techniques and self-supervised adaptation
strategies offer some improvement, much remains to be done
in terms of fully recovering lost performance.

\rnav provides a fast framework to develop and test robust embodied policies, before they can be deployed onto real robots. While \rnav\ currently supports navigation heavy tasks, the supported corruptions can be easily extended to more tasks, as they get popular within the Embodied AI community.

\section{Related Work}
\label{sec:rel_work}
\par \noindent
\textbf{Visual Navigation.} Tasks involving navigation based on egocentric visual inputs have
witnessed exciting progress in recent years~\cite{savva2019habitat,embodiedqa,gordon2018iqa,chen2019audio,gan2020look,chen2020learning}. Some of the widely studied tasks
in this space include \pnav~\cite{anderson2018evaluation}, \onav~\cite{batra2020objectnav} and 
goal-driven navigation where the target is specified by a goal-image~\cite{zhu17}.
Approaches to solve \pnav and \onav can broadly be classified into two categories -- (1) learning
neural policies end-to-end using RL~\cite{wijmans2019decentralized,ye2020auxiliary,savinov18,savva2019habitat,wortsman2019learning} or (2) decomposing navigation into a mapping (building a semantic map) and path planning stage~\cite{chaplot2019learning,chaplot2020object,gupta2017cognitive,ramakrishnan2020occupancy}. 
Recent research has also focused on assessing the ability of polices
trained in simulation to transfer to real-world robots operating in physical spaces~\cite{kadian2019we,robothor}.

\par \noindent
\textbf{Robustness Benchmarks.} Assessing robustness of deep neural models has received quite a bit of attention in recent years~\cite{hendrycks2019benchmarking,recht2019imagenet,hendrycks2020pretrained,andriushchenko2020understanding}. Most relevant and closest to our work is~\cite{hendrycks2019benchmarking},
where authors show that computer vision models are susceptible to
several synthetic visual corruptions, as measured in the proposed ImageNet-C benchmark.
In~\cite{kamann2020benchmarking,michaelis2019benchmarking}, authors
study the effect of similar visual corruptions for semantic segmentation
and object detection on standard static benchmarks.
\rnav integrates several 
visual corruptions from~\cite{hendrycks2019benchmarking}
and adds ones such as low-lighting and crack in the camera-lens,
but within an embodied scenario. Our findings (see Sec.~\ref{sec:findings}) show that visual corruptions affect embodied tasks differently from static tasks like object recognition.
In~\cite{taori2020measuring},
authors repurpose the ImageNet validation split to be used as a benchmark for assessing
robustness to natural distribution shifts (unlike the ones introduced in~\cite{hendrycks2019benchmarking})
and ~\cite{engstrom2020identifying} identifies statistical biases in the
same.
Recently,~\cite{hendrycks2020many}
proposes three extensive benchmarks assessing robustness to image-style, 
geographical location and camera operation.

\par \noindent
\textbf{Real-world RL Suite.} Efforts similar to \rnav have been made in~\cite{dulacarnold2020realworldrlempirical},
where authors
formalize $9$ different challenges holding back RL from real-world use -- including
actuator delays, 
high-dimensional state 
and 
action spaces, latency, and others.
In contrast, \rnav focuses on challenges in the visually rich domains and complexities associated with visual observation. Recently, Habitat~\cite{savva2019habitat} also introduced actuation (from~\cite{murali2019pyrobot}) and visual noise models for navigation tasks. In contrast, \rnav is designed to benchmark robustness of models against a variety of visual and dynamics corruptions ($7$ \vcr\ 
and $4$ \dcr\ corruptions for both \pnav and \onav). %

\par \noindent
\textbf{Adapting Visio-Motor Policies.} Significant progress has been made in the
problem of adapting policies trained with RL from a source to a target environment.
Unlike \rnav, major assumptions involved in such transfer settings are either access to task-supervision
in the target environment~\cite{gordon2019splitnet} or access to paired data from the source and target
environments~\cite{golemo2018sim,truong2021bi}. 
Domain
Randomization (DR)~\cite{akkaya2019solving,raparthy2020generating,lee2019network,muratore2020bayesian} is another common approach to train policies robust to various
environmental factors. Notably,~\cite{lee2019network} perturbs features early in the
visual encoders of the policy network so as to mimic 
DR
and~\cite{muratore2020bayesian} selects optimal 
DR
parameters during training based on sparse data obtained from the the real world.
In absence of task supervision,
another common approach is to optimize self-supervised objectives in the target~\cite{wortsman2019learning,savinov2018episodic}
and has been used to
adapt policies to visual disparities (see Sec.~\ref{sec:findings}) in new environments~\cite{hansen2020self}. To adapt
to changes in transition dynamics, a common approach is to train on a broad family
of dynamics models
and perform system-identification (ex. with domain classifiers~\cite{eysenbach2020off})
in the target environment~\cite{yang2019single,zhou2019environment}.
~\cite{kadian2019we,robothor} studies the extent to which
embodied navigation agents transfer from simulated environments to real-world physical
spaces.
Among these, we investigate two of the most popular approaches -- self-supervised adaptation~\cite{hansen2020self} and aggressive data augmentation and measure if they can help build resistance to \vcr\ corruptions.

\section{\textsc{RobustNav}}
\label{sec:rnav}
We present \rnav, a benchmark 
to assess the robustness of embodied agents to common visual (\vcr) and dynamics (\dcr) corruptions. 
\rnav is built on top of \rthor~\cite{deitke2020robothor}.
In this work, we study the effects 
corruptions 
have on
two kinds of embodied navigation agents -- namely, \pnav (navigate to a specified goal coordinate) and \onav (navigate to an instance of an object category).
While we restrict our experiments to 
navigation, in practice, our
\vcr\ and \dcr\
corruptions can also be extended to 
other embodied tasks that share the same modalities, for instance tasks involving
interacting with objects.

\begin{figure}[t!]
\centering
\includegraphics[width=\linewidth]{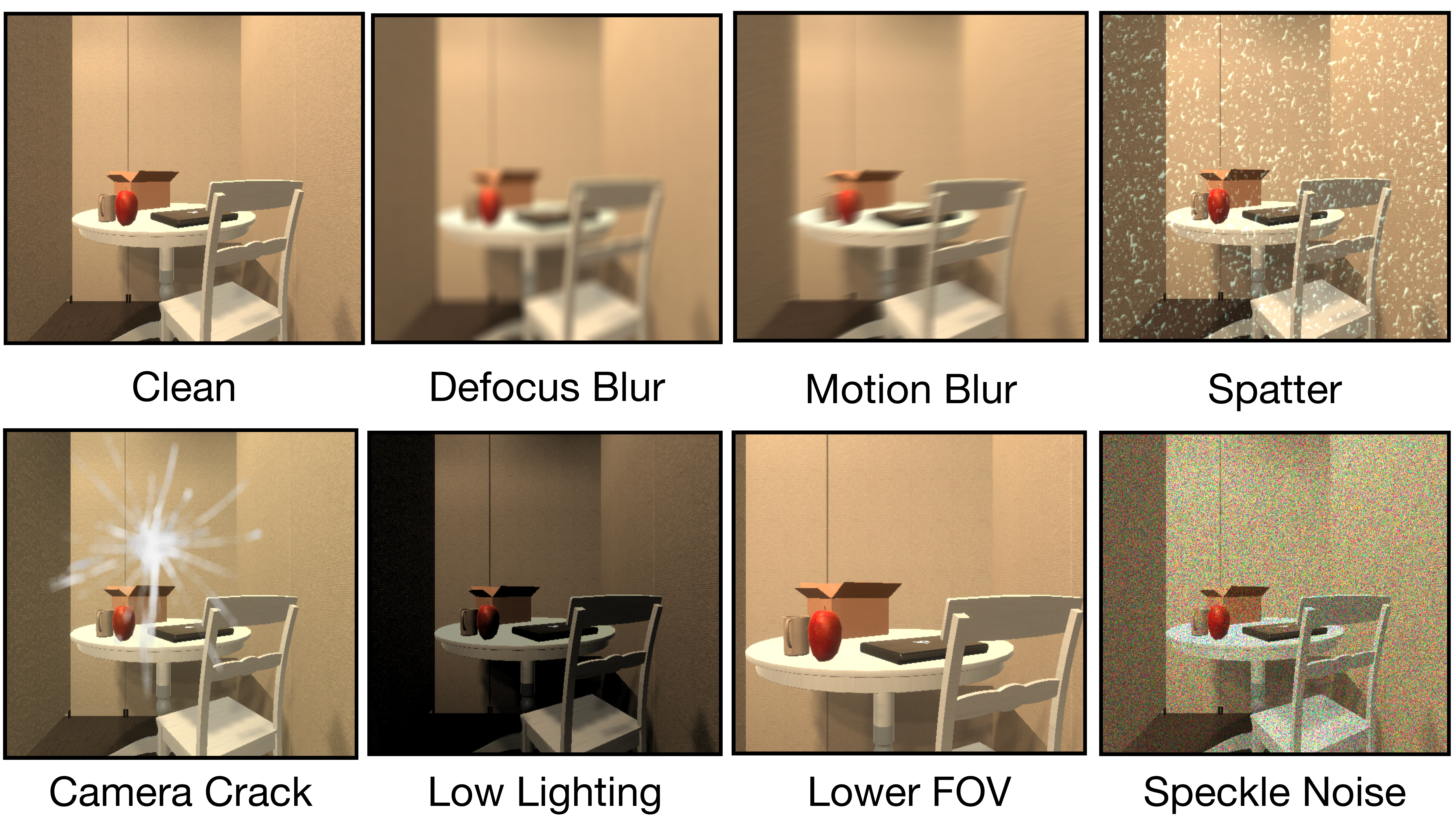}
\caption{\textbf{Visual Corruptions.} 
Visual corruptions \rnav supports in the unseen target
environments. Top-left shows a clean RGB frame and rest show corrupted versions of the same.
}
\label{fig:vis_corr_samples}
\end{figure}

In \rnav, agents are trained 
within 
the training scenes and evaluated on ``corrupt" unseen target scenes. 
Corruptions 
in target scenes 
are drawn from a set of predefined \vcr\ and \dcr\ corruptions. 
As is the case with any form of modeling of corruptions (or noise) in simulation~\cite{hofer2020perspectives,deitke2020robothor}, there will always be an approximation error when 
the \vcr\ and \dcr\ 
corruptions are compared to their real world counterparts.
Our 
\prithvi{aim}
is to ensure that 
the
\rnav 
benchmark acts as a stepping stone towards the larger goal of 
obtaining 
robust agents, ready to be deployed in real world.

To adapt to a corrupt target scene, we provide agents with a ``calibration budget'' -- an upper bound on the number of interactions an agent is allowed to have with the target environment without any external task supervision. This is done to mimic a real-world analog where a shipped robot
is allowed to adapt to changes in the environment by executing a reasonable number of unsupervised interactions. We adopt a modest definition of the calibration-budget based on the number of steps it takes an agent to reasonably recover degraded performance in the most severely corrupted environments when finetuned under complete supervision (see Table.~\ref{tab:vis_corr_adapt}) -- set to $\sim166$k steps for all our experiments.
We attempt to understand if self-supervised adaptation approaches~\cite{hansen2020self} improve performance when
allowed to adapt under this calibration budget (see Sec.~\ref{sec:findings}, resisting corruptions).
We now describe in detail the
\vcr\ and \dcr\
corruptions present in \rnav.

\par \noindent
\textbf{Visual Corruptions.} Visual corruptions are artifacts that degrade the navigation agent's egocentric RGB observation (see Fig.~\ref{fig:vis_corr_samples}). We provide seven visual corruptions within \rnav, four of which are drawn from the set of corruptions and perturbations proposed in~\cite{hendrycks2019benchmarking}
-- 
Spatter, Motion Blur, Defocus Blur and Speckle Noise; 
realistic corruptions that one might expect to see on a real robot.
Spatter emulates occlusion in images due to particles of dirt, water droplets, etc. residing on the camera lens. Motion Blur emulates blur in images 
due to jittery movement of the robot. Defocus Blur occurs when the 
RGB image is out of focus. Speckle Noise emulates granular interference that inherently exists in and degrades the quality of images obtained by the camera (modeled as additive noise with the noise being proportional to the original pixel intensity). Each of these corruptions can manifest at five levels of
severity indicating increase in the extent of visual degradation ($1\rightarrow5$). 

In addition to these, we also add low-lighting (low-lighting conditions in the target environment, has associated severity levels $1\rightarrow5$), lower-FOV (agents operating with a lower camera field of view compared to the one used during training, $79^{\circ}\rightarrow39.5^{\circ}$) and camera-crack (a randomized crack in the camera lens).
For camera-crack, we use fixed random seeds for the $15$ validation scenes which dictate the location and kind of crack on the camera lens.

\begin{figure}[t!]
\centering
\includegraphics[width=\linewidth]{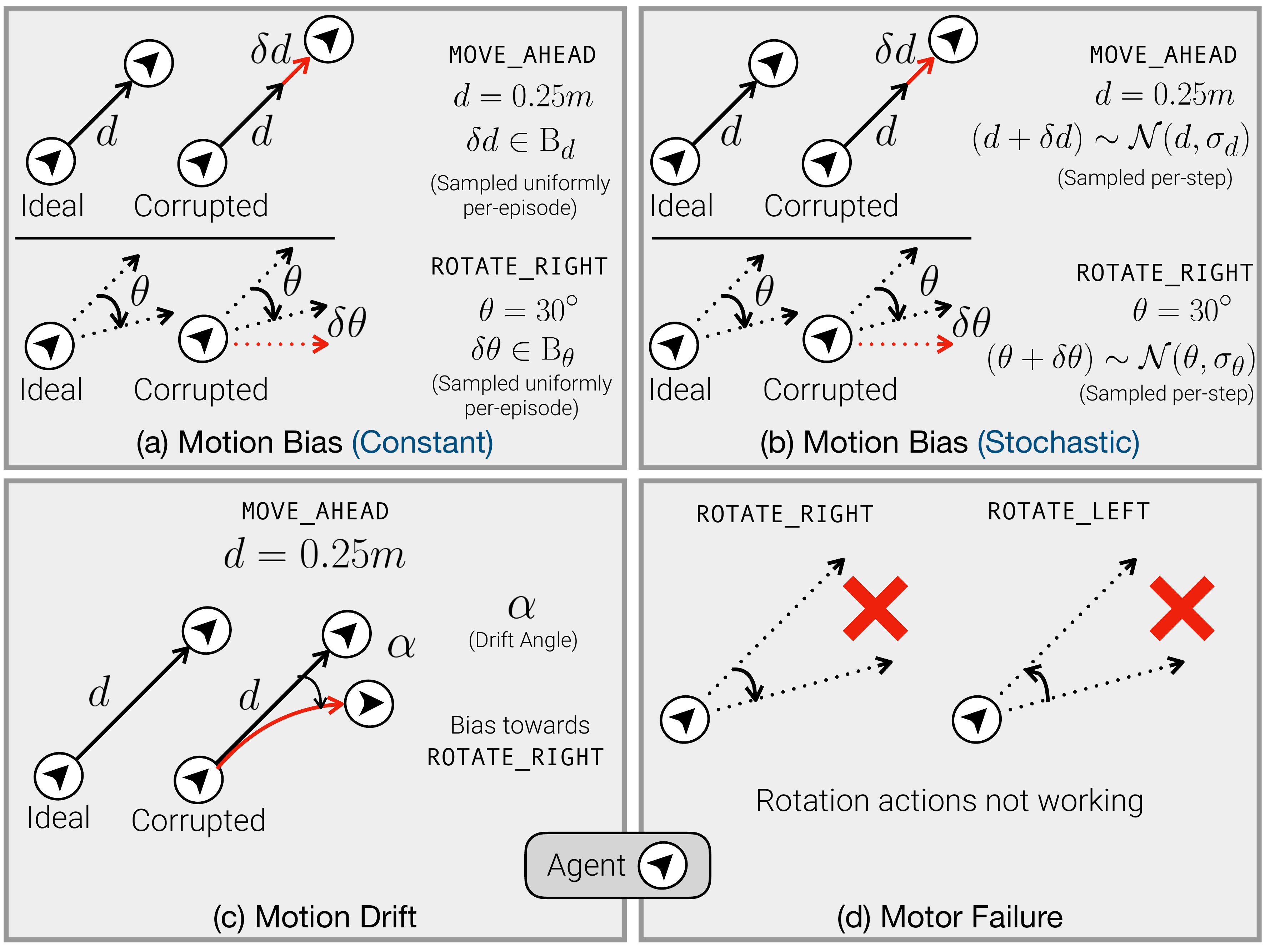}
\caption{\textbf{Dynamics Corruptions.} 
We show the kinds of dynamics corruptions supported in
\rnav. Motion Bias (C \& S) are modeled to mimic friction.
Motion Drift models a setting where translation actions
have a slight bias towards rotating right (or left).
In Motor Failure, the one of the rotation actions fail.
}
\label{fig:dyn_corr_samples}
\end{figure}

\par \noindent
\textbf{Dynamics Corruptions.} Dynamics corruptions affect the transition dynamics of the agents in the target environment (see Fig.~\ref{fig:dyn_corr_samples}). We consider
three classes of dynamics corruptions
-- Motion Bias, Motion Drift and Motor Failure.
Our \dcr\ corruptions are motivated from and
in line with the well-known systematic and/or stochastic drifts (due to error accumulation) and biases in robot motion~\cite{laddmotion,bekrisefficient,garcia2015self,6094632}.

A common dynamics corruption observed in the real world is friction. Unfortunately \rthor 
does not yet natively support multiple friction zones within a scene, as may be commonly observed in a real physical environment (for instance the kitchen floor in a house may have smooth tiles while the bedroom may have rough hardwood floors). In lieu of this, we present the Motion Bias corruption. In the absence of this corruption, the \mv action moves an agent forward by $0.25$m, and rotation \rlf and \rrr actions rotate an agent by $30^{\circ}$ left and right respectively. 
Motion Bias can induce either 
(a) a \textit{constant} bias 
drawn uniformly per-episode from $B_d=\{\pm0.05,\pm0.1,\pm0.15\}$m or $B_{\theta}=\{\pm5^{\circ}, \pm10^{\circ}, \pm15^{\circ}\}$ 
or 
(b) \textit{stochastic} translation and rotation amounts drawn
per-step from $\mathcal{N}(0.25\text{m}, 0.1\text{m})$ and $\mathcal{N}(30.0^{\circ}, 10^{\circ})$ respectively.\footnote{(a) Motion Bias (C) is intended to model scene-level friction,
different floor material in the target environment; (b) Motion Bias (S) is intended
to model high and low friction zones in a scene. Including more
sophisticated models of friction is in the feature roadmap
for \rnav.}

Motion Drift models a setting where an agent's translation movements in the environment include
a slight bias towards turning left or right. Specifically, the \mv action, instead of moving an agent forward $0.25$m in the direction of its heading
(intended behavior), drifts towards the left or right directions stochastically (for an episode) by $\alpha=10^{\circ}$ and takes it to a location 
which deviates in a direction perpendicular to the original heading
by 
a max of $\sim0.043$m.
Motor-failure is the setting where either the \rlf or the \rrr actions malfunction throughout an evaluation episode. 

With the exception of Motion-Bias (S) -- the \textit{stochastic} version -- the agent also operates under standard actuation noise models as calibrated from a LoCoBot in~\cite{robothor}. 
Recently, PyRobot~\cite{murali2019pyrobot} has also introduced LoCoBot
calibrated noise models that demonstrate strafing and drifting. While we
primarily rely on the noise models calibrated in~\cite{deitke2020robothor}, for
completeness, we also include results with the PyRobot noise models.

\par \noindent 
\textbf{Tasks.} \rnav consists of two major embodied navigation tasks -- namely, \pnav and \onav.
In \pnav, an agent is initialized at a random spawn location and orientation in an environment and is asked to navigate to target coordinates specified relative to the agent's position. The agent must navigate based \textbf{only} on sensory inputs from an RGB (or RGB-D) and a GPS + Compass sensor. An episode is declared successful if the agent stops within $0.2$m of the goal location
(by intentionally invoking an \ed action).
In \onav, an agent is instead asked to navigate to an instance of a specified object category (\textit{e.g., Television}, $1$ out of total $12$ object categories) given \textbf{only} ego-centric sensory inputs -- RGB or RGB-D.
An episode is declared successful if the agent stops within $1.0$m of the target object (by
invoking an \ed action) and has the target object in it's egocentric view.
Due to the lack of perfect localization (no GPS + Compass sensor) and the implicit need to ground the specified object within its view, \onav may be considered a harder task compared to \pnav
-- also evident in lower \onav 
performance
(Table.~\ref{tab:pnav_onav_base_results}).

\par \noindent
\textbf{Metrics.} We report performance in terms of the following well established
navigation metrics reported in past works -- Success Rate~\texttt{(SR)} and
Success Weighted by Path Length~\texttt{(SPL)}~\cite{anderson2018evaluation}.
\texttt{SR} indicates the fraction of successful episodes. 
\texttt{SPL} provides a score for the agent's path based on how close it's length is
to the shortest path from the spawn location to the target.
If $\mathbb{I}_{\text{success}}$ denotes whether
an episode is successful (binary indicator), $l$ is the shortest path
length, 
and $p$ is the agent's path length
then
\texttt{SPL}$=\mathbb{I}_{\text{success}}\frac{l}{max(l,p)}$

\par \noindent
\textbf{Scenes.} \rnav is built on top of the \rthor scenes~\cite{robothor}. \rthor consists of
$60$ training and $15$ validation environments based on indoor apartment scenes drawn from
different layouts. To assess robustness in the presence of corruptions, we evaluate
on $1100$ (and $1095$) episodes of varying difficulties (easy, medium and hard)\footnote{
Based on shortest
path lengths -- 
(1) \pnav: $0.00-2.28$ for easy, $2.29-4.39$ for medium, $4.40-9.61$
for hard; (2) \onav: $0.00-1.50$ for easy, $1.51-3.78$ for medium
, $3.79-9.00$ for hard.
}
for \pnav (and \onav) 
across
the $15$ 
val scenes.

\par \noindent
\textbf{Benchmarking.} Present day embodied navigation agents are typically trained without any corruptions. However, we anticipate that researchers may incorporate corruptions as augmentations at training time to improve the robustness of their algorithms in order to make progress on our \rnav\ framework. For the purposes of fair benchmarking, we recommend that future comparisons using \rnav\ do not draw from the set of corruptions reserved for the target scenes -- ensuring the corruptions encountered in the target scenes are indeed ``unseen''.

\section{Experimental Setup}
\label{sec:experiments}
\begin{table}[t]
\footnotesize
\centering
\setlength{\tabcolsep}{2.5pt}
\begin{center}
\resizebox{0.7\columnwidth}{!}{
\begin{tabulary}{0.7\columnwidth}{CLCCC}
\toprule
& \textbf{Corruptions} & Top-1 Acc. $\uparrow$ & Top-5 Acc. $\uparrow$\\
\midrule
\multirow{11}{*} 
\small\texttt{1} &Clean & 69.76 & 89.08\\
\midrule
\multirow{11}{*}
\small\texttt{2} &Camera Crack$^\dagger$ & 57.71\tiny{$\pm$5.82} & 80.27\tiny{$\pm$4.54}\\
\small\texttt{3} &Lower FOV$^*$ & 45.44 & 69.53\\
\small\texttt{4} & Low Lighting & 35.76 & 58.54\\
\small\texttt{5} &Spatter & 19.73 & 39.34\\
\small\texttt{6} &Motion Blur & 10.11 & 22.66\\
\small\texttt{7} &Defocus Blur & 9.39 & 22.25\\
\small\texttt{8} &Speckle Noise & 7.79 & 18.84\\
\bottomrule
\end{tabulary}}
\vspace{2pt}
\caption{\textbf{ImageNet Performance Degradation.} Degradation in classification
performance on the ImageNet validation split under visual corruptions for
ResNet-18~\cite{he2016deep}
trained on ImageNet (used as the agent's visual encoder). Corruptions in 2-8 are present \rnav. 
$^*$Since mimicking lower FOV requires access to camera intrinsics, 
unavailable
for static datasets, we mimic the same by aggressive center-cropping.
$^\dagger$For camera-crack, we report performance over all possible crack settings
present in \rnav.
}
\label{tab:imgnet_vis_corr_perf}
\end{center}
\end{table}
\par \noindent
\textbf{Agent.} Our \pnav agents have 4 actions available to them -- namely, \mv ($0.25$m), \rlf ($30$\degr), \rrr ($30$\degr) and \ed. The action \ed indicates that the agent believes that it has
reached the goal, thereby terminating the episode. During evaluation,
we allow an agent to execute a maximum of 300 steps -- if an agent does not call \ed within
300 steps, we forcefully terminate the episode.
For \onav, in addition to the
aforementioned actions, the agent also has the ability to \lu \prithvi{or} \ld\ -- indicating
change in the agent's view $30$\degr above or below the forward camera horizon. The
agent receives $224\times224$ sized ego-centric observations (RGB or RGB-D).
All agents are trained under LoCoBot calibrated actuation noise models from~\cite{robothor} -- 
$\mathcal{N}(0.25\text{m}, 0.005\text{m})$ for translation and 
$\mathcal{N}(30^{\circ}, 0.5^{\circ})$ for
rotation.
Our agent architectures (akin to~\cite{wijmans2019decentralized}) are composed of a CNN head to process
input observations followed by a recurrent (GRU) policy network (more details in 
Sec.~\ref{sec:agent} of appendix).

\par \noindent
\textbf{Training.} We train our agents using DD-PPO~\cite{wijmans2019decentralized} -- a decentralized, distributed and synchronous version of the Proximal Policy Optimization (PPO)~\cite{schulman2017proximal} algorithm. 
If $\mathtt{R} = 10.0$ denotes the terminal reward obtained at the end of a successful episode 
(with $\mathbb{I_{\text{success}}}$ being an indicator variable denoting whether
an episode was successful),
$\Delta^{\text{Geo}}_t$ denotes the change in geodesic distance to 
target at timestep $t$ from $t-1$
and $\lambda = -0.01$ denotes a slack penalty to encourage efficiency,
then the reward received by the agent at time-step $t$ can be expressed as,

$r_t = \underbrace{\mathtt{R}\,.\,\mathbb{I_{\text{success}}}}_{\text{success reward}} - \underbrace{\Delta^{\text{Geo}}_t}_{\text{reward shaping}} + \underbrace{\lambda}_{\text{slack reward}}$

We train our agents
using the AllenAct~\cite{AllenAct} framework.

\section{Results and Findings}
\label{sec:findings}
\begin{table*}
    \setlength{\tabcolsep}{7pt}
    \centering
\resizebox{0.9\textwidth}{!}{
\begin{tabular}{l c c c cc c cc c cc c cc c}
\toprule
    & & & && \multicolumn{5}{c}{\textbf{\pnav}} && \multicolumn{5}{c}{\textbf{\onav}} \\
    & & & && \multicolumn{2}{c}{\texttt{RGB}} && \multicolumn{2}{c}{\texttt{RGB-D}}  &&
             \multicolumn{2}{c}{\texttt{RGB}} && \multicolumn{2}{c}{\texttt{RGB-D}} \\
    \cmidrule{6-7} \cmidrule{9-10}
    \cmidrule{12-13} \cmidrule{15-16}
    \texttt{\#} & \textbf{Corruption}~$\downarrow$ &  V & D &&
        \textbf{\texttt{SR}}~$\uparrow$ &  \textbf{\texttt{SPL}}~$\uparrow$ &&   \textbf{\texttt{SR}}~$\uparrow$ &  \textbf{\texttt{SPL}}~$\uparrow$ &&
        \textbf{\texttt{SR}}~$\uparrow$ &  \textbf{\texttt{SPL}}~$\uparrow$ &&   \textbf{\texttt{SR}}~$\uparrow$ &  \textbf{\texttt{SPL}}~$\uparrow$ \\

   \midrule
   \band \texttt{1} & Clean &  &  && 98.82 &  83.13 && 98.54 &  84.60 && 31.05 &  14.26 && 35.62 &  17.20\\
   \midrule
    \texttt{2} & Low Lighting &  \checkmark &  && 94.36 &  75.15 && 99.45 &  84.97 && 10.78 &   4.59 && 21.64 &   9.98\\
    \texttt{3} & Motion Blur &  \checkmark &  && 95.72 &  73.37 && 99.36 &  85.36 && 10.59 &   4.03 && 20.27 &   8.29\\
    \texttt{4} & Camera Crack &  \checkmark &  && 82.07 &  63.83 && 95.72 &  81.21 && 7.21 &   3.57 && 24.29 &  12.50\\
    \texttt{5} & Defocus Blur &  \checkmark &  && 75.89 &  53.55 && 99.09 &  85.54 && 5.02 &   2.42 && 19.18 &   7.90\\
    \texttt{6} & Speckle Noise &  \checkmark &  && 67.42 &  48.57 && 98.73 &  84.66 && 9.04 &   3.66 && 18.63 &   7.52\\
    \texttt{7} & Lower-FOV &  \checkmark &  && 42.49 &  31.73 && 89.08 &  73.59 && 9.77 &   3.90 && 9.86 &   4.77\\
    \texttt{8} & Spatter &  \checkmark &  && 33.58 &  24.72 && 98.91 &  84.81 && 6.76 &   2.93 && 21.10 &   9.06\\

   \midrule 
    \texttt{9} & Motion Bias (C) &   & \checkmark && 92.81 &  77.83 && 93.36 &  79.46 && 31.51 &  14.09 && 31.96 &  15.38\\
    \texttt{10} & Motion Bias (S) &   & \checkmark && 94.72 &  76.95 && 96.72 &  79.08 && 30.87 &  14.15 && 35.62 &  16.39\\
    \texttt{11} & Motion Drift &   & \checkmark && 95.72 &  76.19 && 93.36 &  75.08 && 29.68 &  13.58 && 34.06 &  17.03\\
    \texttt{12} & PyRobot~\cite{murali2019pyrobot} (ILQR) Mul. = 1.0 &   & \checkmark && 96.00 &  67.79 && 95.45 &  69.27 && 32.51 &  11.26 && 36.35 &  13.62\\
    \texttt{13} & Motor Failure &   & \checkmark && 20.56 &  17.63 && 20.56 &  17.62 && 4.20 &   2.43 && 6.39 &   3.67\\
    
   \midrule 
    \texttt{14} & Defocus Blur + Motion Bias (S) &  \checkmark & \checkmark && 76.52 &  51.08 && 97.18 &  79.46 && 5.57 &   2.00 && 18.54 &   7.23\\
    \texttt{15} & Speckle Noise + Motion Bias (S) &  \checkmark & \checkmark && 62.69 &  43.31 && 95.81 &  78.27 && 7.85 &   3.73 && 18.54 &   8.16\\
    \texttt{16} & Spatter + Motion Bias (S) &  \checkmark & \checkmark && 33.30 &  23.33 && 95.81 &  78.85 && 7.85 &   3.09 && 21.28 &   9.26\\
    \midrule
    \texttt{17} & Defocus Blur + Motion Drift &  \checkmark & \checkmark && 74.25 &  50.99 && 95.54 &  76.66 && 4.57 &   1.93 && 17.35 &   6.97\\
    \texttt{18} & Speckle Noise + Motion Drift &  \checkmark & \checkmark && 64.42 &  44.73 && 94.36 &  75.23 && 8.49 &   3.67 && 19.82 &   8.61\\
    \texttt{19} & Spatter + Motion Drift &  \checkmark & \checkmark && 32.94 &  23.44 && 95.45 &  76.61 && 6.85 &   2.68 && 19.54 &   8.86\\
    \bottomrule 
\end{tabular}
}
\caption{\textbf{\textsc{PointNav} and \textsc{ObjectNav} Performance.} Degradation in task
  performance of pretrained \pnav (trained for $\sim75$M frames) and \onav (trained for $\sim300$M frames) agents when evaluated
  under \vcr\ and \dcr\ corruptions present in \rnav. 
  \pnav agents have additional access to a GPS-Compass sensor.
  For visual corruptions with controllable
  severity levels, we report results with severity set to $5$ (worst). Performance
  is measured across tasks of varying difficulties (easy, medium and hard).
  Rows are sorted based on SPL values for RGB \pnav agents. Success and SPL values
  are reported as percentages.
   (V = Visual, D = Dynamics)
}
	\label{tab:pnav_onav_base_results}
\end{table*}

In this section, we show that the
performance of \pnav and \onav agents degrades in the presence of
corruptions (see Table.~\ref{tab:pnav_onav_base_results}).
We first highlight how \vcr\ corruptions affect
static vision and embodied navigation tasks
differently
(see Table~\ref{tab:imgnet_vis_corr_perf}). 
Following this, we analyze behaviors that emerge in these agents when operating in the presence of \vcr, \dcr, and \vcr+\dcr\
corruptions. Finally, we investigate whether 
standard data-augmentation and self-supervised adaptation~\cite{hansen2020self} techniques help recover the degraded performance (see Table~\ref{tab:vis_corr_adapt}).

\vspace{\subsectionReduceTop}
\subsection{Degradation in Performance}
\label{sec:perf_degradation}
\vspace{\subsectionReduceBot}
We now present our findings regarding degradation in performance relative
to agents being evaluated in clean (no corruption) target environments (row 1 in 
Table.~\ref{tab:pnav_onav_base_results}).

\par \noindent
\textbf{Visual corruptions affect static and embodied tasks
differently.} In Table~\ref{tab:imgnet_vis_corr_perf}, we report 
object recognition
performance for models trained on the ImageNet~\cite{deng2009imagenet}
train split and evaluated on the corrupt validation splits.
In Table~\ref{tab:pnav_onav_base_results},
we report performance
degradation of \pnav and \onav agents under corruptions (row 1, clean \& rows 2-8 corrupt). 
It is important to note that the nature of tasks (one-shot prediction
vs sequential decision making)
are 
different enough that the difficulty of corruptions for classification
may not indicate the difficulty of corruptions for navigation.
We verify this hypothesis by comparing results in Tables~\ref{tab:imgnet_vis_corr_perf} and \ref{tab:pnav_onav_base_results}
-- for instance, corruptions which are severe for classification
(Defocus Blur and Speckle Noise)
are not as severe for \pnav-\texttt{RGB} 
agents in terms of relative drop from clean performance.
Additionally, 
for Mask-RCNN~\cite{he2017mask} trained on AI2-THOR images,
we note that detection (segmentation)\footnote{For the 12 \onav target classes} mAP$^{0.5:0.95}$ drops from $62.93$ ($66.29$) to $7.96$ ($8.64$) and
$6.56$ ($6.68$) for Spatter (S5) and Low-Lighting (S5), respectively -- unlike
rows 2 \& 8 in Table~\ref{tab:pnav_onav_base_results}, where Spatter appears to be much severe compared to Low-Lighting.
This difference in relative degradation suggests that
that techniques for visual adaptation or robustness
in static settings may not transfer out-of-the-box to embodied tasks, warranting more research in this direction.

\par \noindent
\textbf{Not all corruptions are equally bad.}
While we note that \pnav and \onav agents suffer a drop in
performance from clean settings, not all corruptions
are equally severe. For instance, in \pnav-\texttt{RGB}, while Low Lighting, Motion Blur and 
Motion Bias (C) 
(rows 2, 3, 9 in Table~\ref{tab:pnav_onav_base_results})
lead to a worst-case absolute drop of $<10\%$ in SPL (and $<10\%$ in SR), corruptions
like Spatter and Motor Failure (rows 8, 13) are more extreme and significantly affect task performance (absolute drops of $>57\%$ in SPL, $>65\%$ in SR). For \onav,
however, the drop in performance is more gradual across corruptions
(partly because 
it's 
a harder task
and even clean
performance is fairly low).

\par \noindent
\textbf{A ``clean'' depth sensor helps resisting degradation.} 
We
compare the \texttt{RGB} and \texttt{RGB-D} variants of the trained \pnav and \onav agents
(RGB corrupt, Depth clean) in Table~\ref{tab:pnav_onav_base_results} (corresponding \texttt{RGB} \& \texttt{RGB-D} columns). 
We observe that including
a ``clean'' depth sensor consistently improves resistance to
\vcr, \dcr\ and \vdcr\ corruptions for both \pnav and \onav.
For \pnav, we note that while \texttt{RGB} and \texttt{RGB-D}
variants have comparable clean performance (row 1), 
under severe
corruptions (Spatter, Lower-FOV and Speckle-Noise),
the \texttt{RGB-D} counterparts are ahead roughly by 
an absolute margin of
$36.09-60.09\%$ SPL.
We further observe that, barring exceptions, \pnav~\texttt{RGB-D} agents are 
generally affected minimally by corruptions -- for instance, Low-Lighting
and Motion Blur barely result in any drop in performance.
We hypothesize that this is likely because \texttt{RGB-D} navigation agents are much less reliant on the RGB sensor
compared to the \texttt{RGB} counterparts.
In \onav, an additional depth
sensor generally improves clean performance (row 1 in Table~\ref{tab:pnav_onav_base_results}) which is likely the major
contributing factor for increased resistance to corruptions.
Sensors of different modalities are likely to degrade in different scenarios -- e.g., a depth sensor may continue to 
\prithvi{perceive details}
in low lighting settings. The obtained results suggest that adding multiple sensors, while expensive can help train robust models. 
Additional sensors can also be helpful
for unsupervised adaptation during the calibration
phase. For instance, in the presence of a ``clean''
depth sensor, one can consider comparing
depth 
\prithvi{based}
egomotion estimates
with expected odometry readings in the target
environment to infer changes in 
dynamics.

\newcommand{\analysisWidth}{.21\linewidth}
\begin{figure*}[ht!]
\centering
\begin{tabular}{l cccc}
& \tt{ \small Failed Actions} & \tt{ \small Min. Dist. to Target (m)} & \tt{\small Stop-Fail. (Pos) (\%)} & \tt{\small Stop-Fail (Neg) (\%)}\\
\raisebox{2.5\normalbaselineskip}[0pt][0pt]{\rotatebox[origin=c]{90}{\tt{\pnav}}} & 
\includegraphics[width=\analysisWidth]{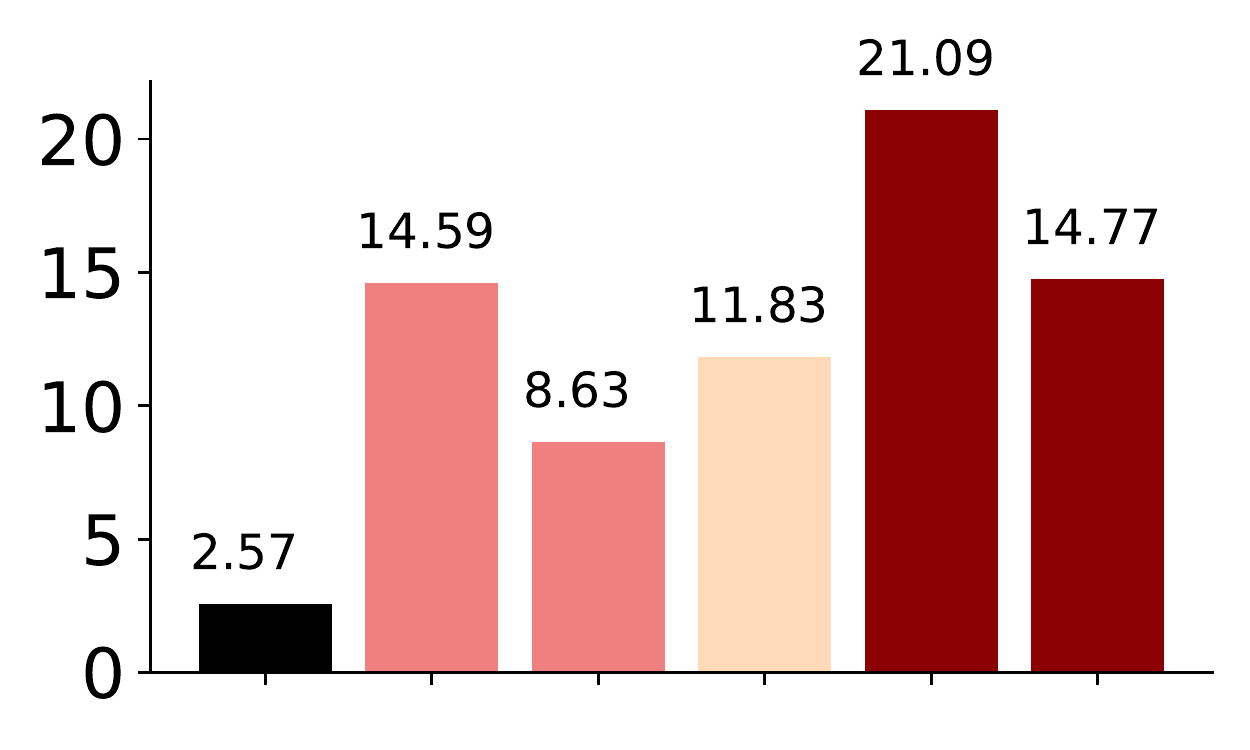}&
\includegraphics[width=\analysisWidth]{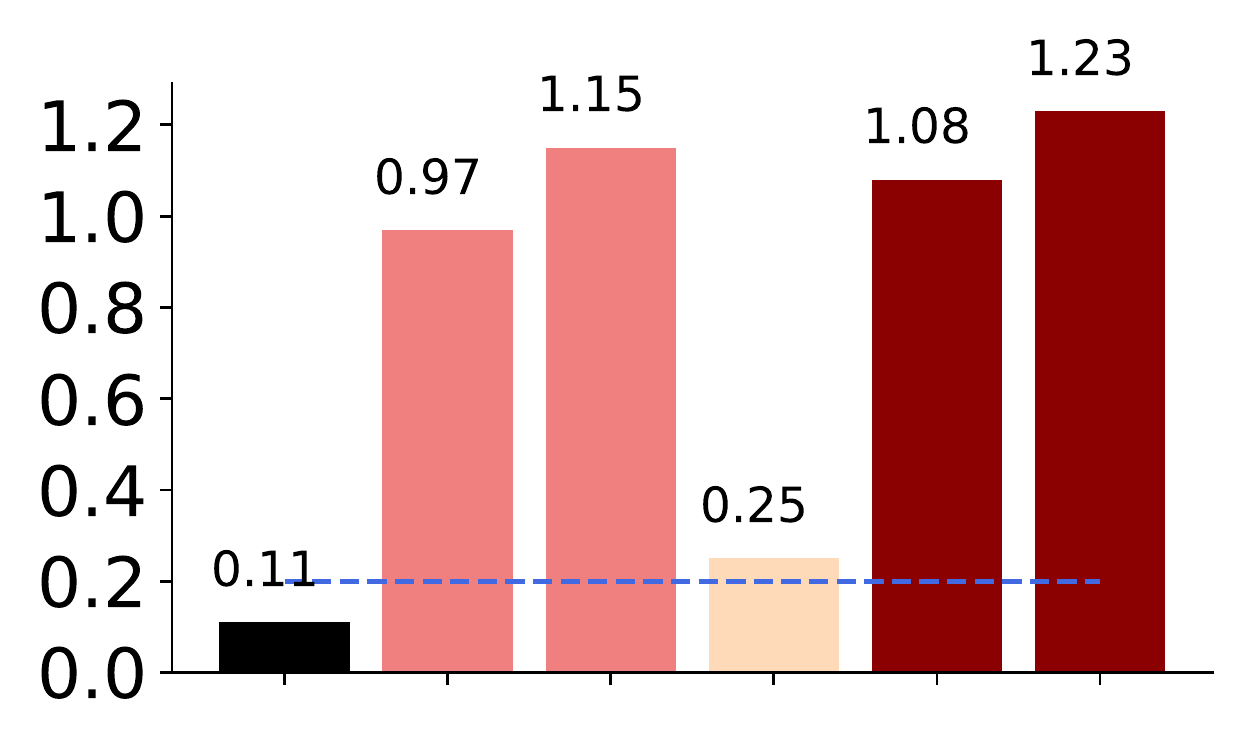} &
\includegraphics[width=\analysisWidth]{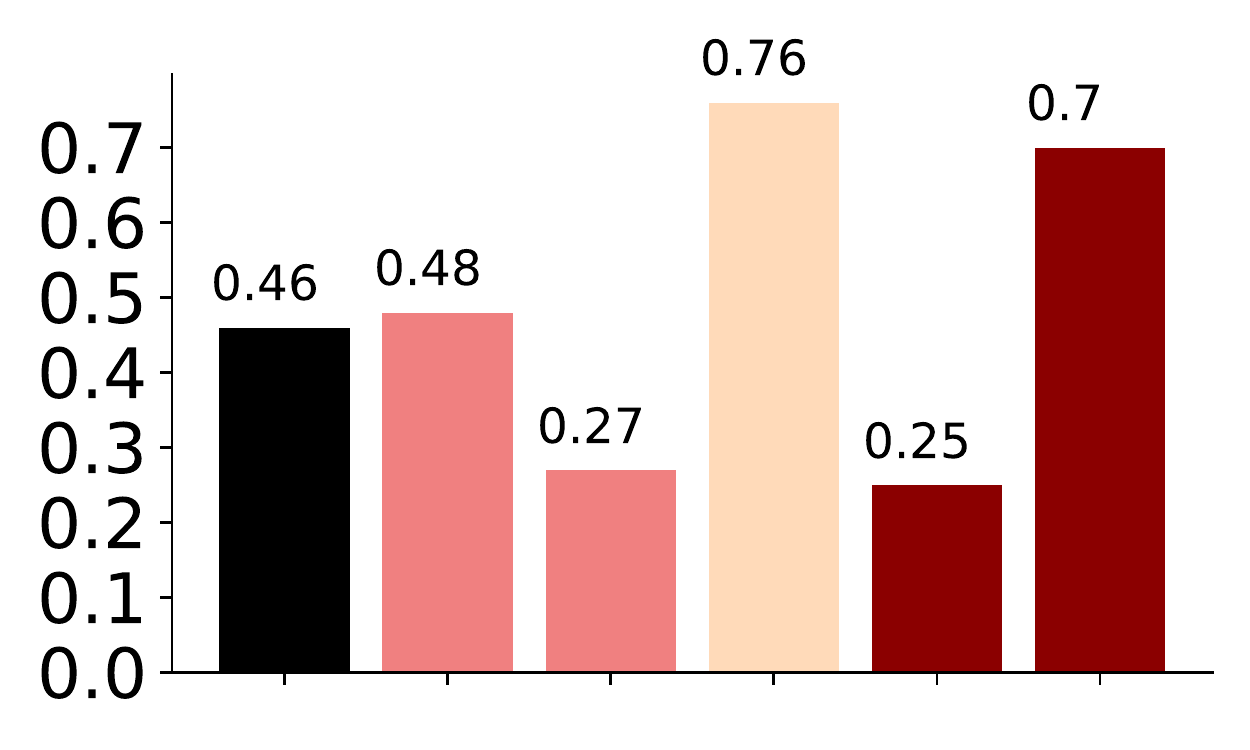}&
\includegraphics[width=\analysisWidth]{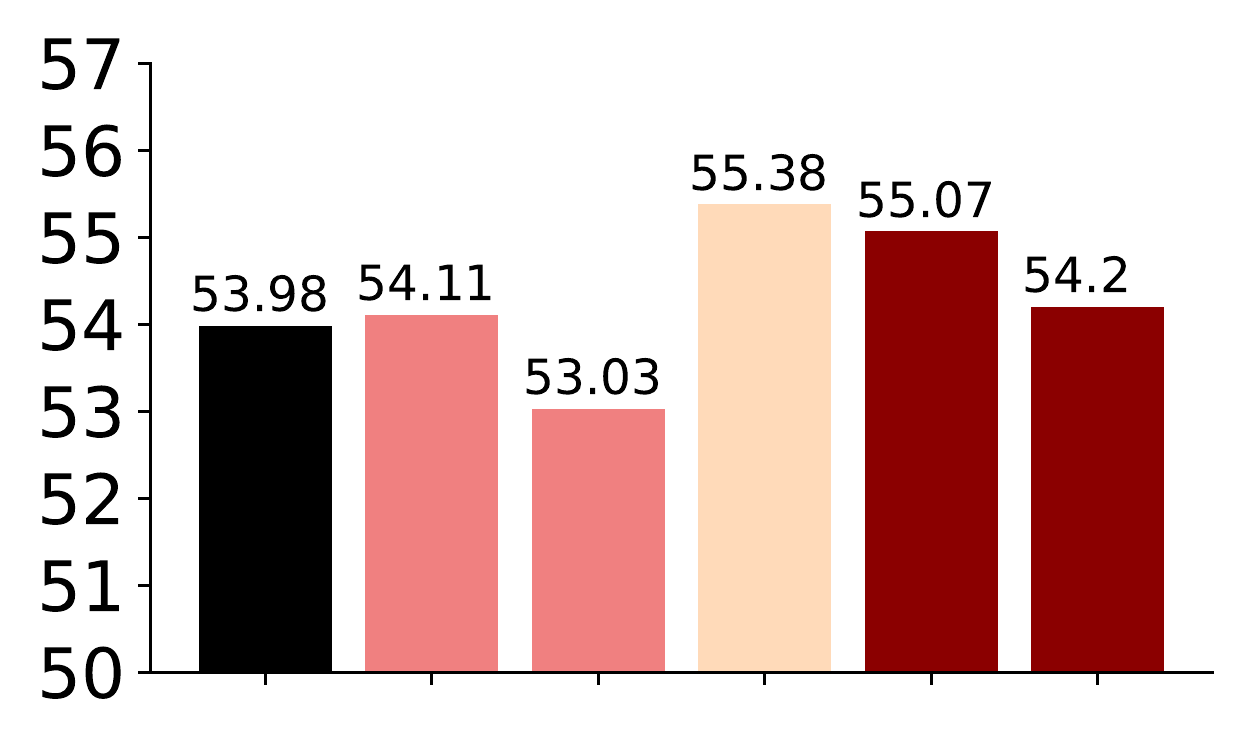}\\
\raisebox{4\normalbaselineskip}[0pt][0pt]{\rotatebox[origin=c]{90}{\tt{\onav}}} & 
\includegraphics[width=\analysisWidth]{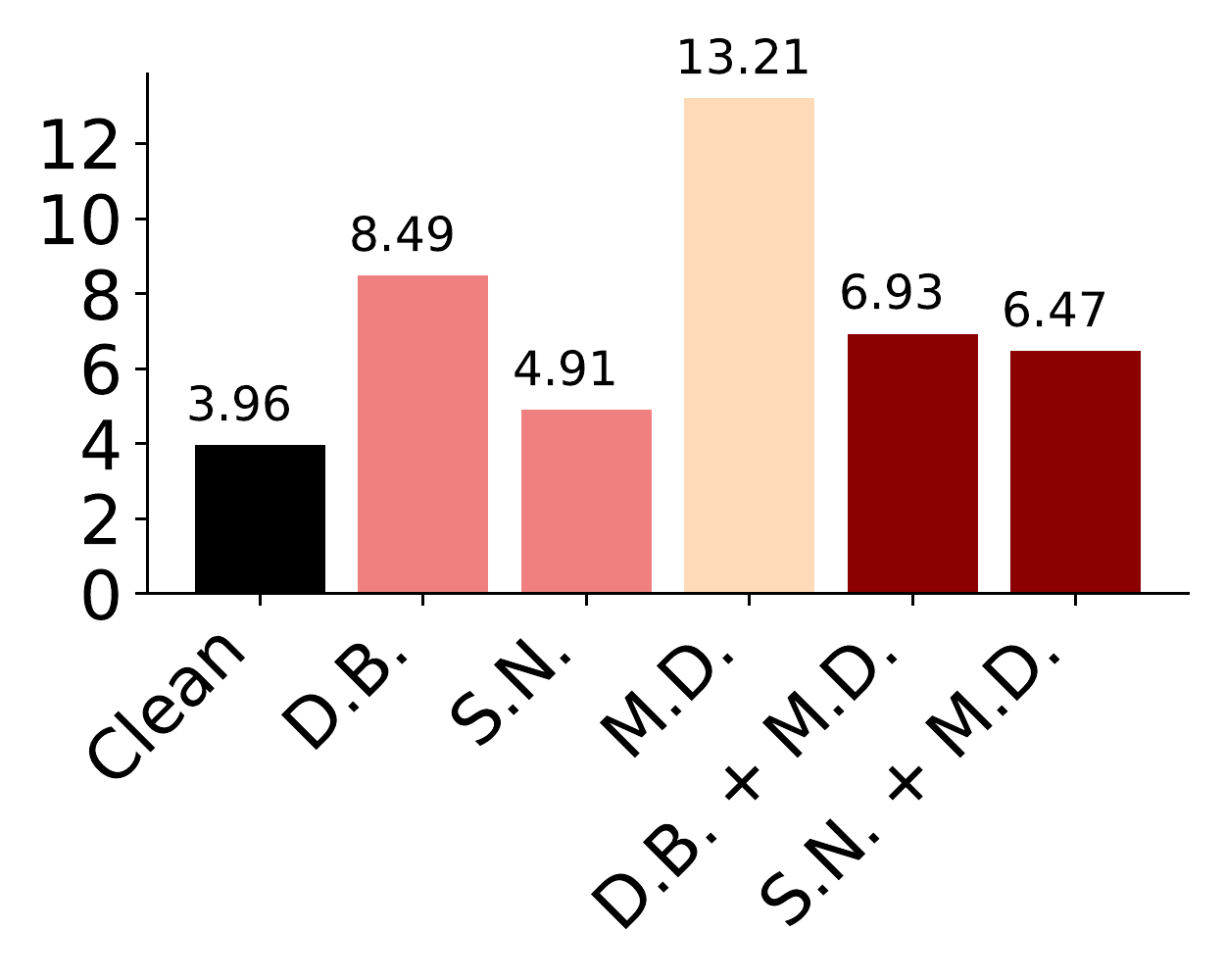}&
\includegraphics[width=\analysisWidth]{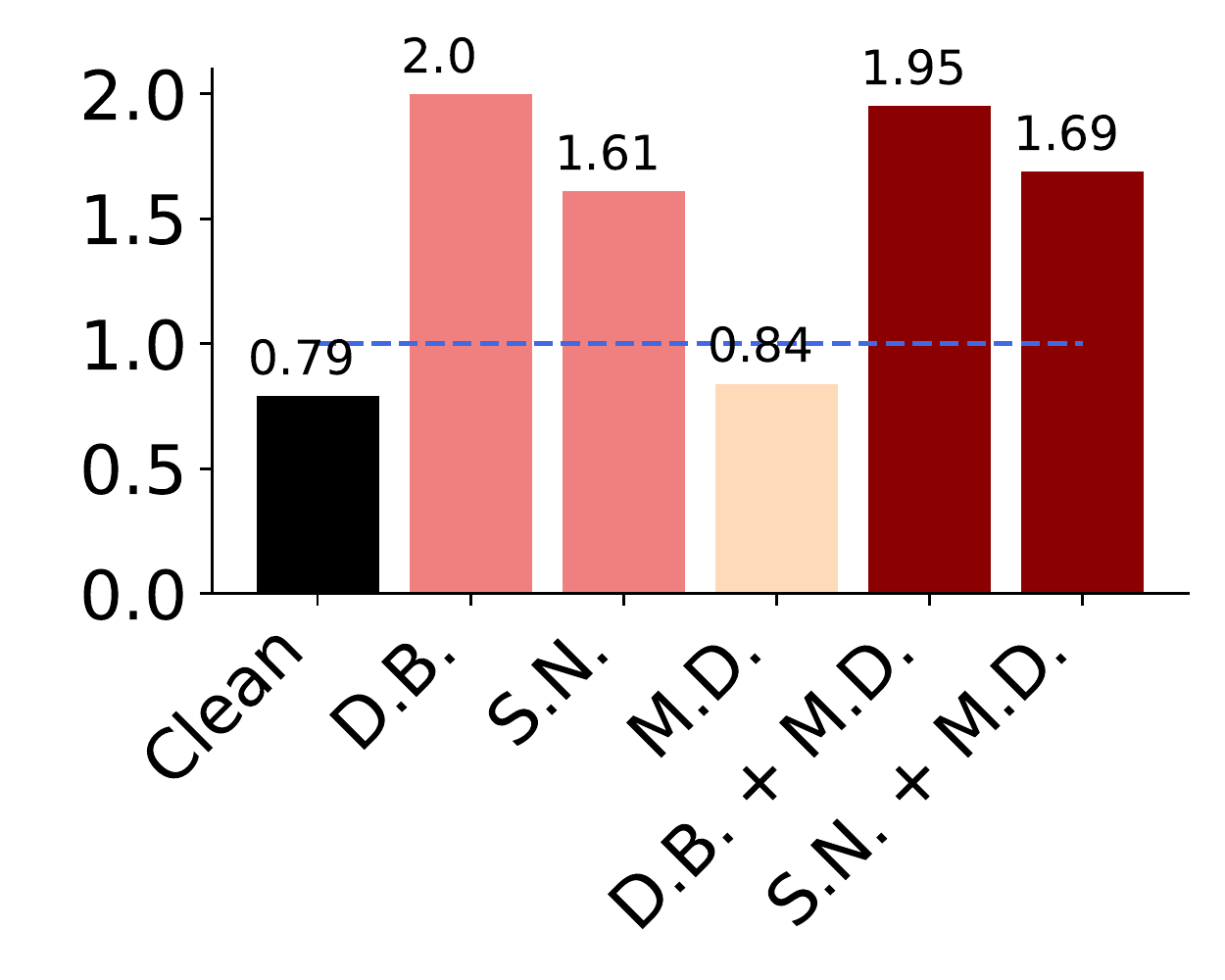}&
\includegraphics[width=\analysisWidth]{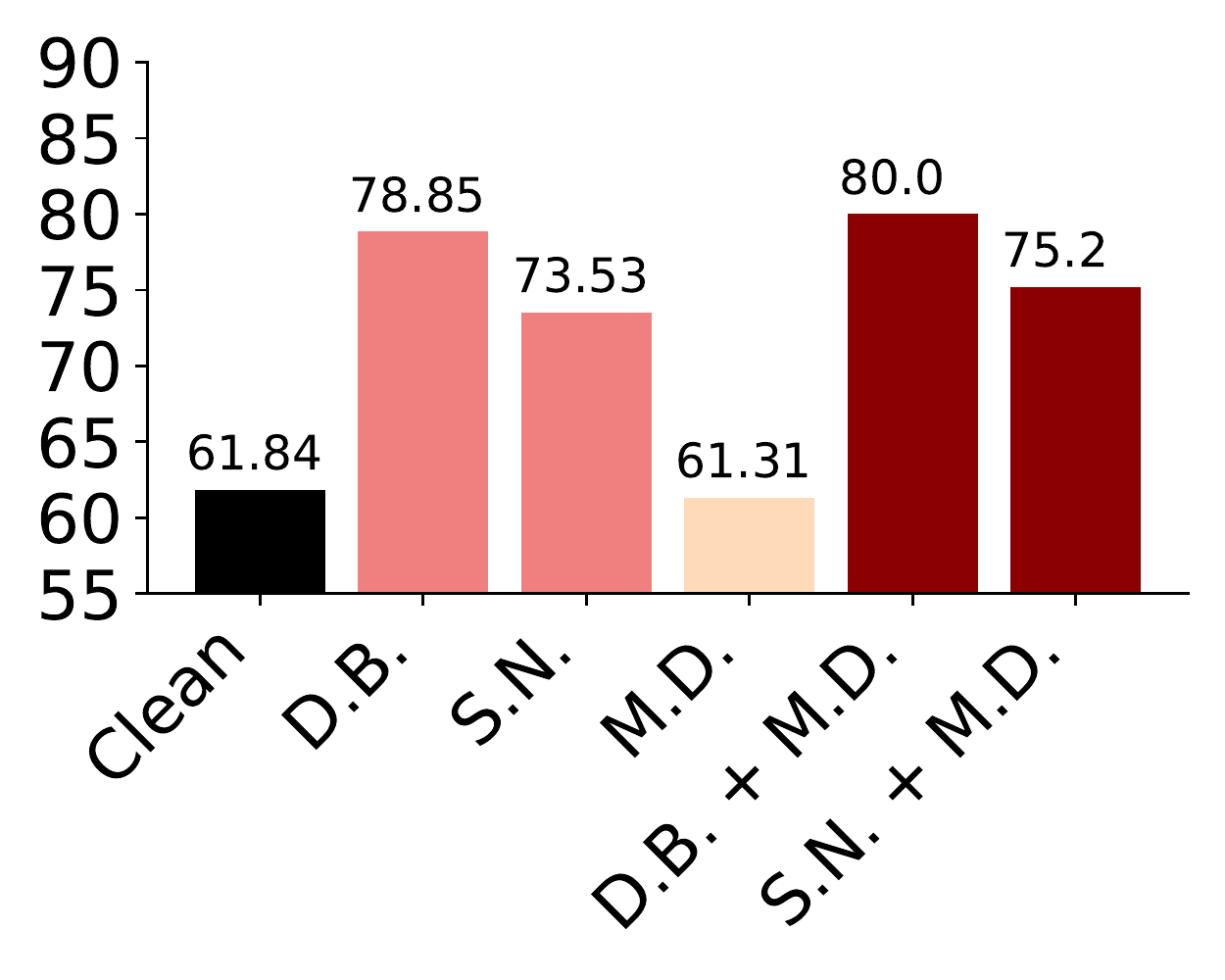}&
\includegraphics[width=\analysisWidth]{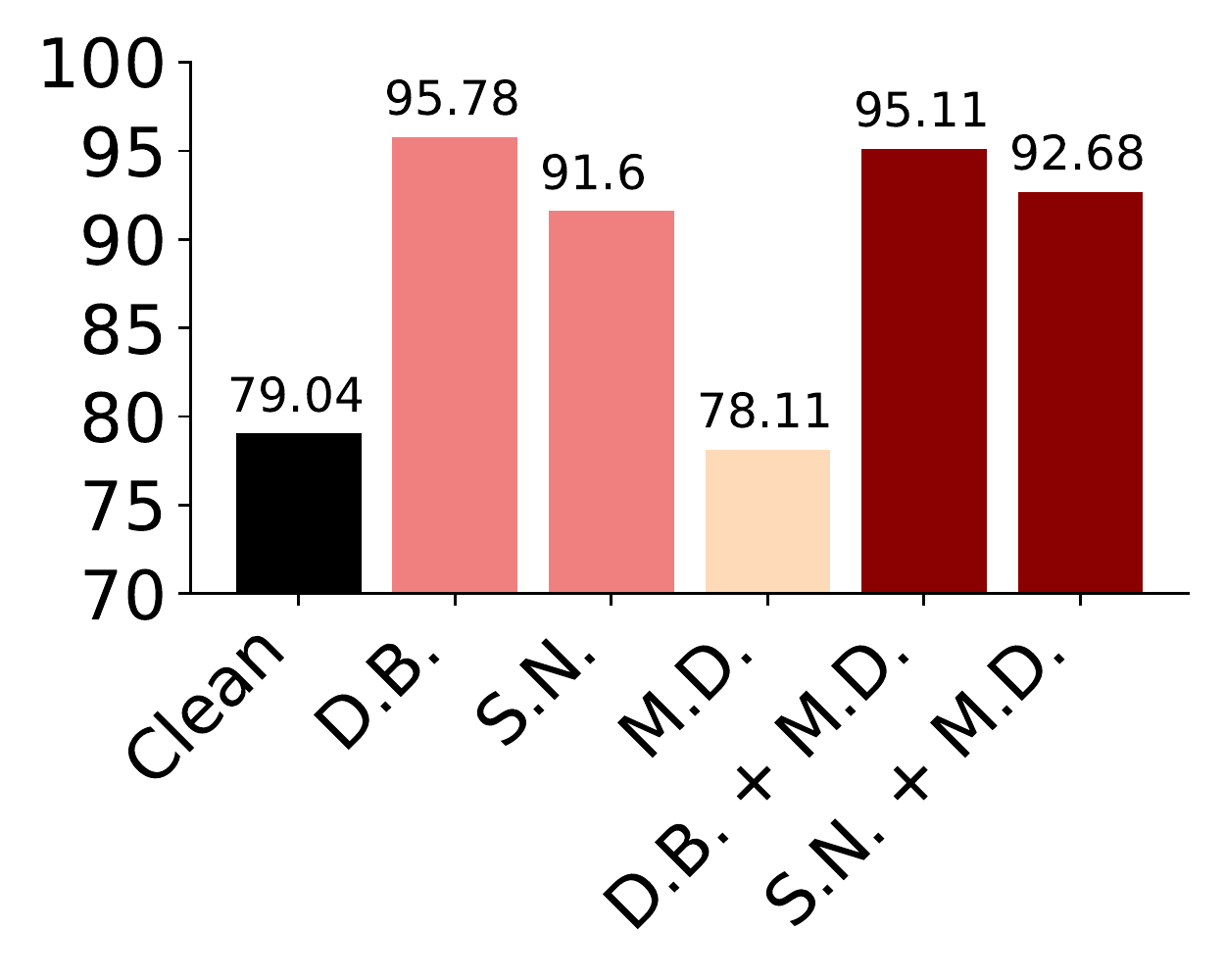}\\
\end{tabular}
\vspace{-7pt}
\caption{\textbf{Agent Behavior Analysis.} To understand agent behaviors, we report the breakdown of four metrics: Number of collisions as observed through \texttt{\small Failed Actions} (\textit{first column}), closest the agent was to target as measured by \texttt{\small Min. Dist. to Target} (\textit{second column}), and failure to appropriately end and episode either when out of range -- \texttt{\small Stop-Fail (Pos)} (\textit{third column}), or in range -- \texttt{\small Stop-Fail (Neg)} (\textit{fourth column}). Each behavior is reported for both \pnav (\textit{top row}) and \onav (\textit{bottom row}) \texttt{RGB} agents within a clean and five corrupt settings: Defocus Blur (D.B.), Speckle Noise (S.N.), Motion Drift (M.D.), Defocus Blur + Motion Drift, and Speckle Noise + Motion Drift. 
\protect\inlinegraphics{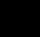} is clean, 
\protect\inlinegraphics{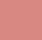} is \vcr~ corruptions, \protect\inlinegraphics{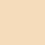} is \dcr~ corruptions and \protect\inlinegraphics{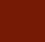} is \vdcr~ corruptions. 
Blue line in col 2 indicates the distance threshold for goal in range.  Severities for S.N. and D.B. are set to $5$ (worst).}
\label{fig:analysis_figure}
\end{figure*}

In \prithvi{Sec.~\ref{sec:deg_res} of appendix}, we further investigate
the degree to which more sophisticated \pnav agents,
composed of map-based architectures,
are susceptible to \vcr\ corruptions.
Specifically,
we evaluate the performance of the winning \pnav entry of
Habitat-Challenge (HC) 2020~\cite{habitat-challenge} -- Occupancy Anticipation (OccAnt)~\cite{ramakrishnan2020occant} on Gibson~\cite{xia2018gibson} val scenes under noise-free, Habitat Challenge conditions and \vcr\ corruptions.
We find that introducing corruptions under noise-free conditions
degrades navigation performance significantly only for \texttt{RGB}
agents. Under HC conditions,
\texttt{RGB-D} agents suffer drop in performance
as RGB noise is replaced with progressively severe \vcr\ corruptions.

\par \noindent
\textbf{Presence of \vdcr\ corruptions further degrades performance.}
Rows 14-19 in Table~\ref{tab:pnav_onav_base_results}
indicate the extent of performance degradation when 
\vdcr\ corruptions are present. 
With the exception of a few cases,
as expected, the drop in performance
is slightly more pronounced compared to the presence of just 
\vcr\
or 
\dcr\ corruptions.
The relative drop in performance from
\vcr\ $\rightarrow$ \vdcr\ is more pronounced for \onav as opposed
to \pnav.

\par \noindent
\textbf{Navigation performance for \texttt{RGB} agents degrades consistently with escalating episode difficulty.} 
Recall
that 
we evaluate navigation performance over epsisodes of varying difficulty levels
(see Sec.~\ref{sec:rnav}).
We break down the performance
of \pnav \& \onav agents by episode difficulty levels 
\prithvi{(in Sec.~\ref{sec:deg_res} of appendix).}
Under ``clean'' settings, we find that \pnav (\texttt{RGB} and \texttt{RGB-D})
have comparable performance
across all difficulty levels.
Under corruptions, we note that unlike the \texttt{RGB-D} counterparts,
\prithvi{performance of \pnav-\texttt{RGB} agents}
consistently deteriorates as the episodes become harder.
\prithvi{\onav (both \texttt{RGB} \& \texttt{RGB-D}) agents show a similar trend
of decrease in navigation performance with increasing episode difficulty.}

\vspace{\subsectionReduceTop}
\subsection{Behavior of Visual Navigation Agents}
\label{sec:behavior_analysis}
\vspace{\subsectionReduceBot}
We now study the idiosyncrasies (see Fig~\ref{fig:analysis_figure}) exhibited by these agents (\pnav-\texttt{RGB} and \onav-\texttt{RGB})
which leads to their degraded performance.

\par \noindent
\textbf{Agents tend to collide more often.} 
Fig~\ref{fig:analysis_figure} 
(\textit{first column}, bars color-coded based on the kind of corruption) %
shows the average number of failed actions under corrupt settings. In our framework, failed actions occur as a consequence of colliding with objects, walls, etc.
While corruptions generally lead to increased
collisions, we note that
adding a \dcr\ corruption 
in addition to 
a \vcr\ one (D.B. $\rightarrow$ D.B. + M.D. \& S.N. $\rightarrow$ S.N. + M.D.)
increases the number of collisions
over 
\vcr\ or \dcr\ corruptions -- \dcr\
corruptions lead to unforeseen changes in
dynamics (actions working unexpectedly), which likely contributes to an uptick in collisions.

\begin{table*}[ht!]
\footnotesize
\centering
\setlength{\tabcolsep}{7pt}
\resizebox{\linewidth}{!}{
\begin{tabulary}{\linewidth}{L  CC  CC  CC  CC  CC}
 \toprule
 \textbf{Approach} &  \multicolumn{10}{c}{\textbf{Visual Corruption} 
 }\\
 & \multicolumn{2}{c}{Clean} & \multicolumn{2}{c}{Lower-FOV} & \multicolumn{2}{c}{Defocus Blur} & \multicolumn{2}{c}{Camera Crack} & \multicolumn{2}{c}{Spatter}\\
 \midrule
  &\texttt{\textbf{SR}}$\uparrow$ & \texttt{\textbf{SPL}}$\uparrow$ &  \texttt{\textbf{SR}}$\uparrow$ & \texttt{\textbf{SPL}}$\uparrow$ &  \texttt{\textbf{SR}}$\uparrow$ & \texttt{\textbf{SPL}}$\uparrow$ &  \texttt{\textbf{SR}}$\uparrow$ & \texttt{\textbf{SPL}}$\uparrow$ &  \texttt{\textbf{SR}}$\uparrow$ & \texttt{\textbf{SPL}}$\uparrow$ \\
 \midrule
 \band \texttt{1} Nav. Loss&  98.82 &  83.13 & 42.49 &  31.73 & 75.89 &  53.55 & 82.07 &  63.83 & 33.58 &  24.72 \\
 \yband \texttt{2} Nav. Loss + AP&  98.45 &  83.28 & 45.68 &  35.14 & 83.35 &  61.51 & 72.70 &  56.82 & 20.38 &  15.70\\
 \yband \texttt{3} Nav. Loss + AP + SS-Adapt&  37.31 &  31.03 & 32.94 &  26.09 & 40.95 &  33.35 & 57.87 &  46.72 & 14.19 &  10.29 \\
 \gband \texttt{4} Nav. Loss + RP&  98.73 &  82.53 & 44.95 &  32.74 & 32.21 &  22.47 & 67.06 &  53.70 & 23.48 &  18.63 \\
 \gband \texttt{5} Nav. Loss + RP + SS-Adapt&  94.63 &  77.25 & 50.59 &  36.10 & 79.16 &  62.74 & 60.42 &  49.37 & 61.06 &  47.16 \\
 \rband \texttt{6} Nav. Loss + Data Aug&  98.45 &  81.08 & 71.70 &  54.54 & 81.26 &  61.32 & 88.44 &  71.57 &  23.93 &  18.41 \\
 \bband \texttt{7} Finetune Nav. Loss on Target&     -&  -& 72.88 & 61.82 & 97.18 & 80.32  & 96.54 & 80.92 & 91.81 & 77.38  \\

\bottomrule
\end{tabulary}
}
\vspace{1pt}
\caption{\textbf{Resisting Visual Corruptions.} To assist near-term progress, we study
if standard approaches towards training visually robust models or adapting to 
visual disparities can help resisting visual corruptions. 
All agents in rows 1-7 are \pnav \texttt{RGB} agents pre-trained for $\sim75$M frames.
Agents in rows 3 \& 5 have obtained by running adaptation for $\sim166$k steps. Agents in row 7 provide
an anecdotal
upper bound indicating attainable improvements when finetuned with task-supervision
under the calibration budget -- set to $\sim166$k steps.
For visual corruptions with controllable severity levels, we report
results with severity set to 5 (worst).
}
\label{tab:vis_corr_adapt}
\end{table*}

\par \noindent
\textbf{Agents tend to be farther from the target.}
Fig~\ref{fig:analysis_figure} 
(\textit{second column}) %
shows the minimum distance from the target over the course of an
episode. While we note that as corruptions become progressively
severe, agents tend to terminate farther away from the target 
(see Sec.~\ref{sec:behav} of appendix), 
Fig~\ref{fig:analysis_figure} 
(\textit{second column}) %
indicates that the overall proximity of the agent to the goal
over an episode decreases -- minimum distance to target
increases as we go from Clean $\rightarrow$ \vcr\ or \dcr;
\vcr\ or \dcr\ $\rightarrow$ \vdcr.
While this may be intuitive in the presence of a \dcr\ corruption, it is interesting to note that this trend is also consistent for \vcr\ corruptions (Clean $\rightarrow$ D.B. or S.N.).

\par \noindent
\textbf{Corruptions hurt \onav stopping mechanism.}
Recall that for both \pnav and \onav, success depends on the notion
of ``intentionality''~\cite{batra2020objectnav} -- the agent calls
 an \ed action when it believes it has reached the goal.
In Fig~\ref{fig:analysis_figure} 
(\textit{last two columns}) %
we aim to
understand how corruptions affect this stopping mechanism.
Specifically, we look at two quantitative measures -- (1) Stop-Failure
 (Positive), the proportion of times the agent invokes
 an \ed action when the goal is not range; and (2) Stop-Failure (Negative), the proportion of times the agent does not invoke an
 \ed action when the goal is in range, out of the number of times the goal is in range.\footnote{
 The
 goal in range 
 criterion for \pnav checks if the target is within the threshold
 distance. For \onav, this includes an additional visibility criterion.
 }

We observe that prematurely calling an \ed action
is a significant issue only for \onav (Fig~\ref{fig:analysis_figure} (\textit{third column})) -- which becomes
more pronounced as corruptions become progressively severe (Clean $\rightarrow$ D.B. or S.N.; M.D. $\rightarrow$ D.B. + M.D. or S.N. + M.D.). Similarly,
the inability of an agent to invoke an \ed action is also more pronounced
for \onav as opposed to \pnav (Fig~\ref{fig:analysis_figure} (\textit{fourth column})). 
To investigate the extent to which this impacts
the agent's performance, we compare the agent's Success Rate (SR) with
a setting where the agent is equipped with an oracle stopping mechanism
(call \ed as soon as the goal is in range). We find that this makes a significant
difference only for \onav -- 
absolute $+7.12\%$ for Clean, $+7.76\%$ for M.D. and
$+13.88\%$ for D.B. + M.D.
We hypothesize that equipping agents with robust
stopping mechanisms can significantly improve performance on \rnav.
For instance, equipping the agent with a progress monitor module~\cite{ma2019self}
(estimating progress made towards the goal in terms of distance)
robust to \vcr\ corruptions can potentially help decide when explicitly
to invoke an \ed action in the target environment.

\vspace{\subsectionReduceTop}
\subsection{Resisting Corruptions}
\label{sec:resist_deg}
\vspace{\subsectionReduceBot}
To assist near-term progress, we investigate if some standard approaches
towards training robust models or adapting to visual disparities can help 
resisting
\vcr\ corruptions under a calibration budget (Sec.~\ref{sec:rnav}) -- set to $\sim166$k steps.\footnote{Based on the number of steps it takes an agent to reasonably recover degraded performance in corrupted environments when finetuned with complete task supervision.}

\par \noindent
\textbf{Extent of attainable improvement by finetuning under task supervision.}
As an anecdotal upper bound on attainable improvements under the calibration-budget, we also report the extent
to which degraded performance can be recovered when fine-tuned under complete task supervision.
We report these results for \vcr\ corruptions in Table~\ref{tab:vis_corr_adapt} (row 7).
We note that unlike Lower-FOV,
the agent is able to almost recover performance for Defocus Blur, Camera-Crack and
Spatter (Table.~\ref{tab:vis_corr_adapt}, rows 1,7).

\par \noindent
\textbf{Do data-augmentation strategies help?}
In Table~\ref{tab:vis_corr_adapt}, we study if data-augmentation strategies improve zero-shot resistance to \vcr\ corruptions (rows 1,6). We compare \pnav \texttt{RGB} agents trained with
Random-Crop, Random-Shift and Color-Jitter (row 6) with the vanilla versions (row 1) and 
find that while data augmentation (row 6)
offers some improvements (Spatter being an exception) over degraded performance (row 1) --
absolute improvements of
($22.81\%$ SPL, $29.21\%$ SR) for Lower-FOV, ($7.77\%$ SPL, $5.37\%$ SR) for Defocus Blur and
($7.74\%$ SPL, $6.37\%$ SR) for Camera-Crack, obtained performance
is still significantly below Clean settings (row 1, Clean col).
Improvements are more pronounced for Lower-FOV 
compared to others (likely due to Random-Shift and Random-Crop).
We note that
data-augmentation
provides improvements only for a subset of
\vcr\ corruptions and when it does, obtained improvements are still not
sufficient enough to recover lost performance.

\par \noindent
\textbf{Do self-supervised adaptation approaches help?} 
In the absence of reward supervision in the target environment, Hansen \etal~\cite{hansen2020self} proposed Policy Adaptation during Deployment (PAD) -- source pretraining with an auxiliary supervised objective and optimizing only the self-supervised objective when deployed in the target environment. 
We investigate the degree to which PAD helps adapting to the target
environments in \rnav.
The adopted self-supervised tasks are 
(1) Action-Prediction (AP) -- given two successive observations in a trajectory, predict
the intermediate action and (2) Rotation-Prediction (RP) -- rotate the input observation
by $0$\degr, $90$\degr, $180$\degr, or $270$\degr before feeding it to the agent 
and task an
additional auxiliary head 
with predicting the rotation.
We report numbers with AP (rows 2,3)
and RP (rows 4,5) in 
Table.~\ref{tab:vis_corr_adapt}.
For AP, we find that (1) pre-training (row 2 vs row 1) results in
little or no improvements over degraded performance (maximum absolute improvements
of $7.96\%$ SPL, $7.46\%$ SR for Defocus Blur)
and (2) further adaptation (row 3 vs rows 2,1) under calibration budget consistently degrades performance.
For RP, we observe that (1) with the exception of Clean and Lower-FOV, pre-training
(row 4 vs row 1) results in worse performance and (2) while self-supervised
adaptation under corruptions improves performance over pre-training (row 5 vs row 4),
it is still significantly below Clean settings (row 1, Clean col)
-- minimum absolute gap of $20.39\%$ SPL, $19.66\%$ SR
between Defocus Blur (row 5) and Clean (row 1).
While
improvements over degraded performance might highlight the utility
of PAD (with AP / RP) as a potential unsupervised adaptation approach,
there is still a long way to go in terms of closing the performance
gap between clean and corrupt settings.

\section{Conclusion}
\label{sec:conclusion}
In summary, as a step towards assessing general
purpose robustness of embodied navigation agents,
we propose \rnav, a challenging framework well-suited to benchmark the robustness of embodied navigation agents, with a wide variety of \textit{visual} and \textit{dynamics} corruptions.
To succeed on \rnav, an agent must be insensitive to
corruptions and also be able to adapt to unforeseen changes in
new environments with minimal interaction. We find that standard
\pnav and \onav agents underperform (or fail) significantly in the presence
of corruptions and 
while standard techniques to improve robustness or
adapt to environments with visual disparities (data-augmentation, self-supervised adaptation) provide
some improvements, 
a large room for improvement remains
in terms of fully recovering lost navigation performance.
Lastly, we plan on evolving \rnav in terms of the sophistication and
diversity of corruptions as more features are supported in the
underlying simulator.
We release \rnav in \rthor, and hope
that our findings provide insights into developing more
robust navigation agents.
\par \noindent
\textbf{Acknowledgements.} We thank Klemen Klotar, Luca Weihs, Martin Lohmann, Harsh Agrawal and Rama Vedantam 
 for fruitful discussions and valuable feedback. 
 We thank Winson Han for helping out with the Camera-Crack \vcr\ corruption.
 We also thank Vishvak Murahari for helping out with the ImageNet experiments and for sharing code for the Mask-RCNN experiments. This work is supported by the NASA University Leadership Initiative (ULI) under grant number 80NSSC20M0161.

{\small
\bibliographystyle{ieee_fullname}
\bibliography{egbib,main}
}

\clearpage
\appendix
\section{Appendix}
\localtableofcontents

\subsection{Overview}
\label{sec:oview}
This appendix is organized as follows.
In Sec.~\ref{sec:task_spec}, we describe in detail the task
specifications for \pnav and \onav. In Sec.~\ref{sec:agent},
we provide details about the 
architecture adopted for \pnav and \onav agents and how they
are trained.
In Sec.~\ref{sec:behav}, we include more 
plots demonstrating the kinds of behaviors
\pnav and \onav agents exhibit under corruptions (\texttt{\texttt{RGB}-D} variants
in addition to the \texttt{RGB} variants in Sec.~\ref{sec:behavior_analysis} of the main paper).
In Sec.~\ref{sec:deg_res}, we provide more results demonstrating degradation
in performance at severity set to $3$ (for \vcr\ corruptions with controllable
severity levels; excluded from the main paper due to space
constraints) and break down performance degradation by episode difficulty.

\subsection{Task Specifications}
\label{sec:task_spec}
We describe the task-specifications (as outlined in Sec.~\ref{sec:rnav} of the main paper) for the ones
included in \rnav in detail. Note that while \rnav currently supports navigation heavy tasks,
the corruptions included can easily be extended to other embodied tasks that share the same
modalities, for instance, tasks involving vision and language guided navigation or having interaction
components.

\par \noindent
\textbf{\pnav.} In \pnav, an agent is spawned at a random location
and orientation
in an environment
and asked to navigate to goal coordinates specified relative to the agent's position. 
This is equivalent to the agent being
equipped with a GPS+Compass sensor (providing relative location and orientation with respect
to the agent's
current position). Note that the agent does not have access to any ``map" of the
environment and must navigate based solely on sensory inputs from a visual \texttt{RGB} (or \texttt{\texttt{RGB}-D})
and GPS+Compass sensor. An episode is declared successful if the \pnav agent stops (by
``intentionally'' invoking an \ed action) within $0.2$m of the goal location.

\begin{figure}[t!]
\centering
\includegraphics[width=\linewidth]{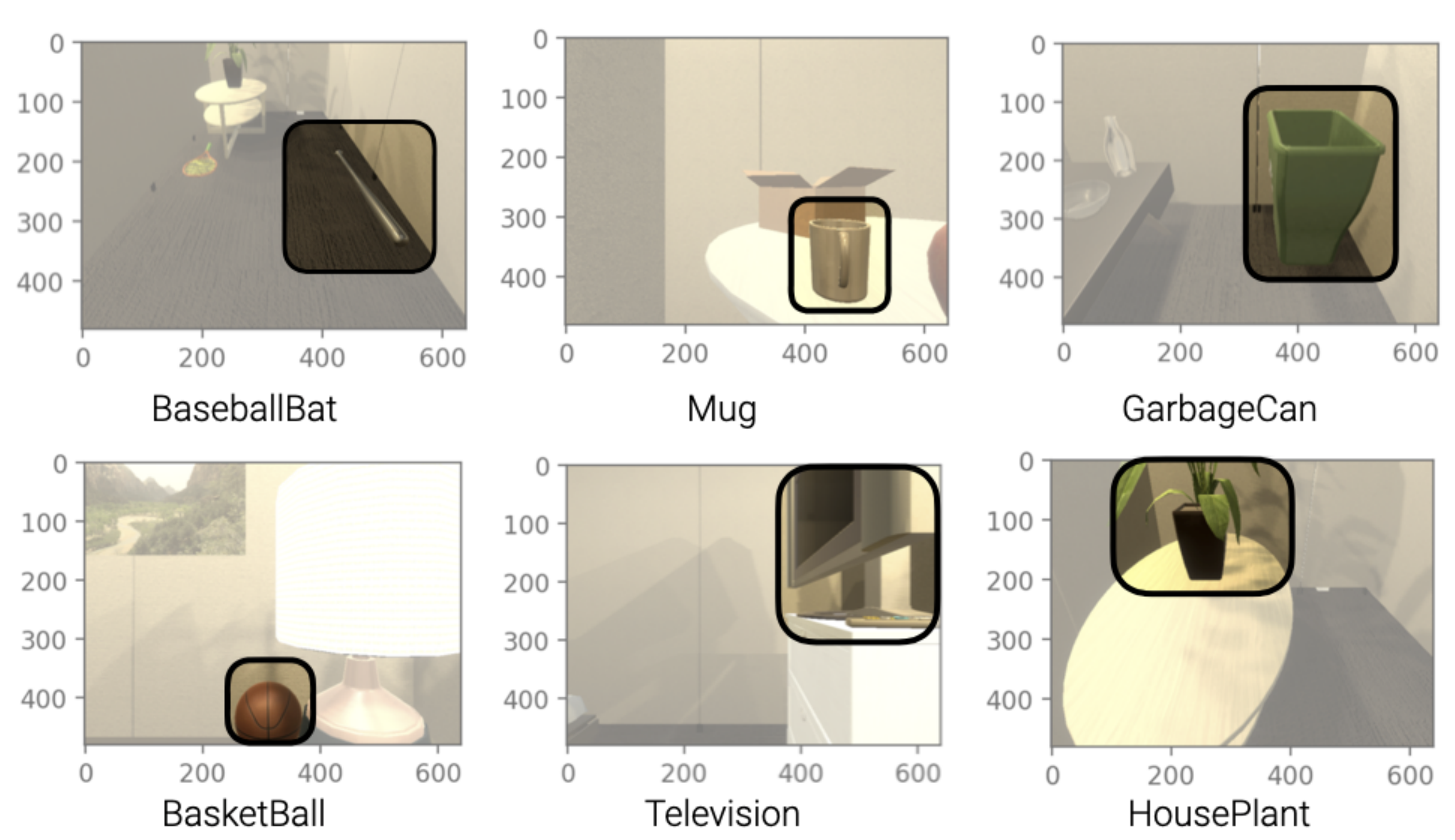}
\caption{\textbf{\onav Target Objects.} We present a few examples of the
target objects considered for \onav agents in \rthor as viewed from
the agent's ego-centric \texttt{RGB} frame under successful episode termination
conditions.}
\label{fig:objnav_targets}
\end{figure}

\par \noindent
\textbf{\onav.} In \onav, an agent is spawned at a random location and orientation in
an environment as is asked to navigate to a specified ``object'' category (e.g, \textit{Television})
that exists in the environment. Unlike \pnav, an \onav agent does not have access to
a GPS+Compass sensor and must navigate based solely on the specified target and
visual sensor inputs -- \texttt{RGB} (or \texttt{\texttt{RGB}-D}). An episode is declared successful if the \onav
agent (1) stops (by
``intentionally'' invoking an \ed action) within $1.0$m of the target object and (2)
has the target object within it's ego-centric view. We consider $12$ object categories
present in the \rthor scenes for our \onav experiments. These are AlarmClock, Apple, BaseballBat, BasketBall,
Bowl, GarbageCan, HousePlant, Laptop, Mug, SprayBottle, Television and Vase
(see Fig.~\ref{fig:objnav_targets} for a few examples in the agent's ego-centric frame).

\subsection{Navigation Agents}
\label{sec:agent}
\begin{figure}[t!]
\centering
\includegraphics[width=\linewidth]{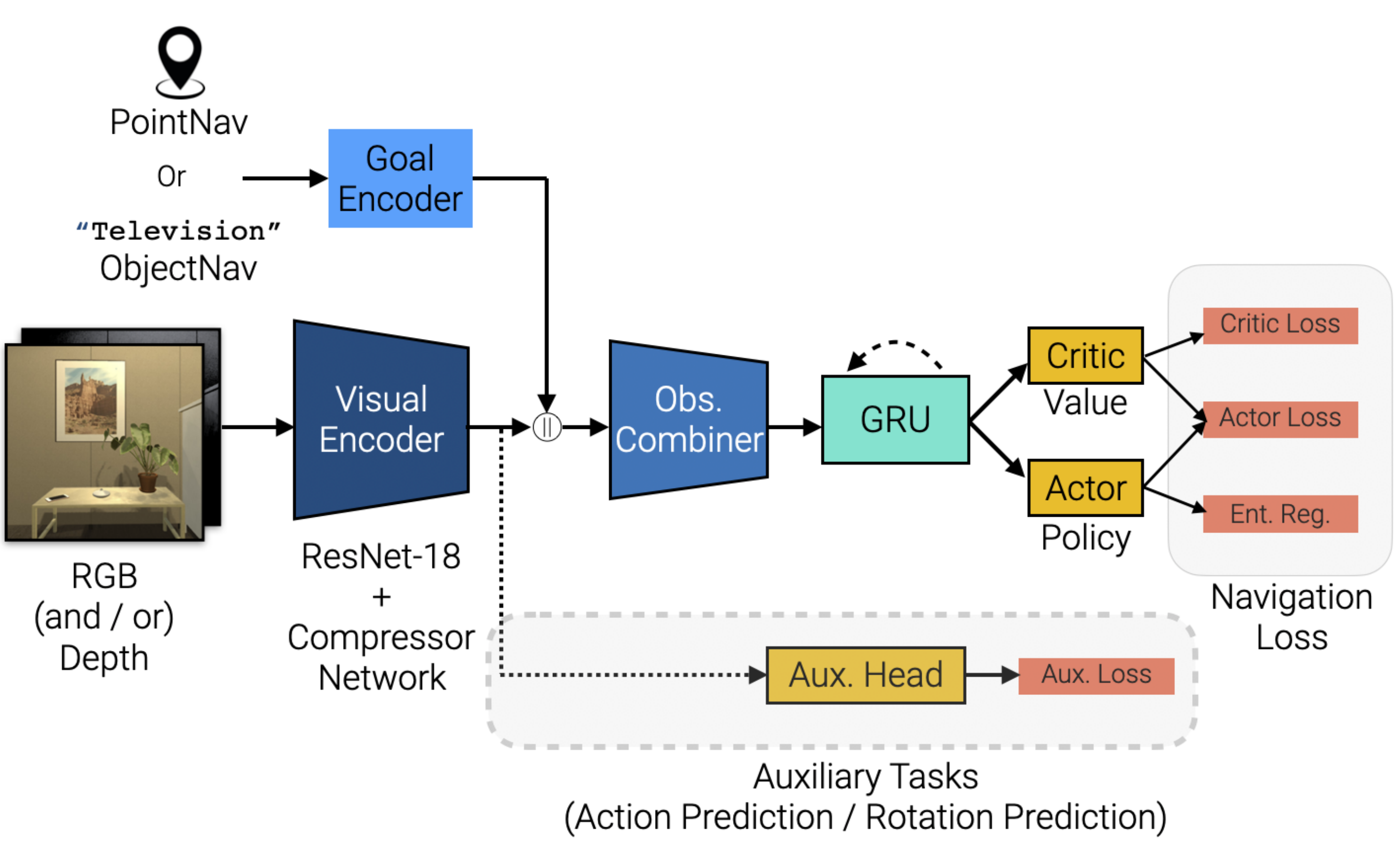}
\caption{\textbf{Agent Architecture.} We show the 
general architecture adopted for our \pnav and \onav agents --
convolutional units to encode observations followed by
recurrent policy networks. The auxiliary task heads are used
when we consider pre-training or adaptation using PAD~\cite{hansen2020self}.}
\label{fig:agent_arch}
\end{figure}

We highlight describe the architecture of
the agents studied in \rnav and provide additional training
details.
\par \noindent
\textbf{Base Architecture.}
We consider standard neural architectures (akin to~\cite{wijmans2019decentralized,AllenAct})
for both \pnav and \onav -- convolutional units to encode
observations followed by recurrent policy networks to predict
action distributions.
Concretely, our agent architecture consists of four major components -- a visual encoder, a goal
encoder, a target observation combiner and a policy network (see Fig.~\ref{fig:agent_arch}). The visual encoder (for \texttt{RGB} and \texttt{\texttt{RGB}-D}) agents
consists of a frozen ResNet-18~\cite{he2016deep} encoder (till the last residual block) pretrained on ImageNet~\cite{deng2009imagenet}, followed by
a learnable compressor network consisting of two convolutional layers of kernel size 1, each followed by ReLU activations
($512\times7\times7\rightarrow128\times7\times7\rightarrow32\times7\times7$).
The goal
encoder encodes the specified target -- a goal location in polar coordinates ($r,\theta$) for \pnav and the target
object token (\textit{e.g., Television}) for \onav. For \pnav, the goal is encoded via
a linear layer ($2\times32$). For \onav, the goal is encoded via an embedding layer ($12\times32$) set to
encode one of the 12 object categories.
The goal embedding and output of the visual encoder are then
concatenated and
further passed through the target observation combiner network consisting of two convolutional layers of kernel size 1 ($64\times7\times7\rightarrow128\times7\times7\rightarrow32\times7\times7$).
The output of the target observation combiner is flattened and then
fed to the policy network -- specifically, to a single layer GRU (hidden size $512$), 
followed by
linear actor and critic heads used to predict
action distributions and value estimates.

\par \noindent
\textbf{Auxiliary Task Heads.} In Sec.~\ref{sec:resist_deg} of the main paper,
we investigate if self-supervised approaches, particularly,
Policy Adaptation during Deployment (PAD)~\cite{hansen2020self} help
in resisting performance degradation due to \vcr\ corruptions.
Incorporating PAD involves training the vanilla agent architectures (as highlighted before)
with self-supervised tasks (for pre-training as well as
adaptation in a corrupt target environment) -- namely, Action
Prediction (AP) and Rotation Prediction (RP). In Action-Prediction (AP), given two successive observations in a trajectory, an
auxiliary head is tasked with
predicting
the intermediate action 
and in Rotation-Prediction (RP), the input observation
is rotated
by $0$\degr, $90$\degr, $180$\degr, or $270$\degr uniformly
at random
before feeding to the agent and an auxiliary head is asked to
to predict the rotation bin. For both AP and RP,
the auxiliary task heads operate on the encoded visual
observation (as shown in Fig.~\ref{fig:agent_arch}). To gather
samples in the target environment (corrupt or otherwise), we use data
collected from trajectories under the 
source (clean) pre-trained
policy
-- \textit{i.e.,} the visual encoder is updated
online as observations are encountered under the pre-trained
policy.

\par \noindent
\textbf{Training and Evaluation Details.}
As mentioned earlier, we train our agents with DD-PPO~\cite{wijmans2019decentralized} (a decentralized, distributed version of the Proximal Policy Optimization Algorithm~\cite{schulman2017proximal}) with Generalized Advantage Estimation~\cite{schulman2015high}. We use rollout lengths $T=128$, 4 epochs of PPO with 1 mini-batch per-epoch.
We set the discount factor to $\gamma=0.99$, GAE factor to $\tau=0.95$, PPO clip parameter to
$0.1$, value loss coefficient to $0.5$ and clip the gradient norms at $0.5$. We use the the Adam optimizer~\cite{kingma2014adam}
with a learning rate of $3e-4$ with linear decay.
The reward structure used is as follows -- if $\mathtt{R} = 10.0$ denotes the terminal reward obtained at the end of a ``successful'' episode and $\lambda = -0.01$ denotes a slack penalty to encourage efficiency,
then the reward received by the agent at time-step $t$ can be expressed as,
\begin{align}
  r_t =
    \begin{cases}
      \mathtt{R} \cdot \mathbb{I}_{\text{Success}} & \text{if $a_t=$ \ed}\\
      -\Delta^{\text{Geo}}_t + \lambda & \text{otherwise}
    \end{cases}       
\end{align}
where $\Delta^{\text{Geo}}_t$ is the change in geodesic distance to the goal,
$a_t$ is the action taken by the agent and
$\mathbb{I}_{\text{Success}}$ indicates where the episode
was successful ($1$) or not ($0$).
During evaluation,
we allow an agent to execute a maximum of $300$ steps -- if an agent doesn't call \ed within
$300$ steps, we forcefully terminate the episode.
All agents are trained under LocoBot calibrated actuation noise models from~\cite{robothor} -- $\mathcal{N}(0.25\text{m}, 0.005\text{m})$ for translation and 
$\mathcal{N}(30^{\circ}, 0.5^{\circ})$ for
rotation. During evaluation, with the exception of circumstances when Motion Bias (S) is
present, we use the same actuation noise models (in addition to \dcr\ corruptions when applicable).
We train our \pnav agents for $\sim75$M steps and \onav agents for $\sim300$M steps (both \texttt{RGB} and \texttt{RGB-D} variants).

\subsection{Behavior Analysis}
\label{sec:behav}
\newcommand{\suppanalysisWidth}{.15\linewidth}
\begin{figure*}[ht!]
\centering
\begin{tabular}{l ccccc}
& \scriptsize{\texttt{Failed Actions}} & \scriptsize{\texttt{Term. Dist. to Target (m)}} & \scriptsize{\texttt{Min. Dist. to Target (m)}} & \scriptsize{\texttt{Stop-Fail. (Pos) (\%)}} & \scriptsize{\texttt{Stop-Fail (Neg) (\%)}}\\
\raisebox{1.5\normalbaselineskip}[0pt][0pt]{\rotatebox[origin=c]{90}{\tt{\scriptsize{PNav-\texttt{RGB}}}}} & 
\includegraphics[width=\suppanalysisWidth]{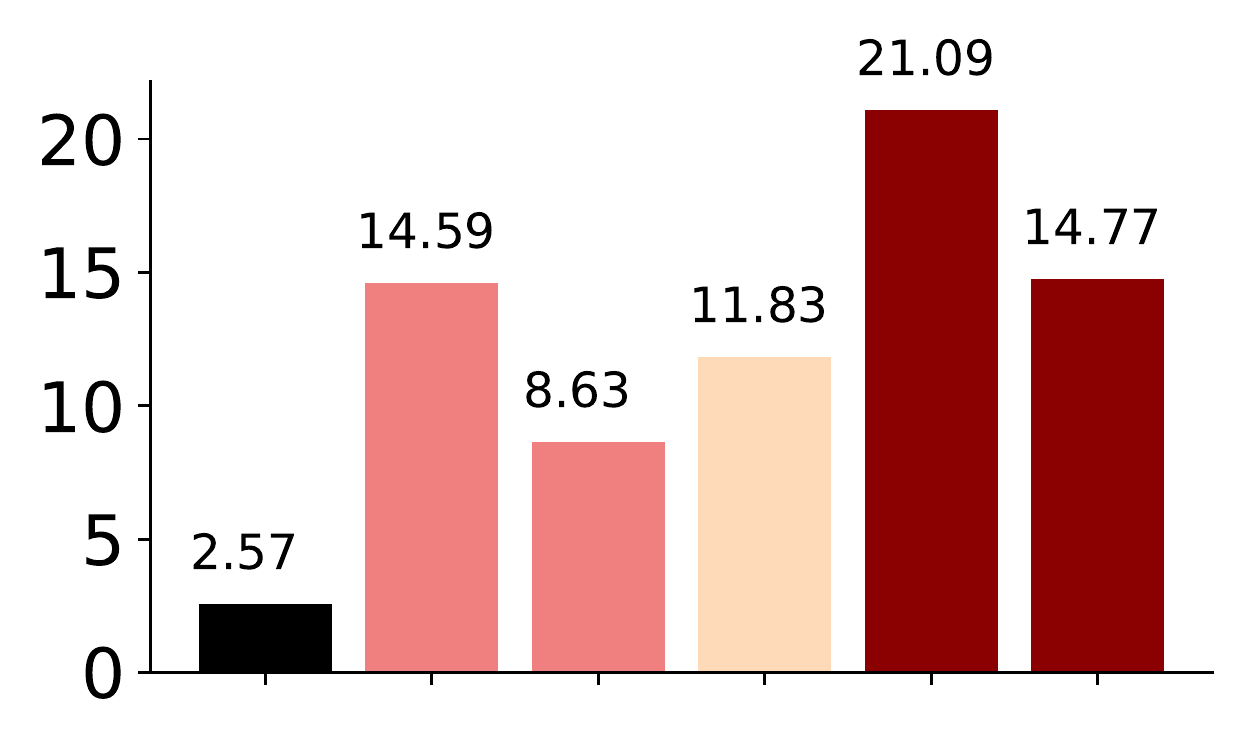}&
\includegraphics[width=\suppanalysisWidth]{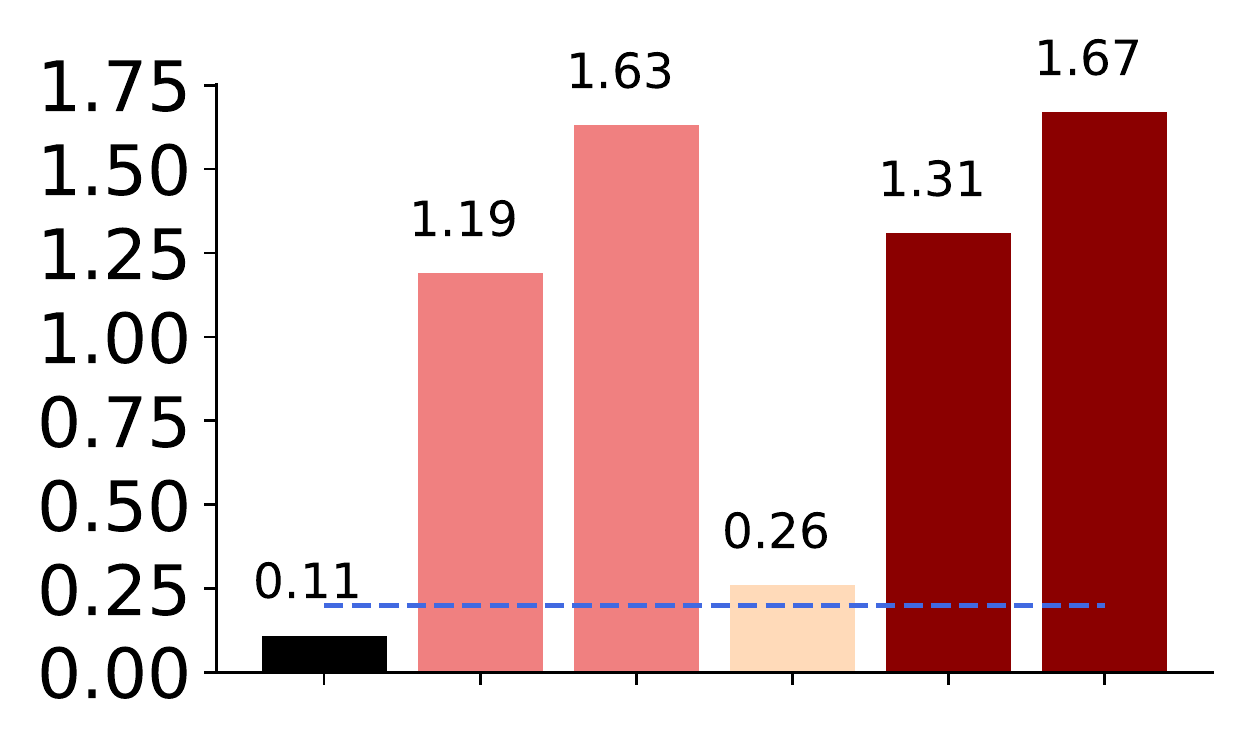} &
\includegraphics[width=\suppanalysisWidth]{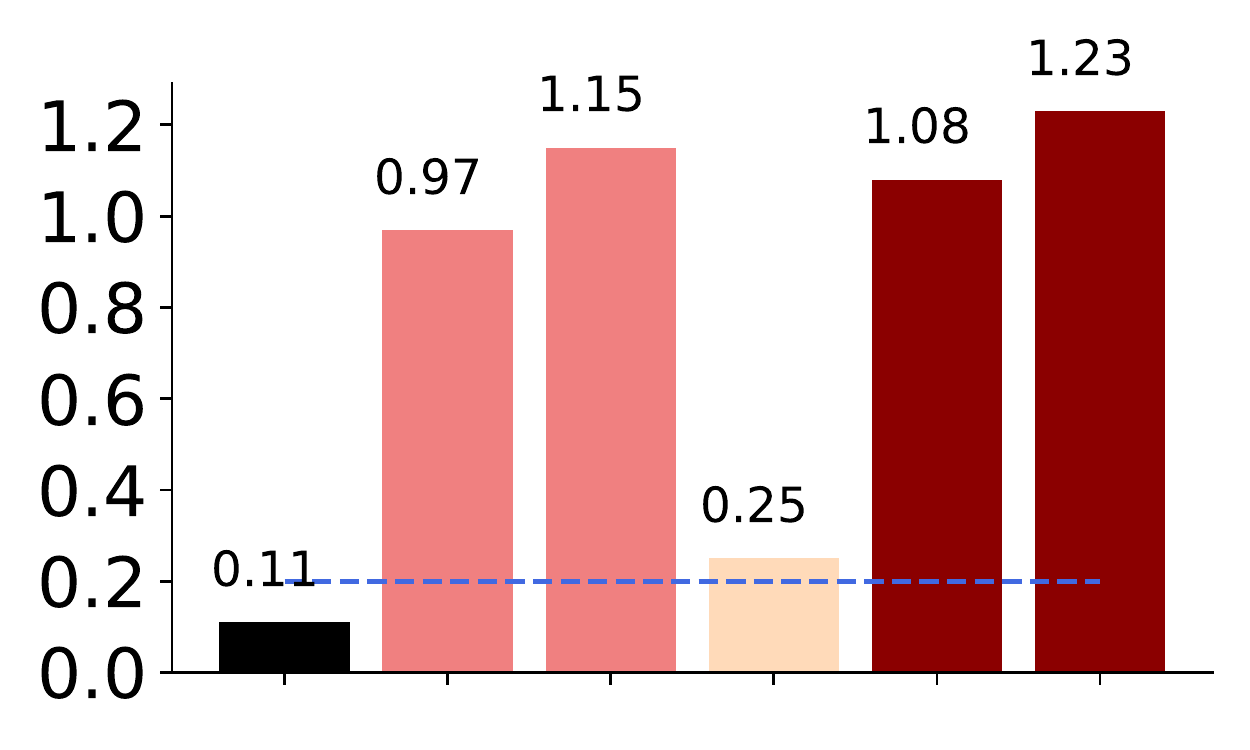} &
\includegraphics[width=\suppanalysisWidth]{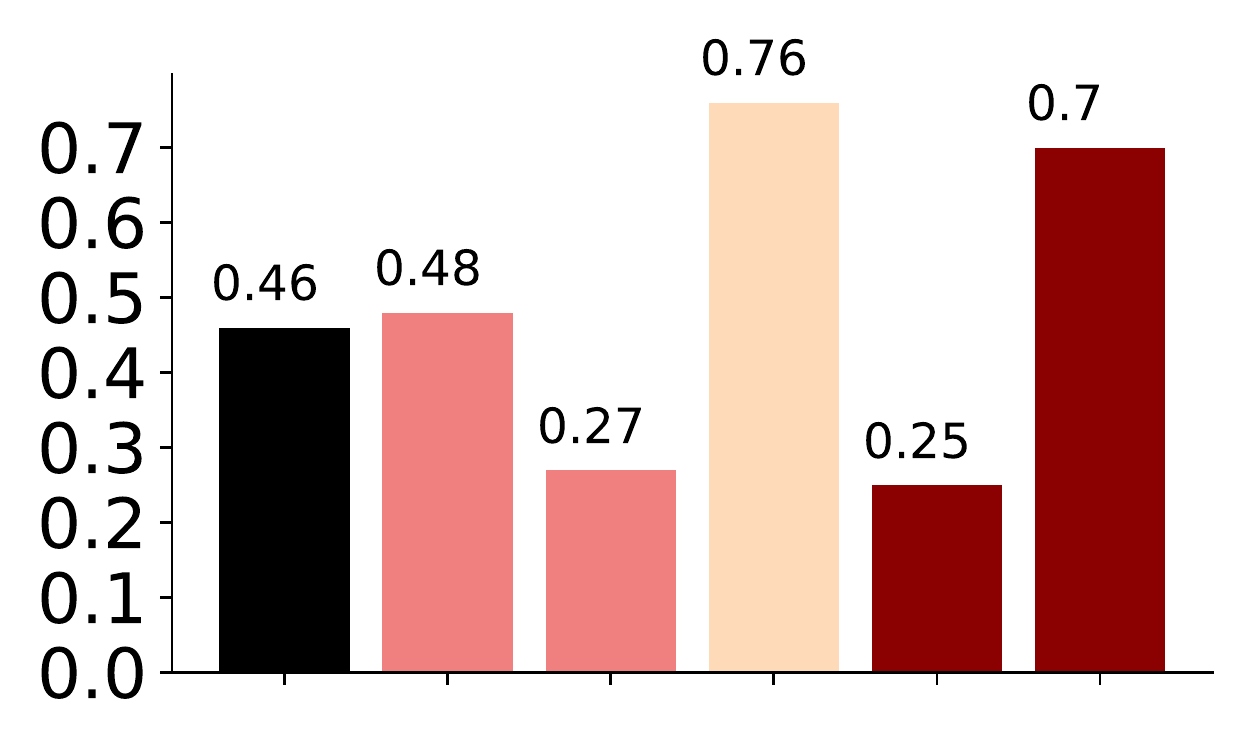}&
\includegraphics[width=\suppanalysisWidth]{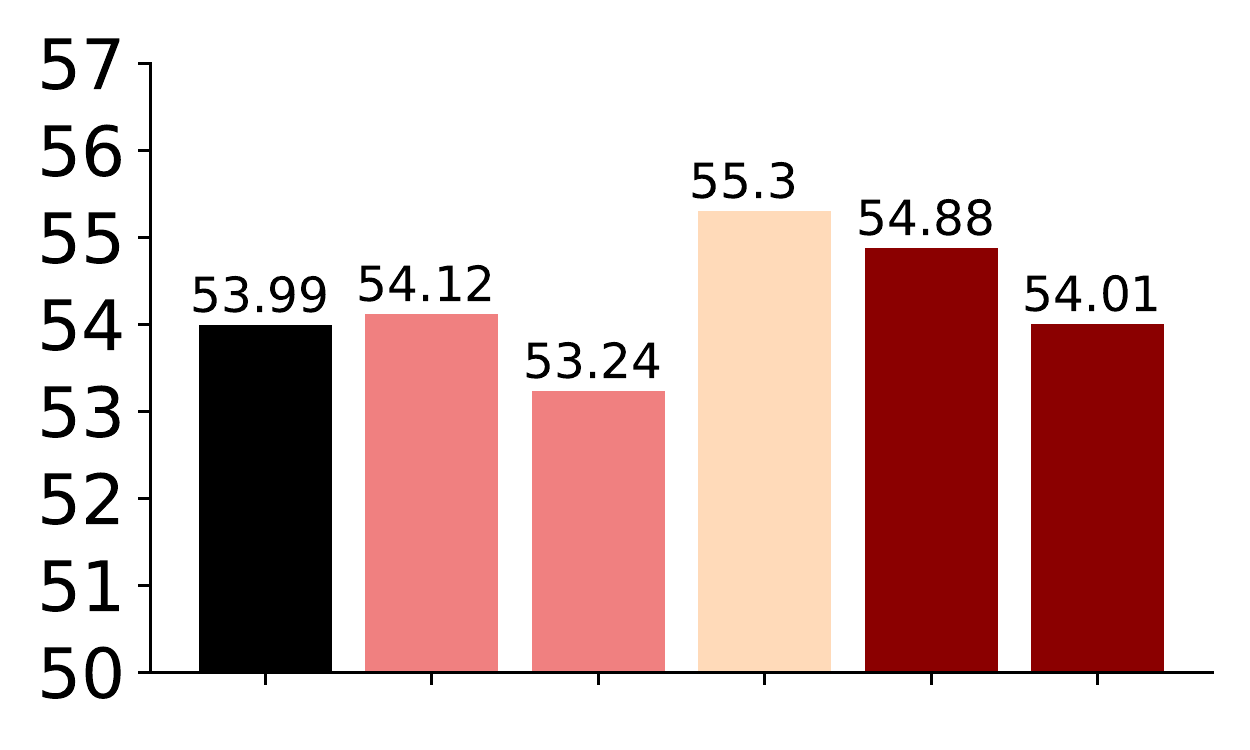}\\

\raisebox{1.5\normalbaselineskip}[0pt][0pt]{\rotatebox[origin=c]{90}{\tt{\scriptsize{PNav-\texttt{RGB}D}}}} & 
\includegraphics[width=\suppanalysisWidth]{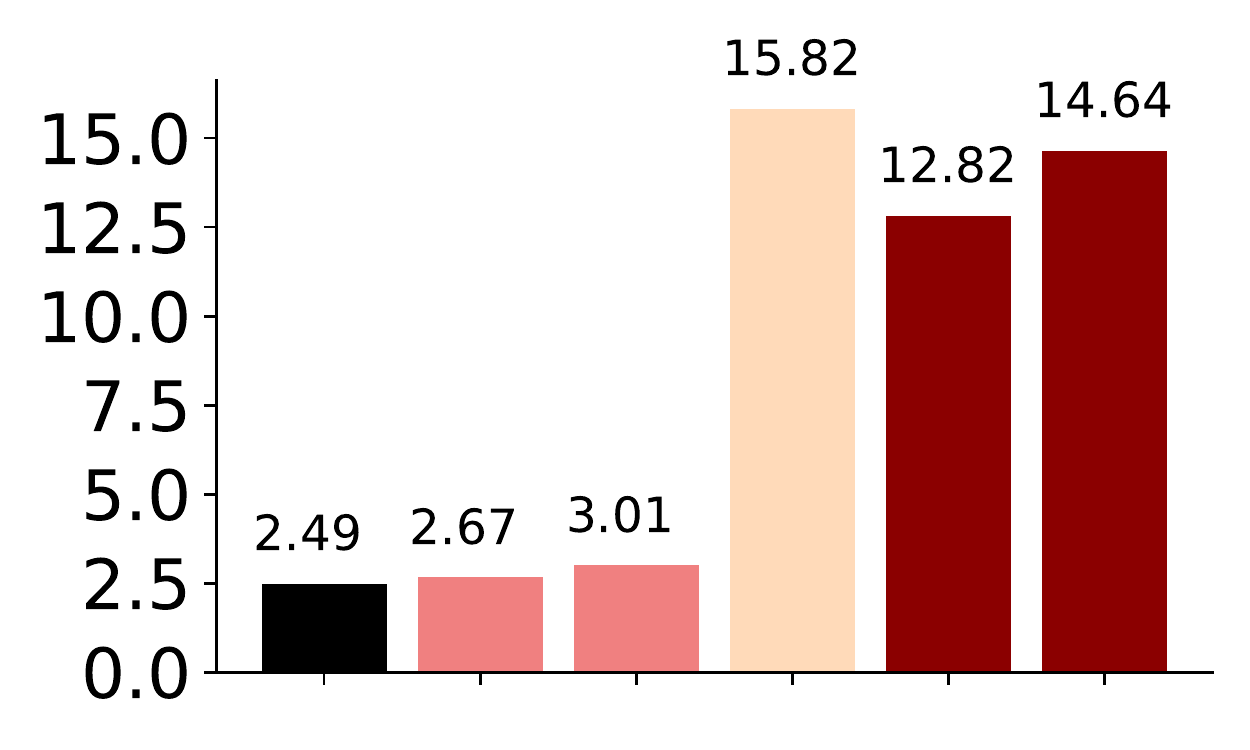}&
\includegraphics[width=\suppanalysisWidth]{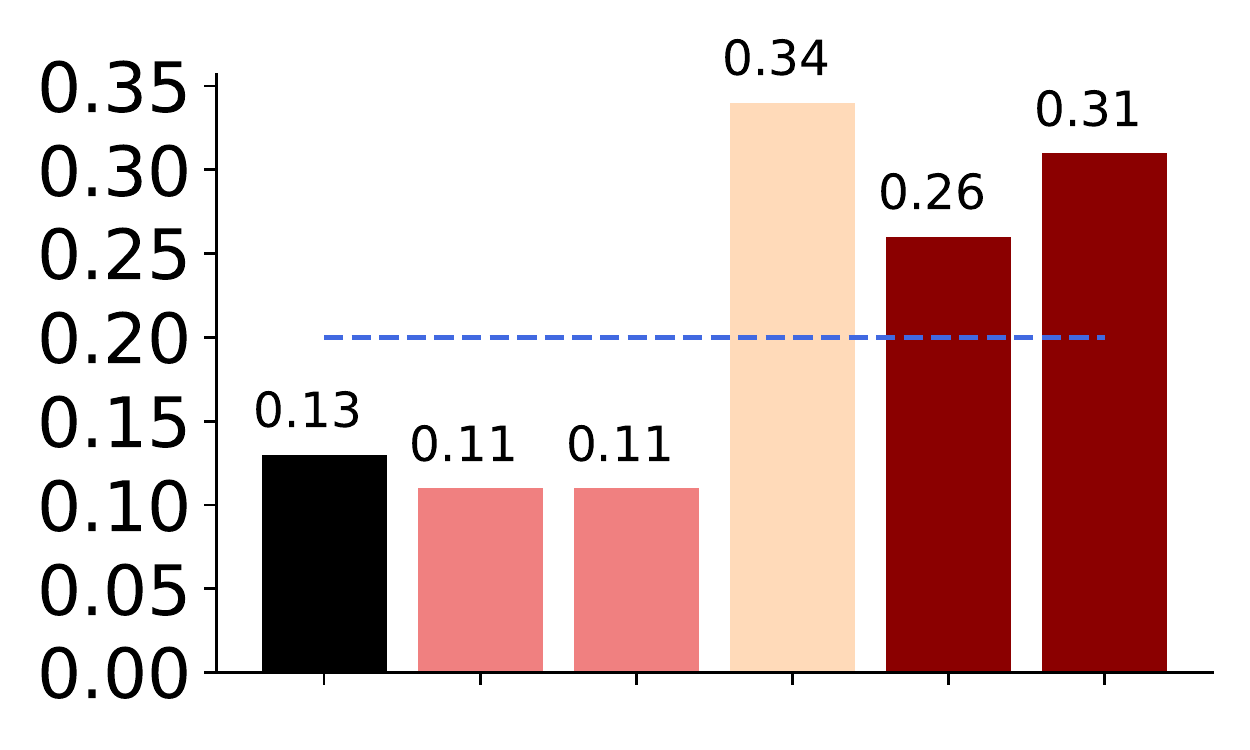} &
\includegraphics[width=\suppanalysisWidth]{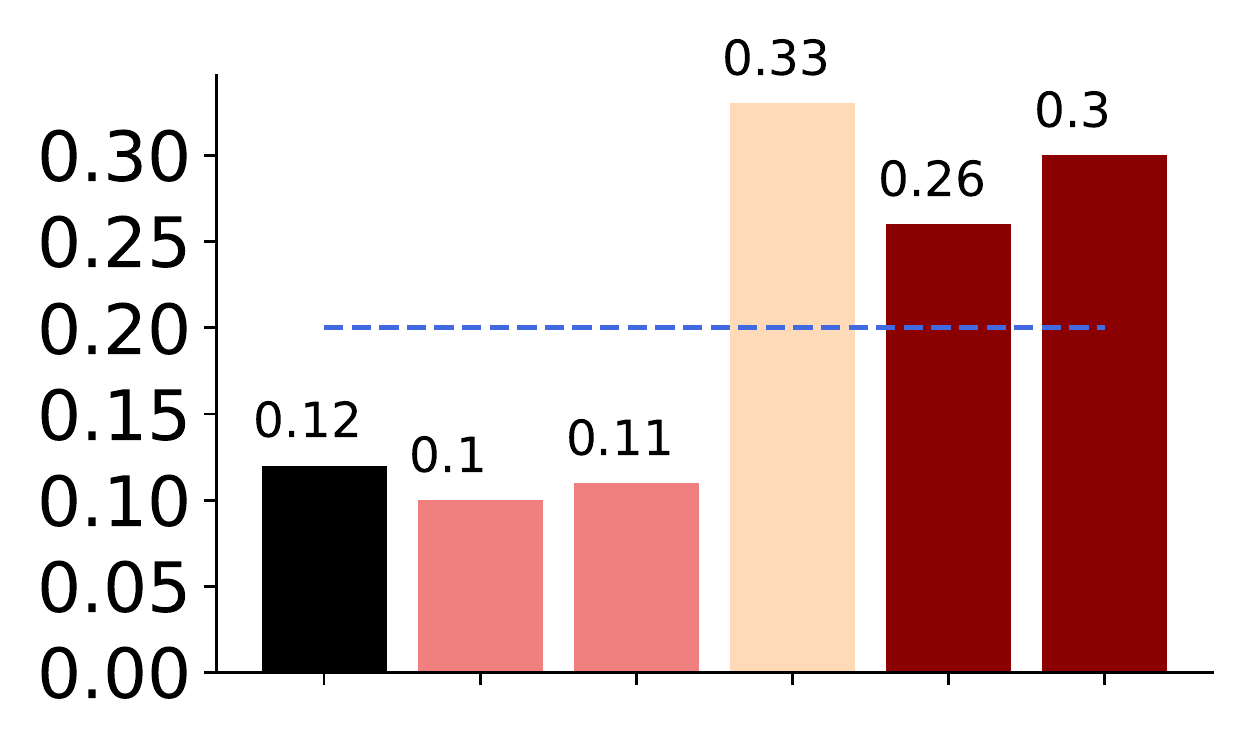} &
\includegraphics[width=\suppanalysisWidth]{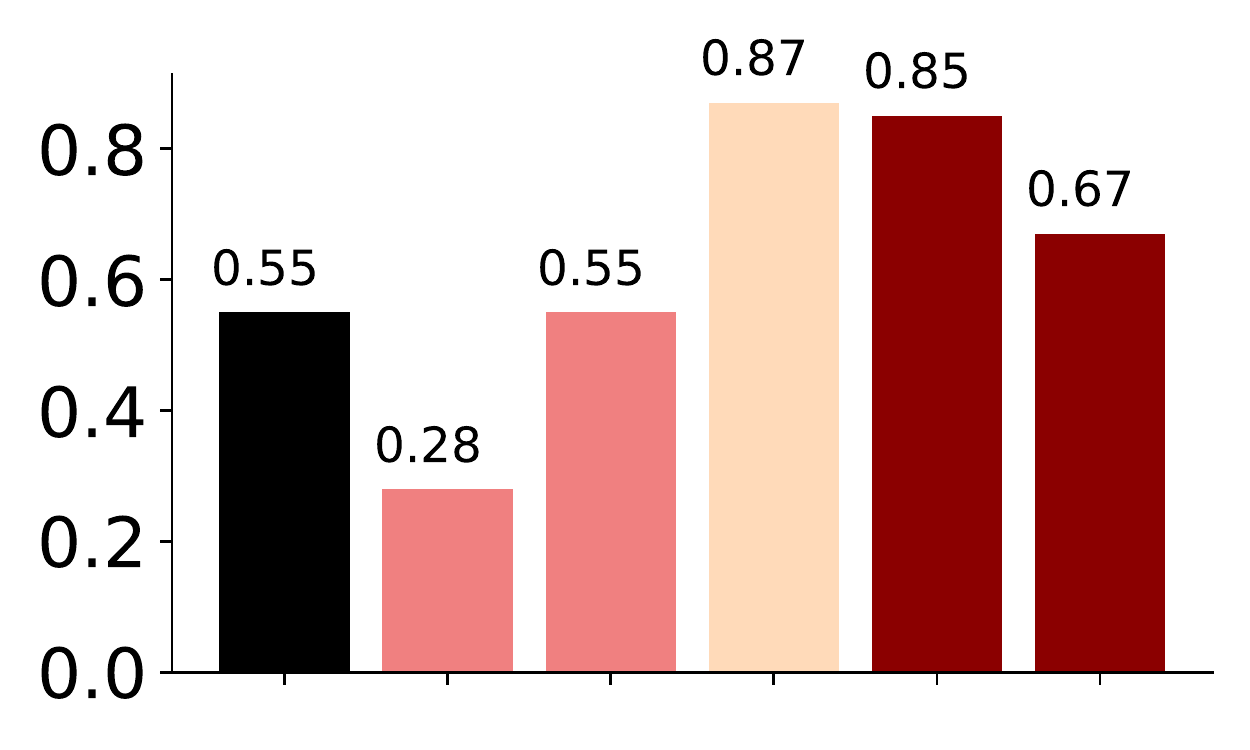}&
\includegraphics[width=\suppanalysisWidth]{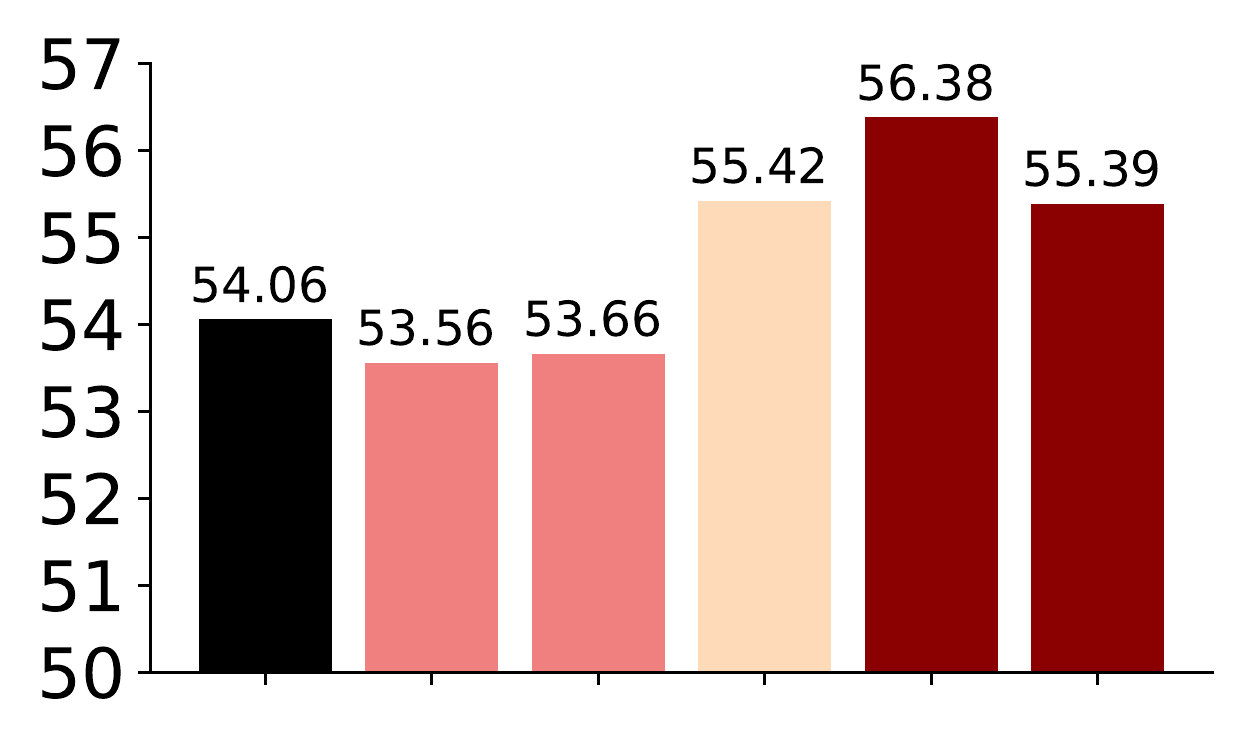}\\
\raisebox{1.5\normalbaselineskip}[0pt][0pt]{\rotatebox[origin=c]{90}{\tt{\scriptsize{ONav-\texttt{RGB}}}}} & 
\includegraphics[width=\suppanalysisWidth]{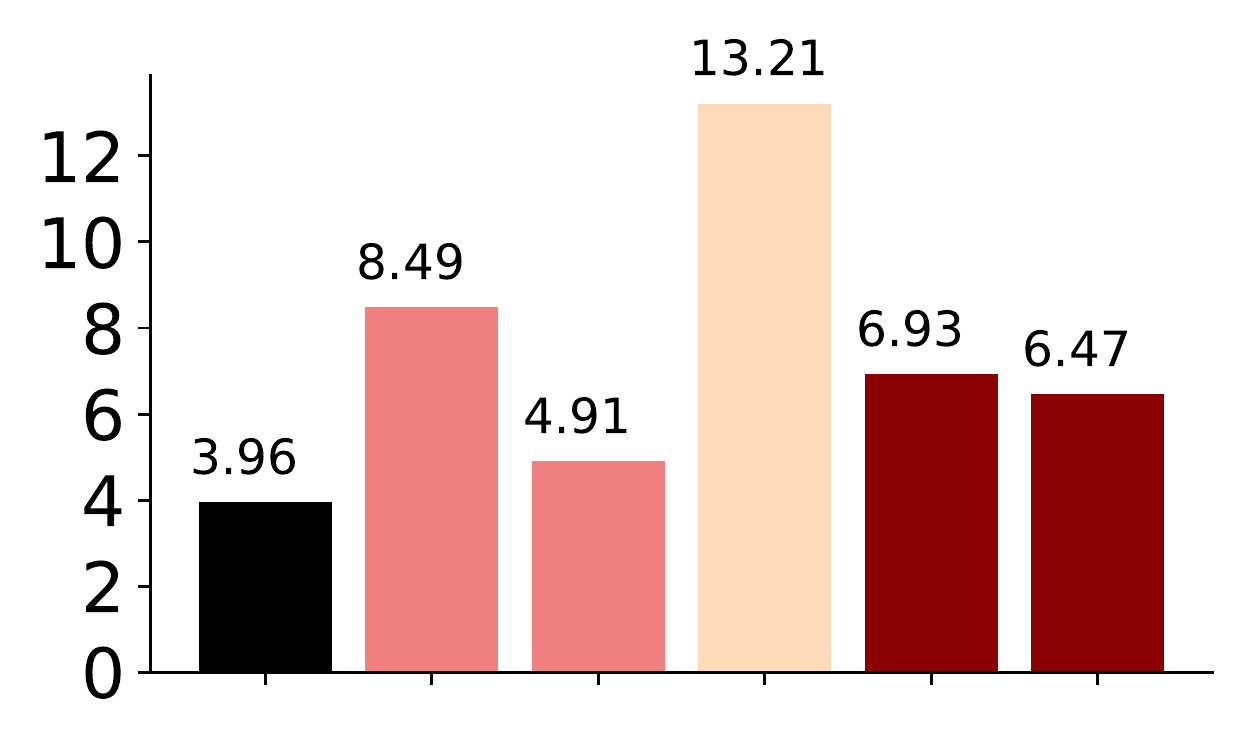}&
\includegraphics[width=\suppanalysisWidth]{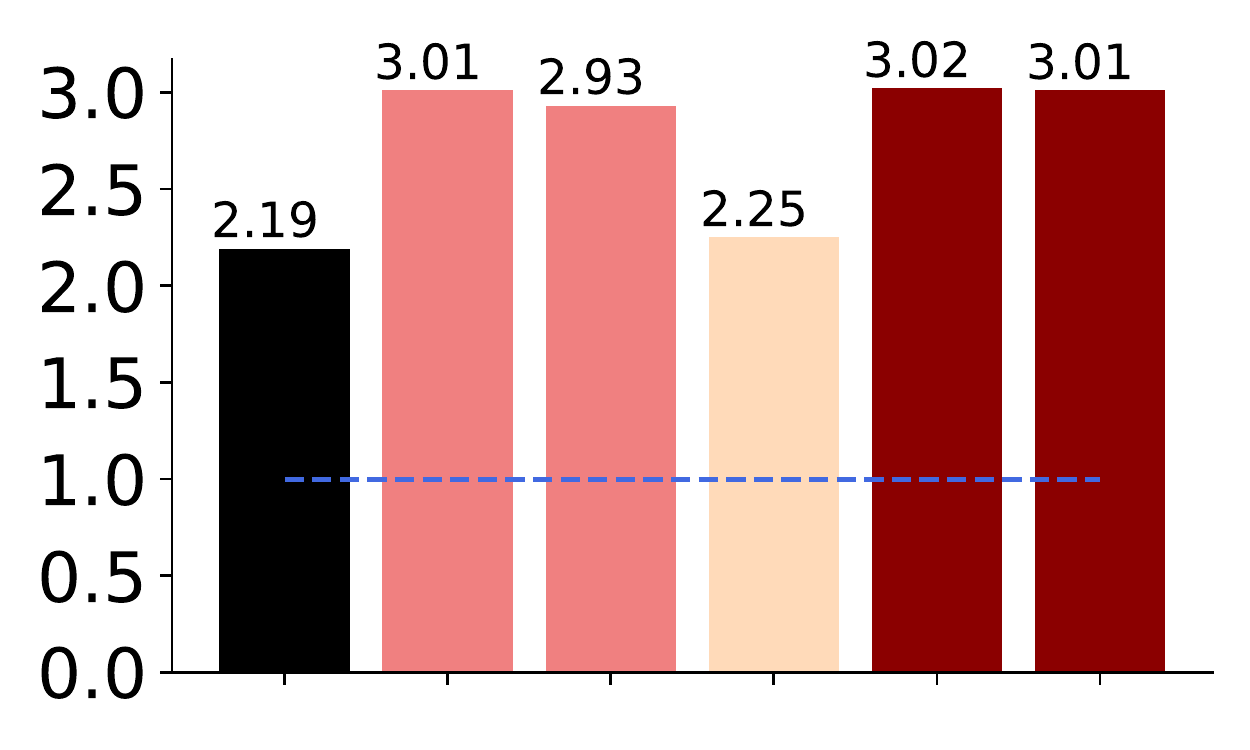}&
\includegraphics[width=\suppanalysisWidth]{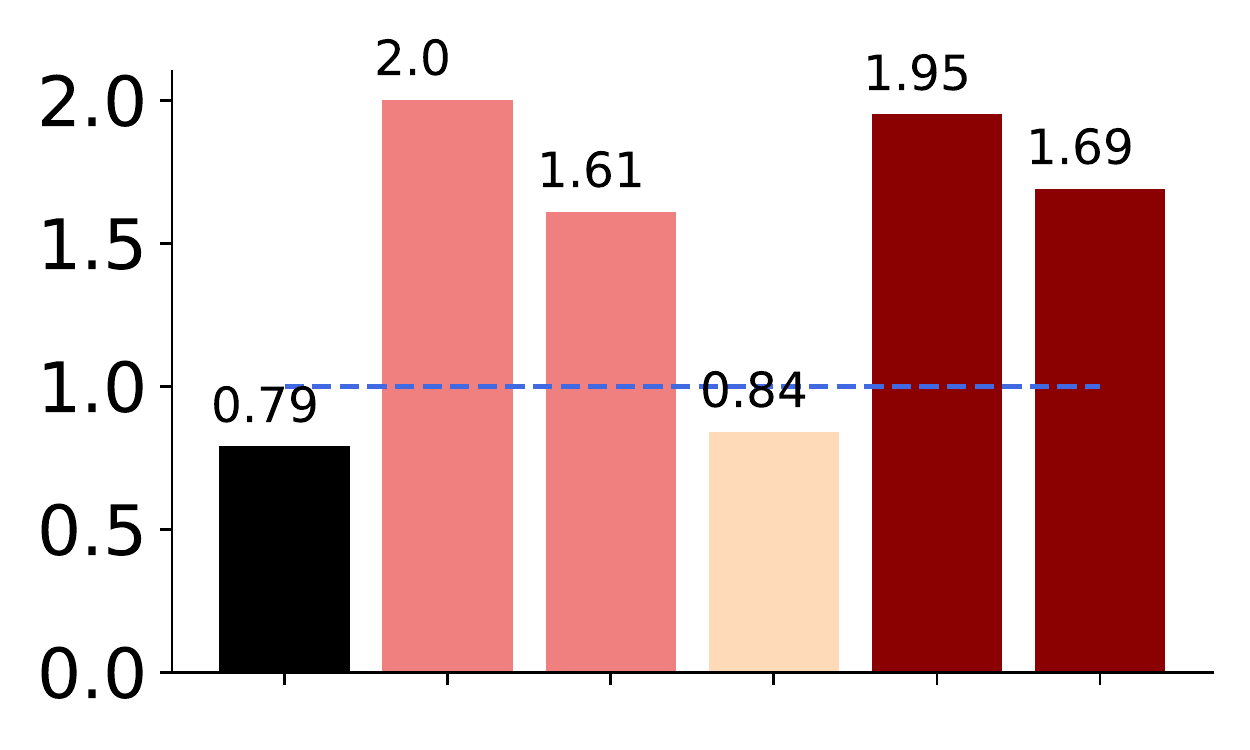}&
\includegraphics[width=\suppanalysisWidth]{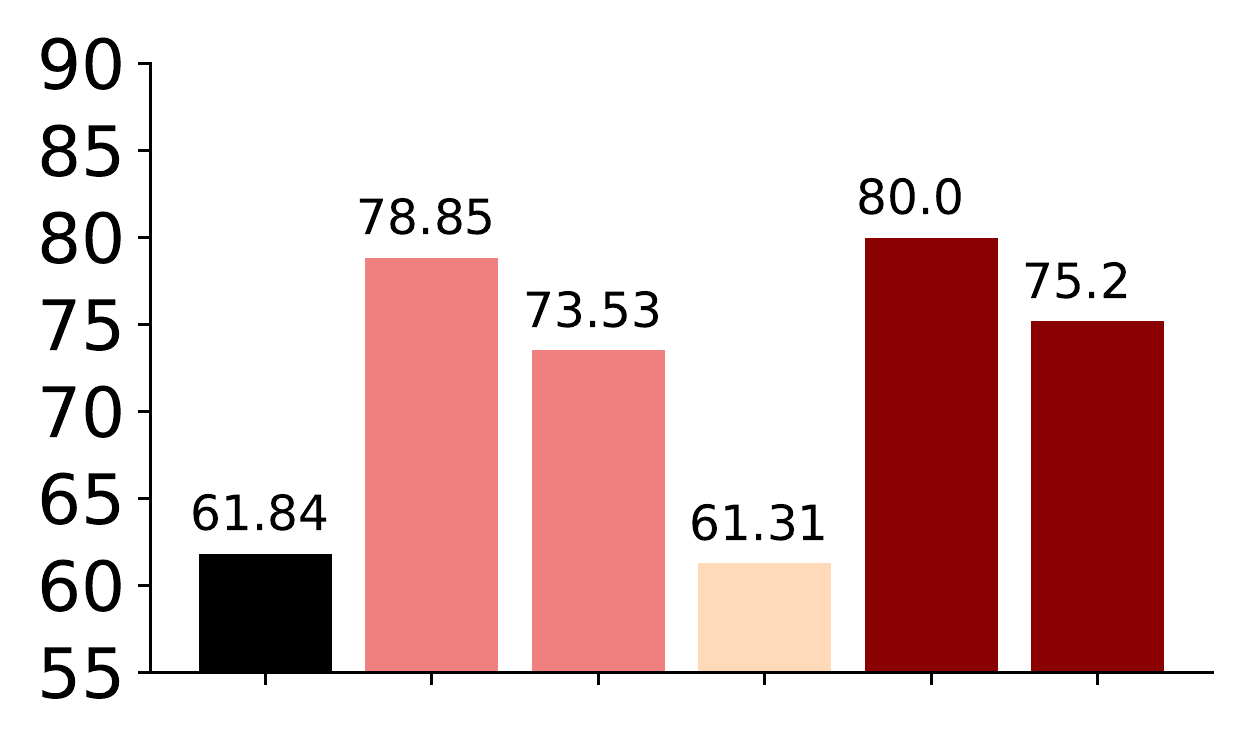}&
\includegraphics[width=\suppanalysisWidth]{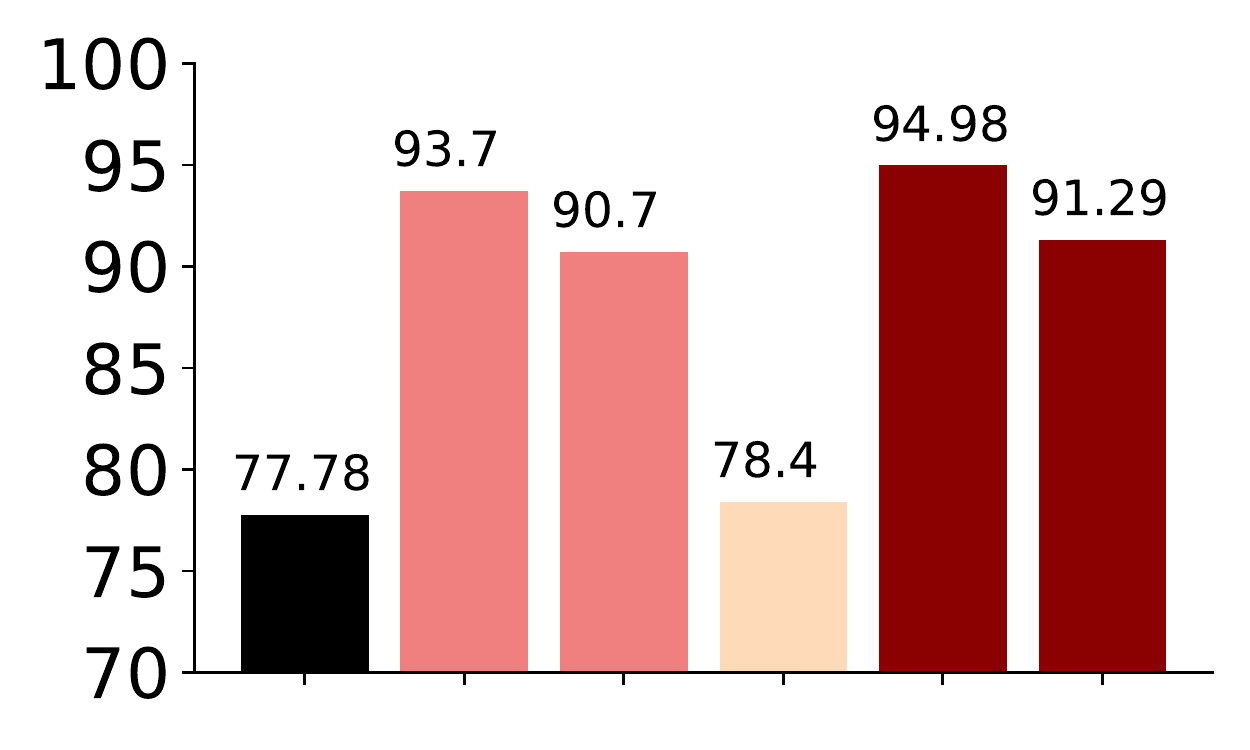}\\

\raisebox{3\normalbaselineskip}[0pt][0pt]{\rotatebox[origin=c]{90}{\tt{\scriptsize{ONav-\texttt{RGB}D}}}} & 
\includegraphics[width=\suppanalysisWidth]{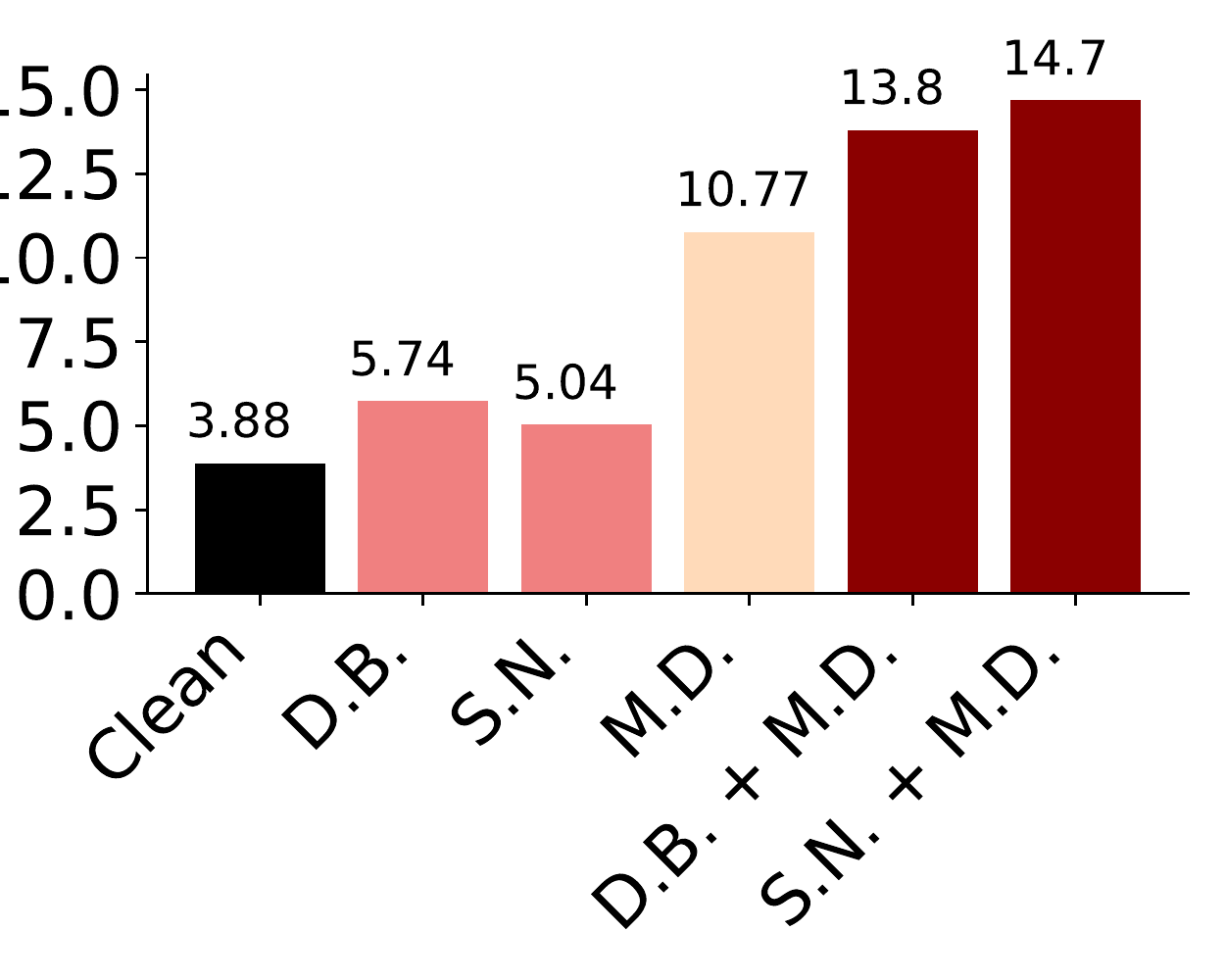}&
\includegraphics[width=\suppanalysisWidth]{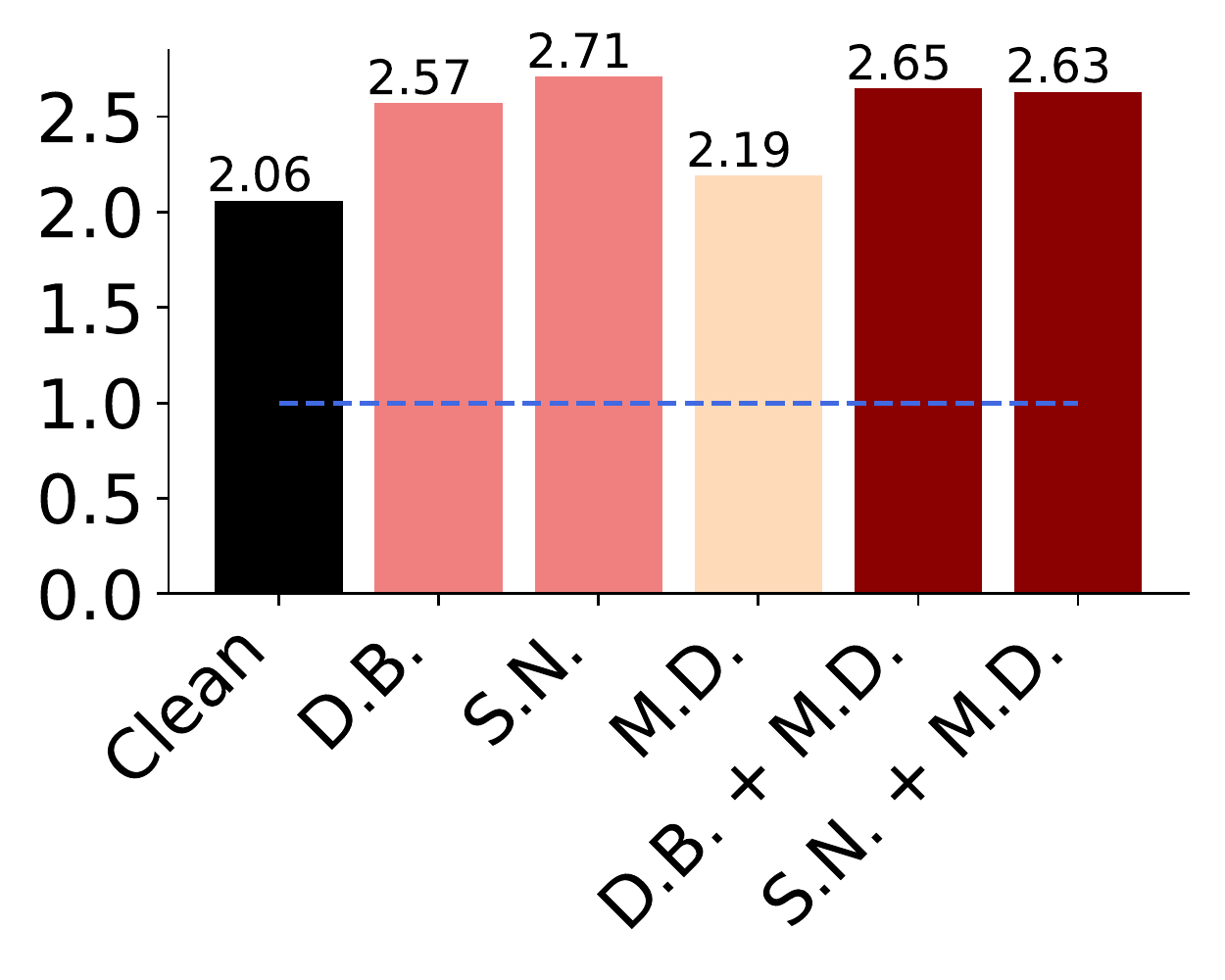}&
\includegraphics[width=\suppanalysisWidth]{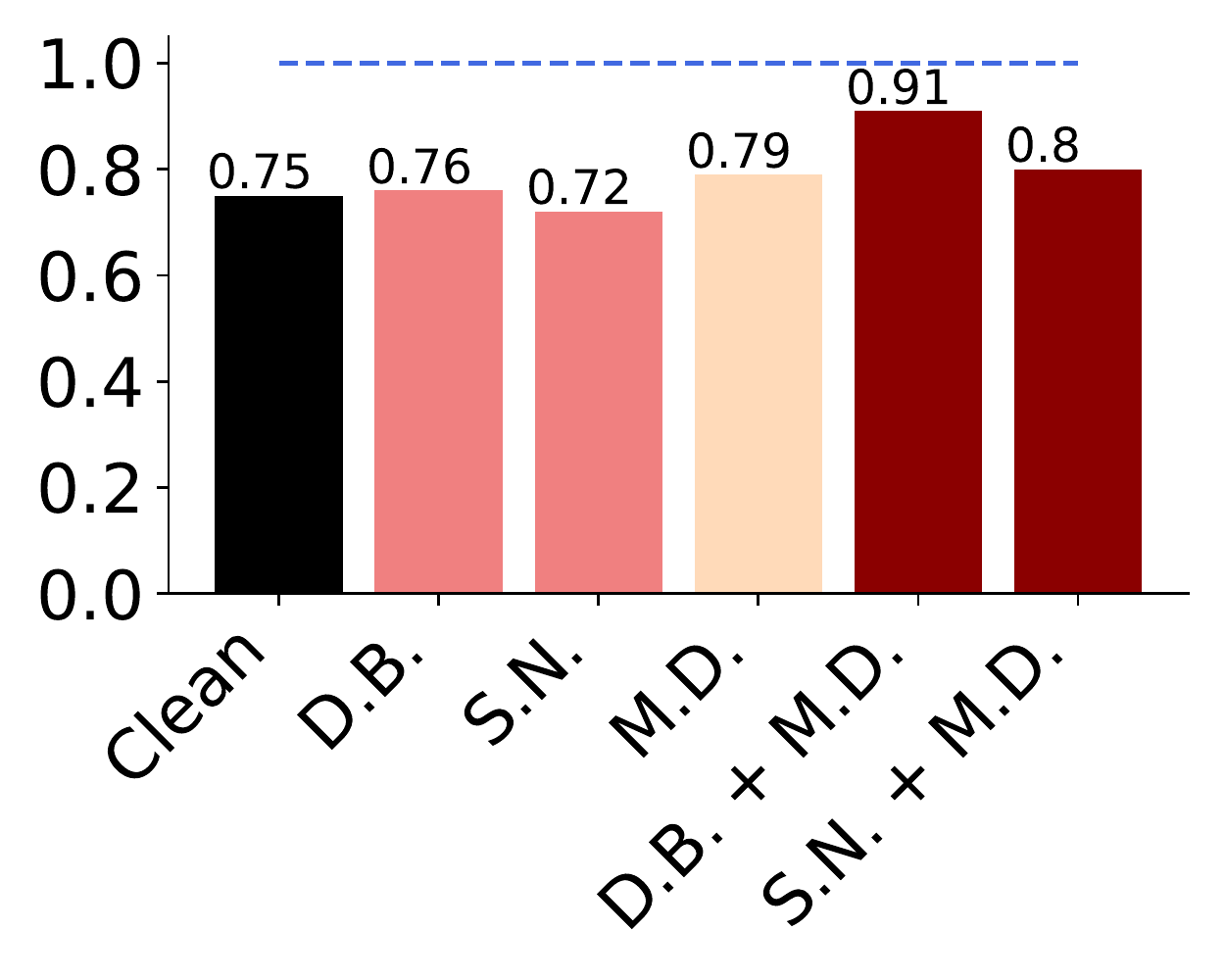}&
\includegraphics[width=\suppanalysisWidth]{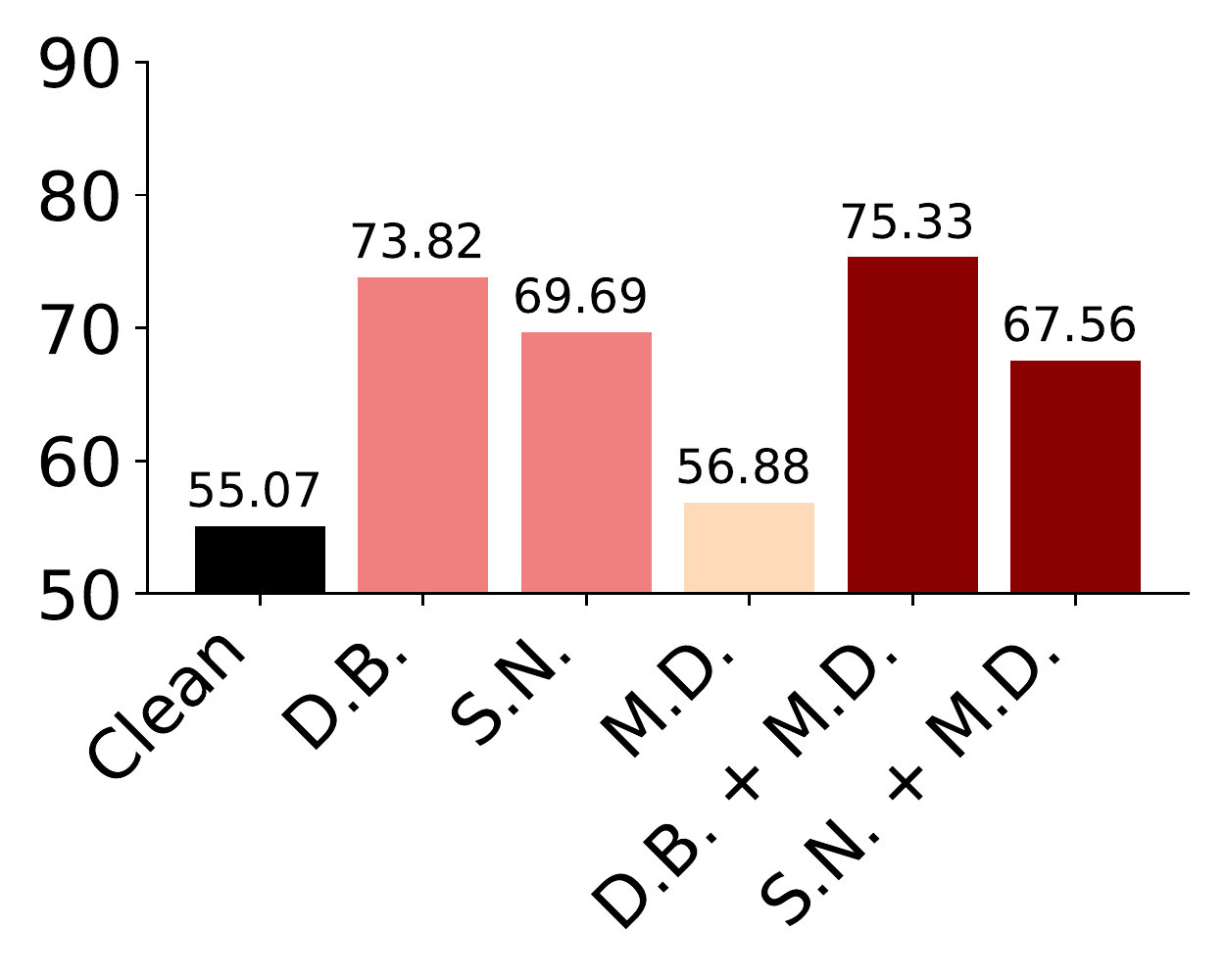}&
\includegraphics[width=\suppanalysisWidth]{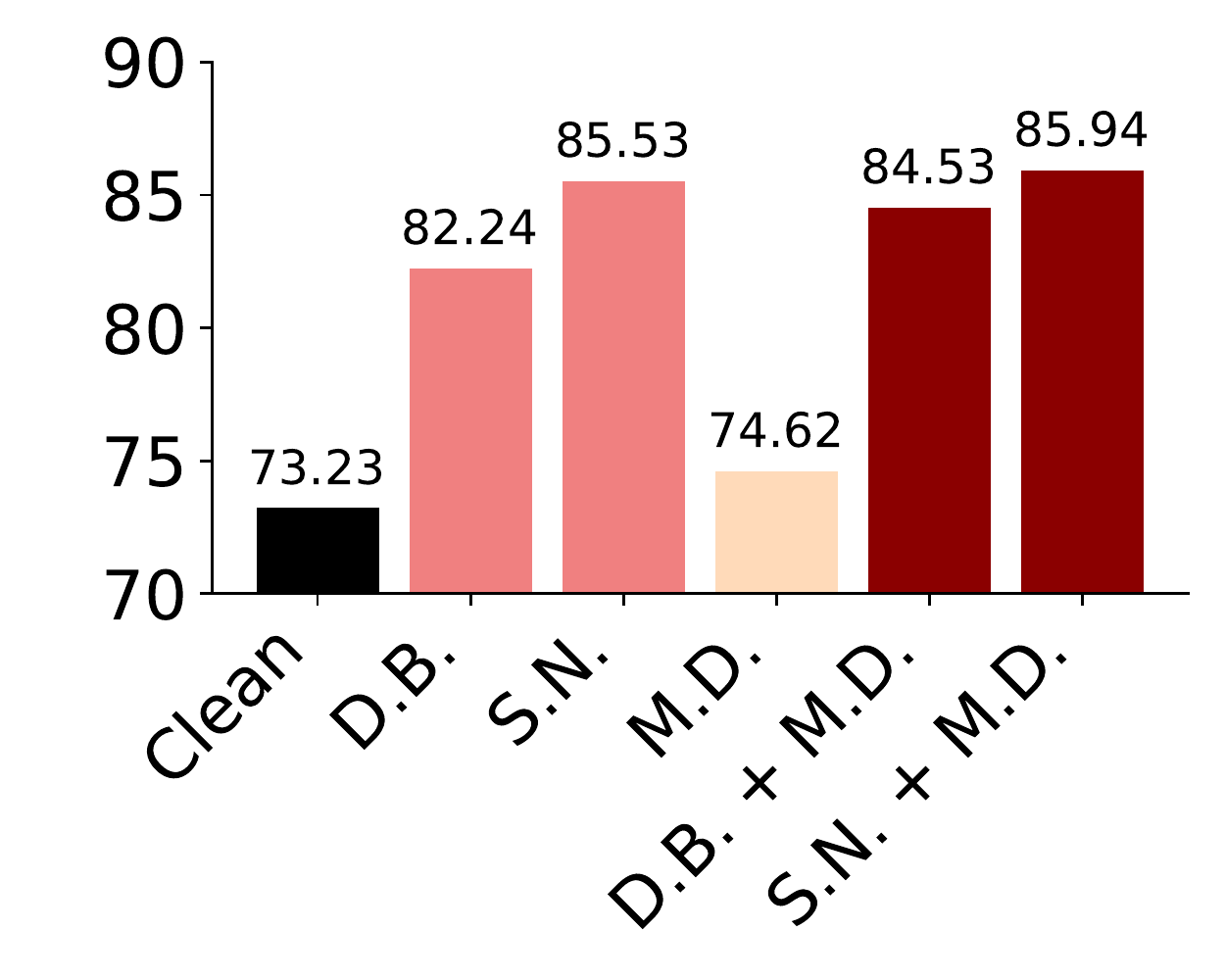}\\
\end{tabular}
\caption{\textbf{Agent Behavior Analysis.} To understand agent behaviors, we report the breakdown of four metrics: Number of collisions as observed through \texttt{\small Failed Action} (\textit{first column}), 
distance to target at episode termination as measured by \texttt{\small Term. Dist. to Target} (\textit{second column)},
closest agent was to target as measured by \texttt{\small Min. Dist. to Target} (\textit{third column}), and failure to appropriately end an episode either when out of range -- \texttt{\small Stop-Fail (Pos)} (\textit{fourth column}), or in range -- \texttt{\small Stop-Fail (Neg)} (\textit{fifth column}). Each behavior is reported for both \pnav (\texttt{RGB}-\textit{first row}, \texttt{RGB}D-\textit{second row}) and \onav (\texttt{RGB}-\textit{third row}, \texttt{RGB}D-\textit{fourth row}) within a clean and five corrupt settings: Defocus Blur (D.B.), Speckle Noise (S.N.), Motion Drift (M.D.), Defocus Blur + Motion Drift, and Speckle Noise + Motion Drift. \protect\inlinegraphics{figures/clean.pdf} is clean, 
\protect\inlinegraphics{figures/visual.pdf} is \vcr~ corruptions, \protect\inlinegraphics{figures/dynamics.pdf} is \dcr~ corruptions and \protect\inlinegraphics{figures/visdyn.pdf} is \vdcr~ corruptions. Blue lines in column 2 and 3 indicate the distance threshold for goal in range. Severities for S.N. and 
D.B. are set to $5$ (worst).}
\label{fig:supp_analysis_figure}
\vspace{-10pt}
\end{figure*}

\newcommand{\osrWidth}{.40\columnwidth}
\begin{figure}[ht!]
\centering
\begin{tabular}{l cc}
& \tt{\scriptsize{ONav-\texttt{RGB}}} & \tt{\scriptsize{ONav-\texttt{RGB}D}}\\
\raisebox{4.5\normalbaselineskip}[0pt][0pt]{\rotatebox[origin=c]{90}{\scriptsize{\texttt{\textbf{SR}$_{\text{Or}}$} - \texttt{\textbf{SR}}}}} &
\includegraphics[width=\osrWidth]{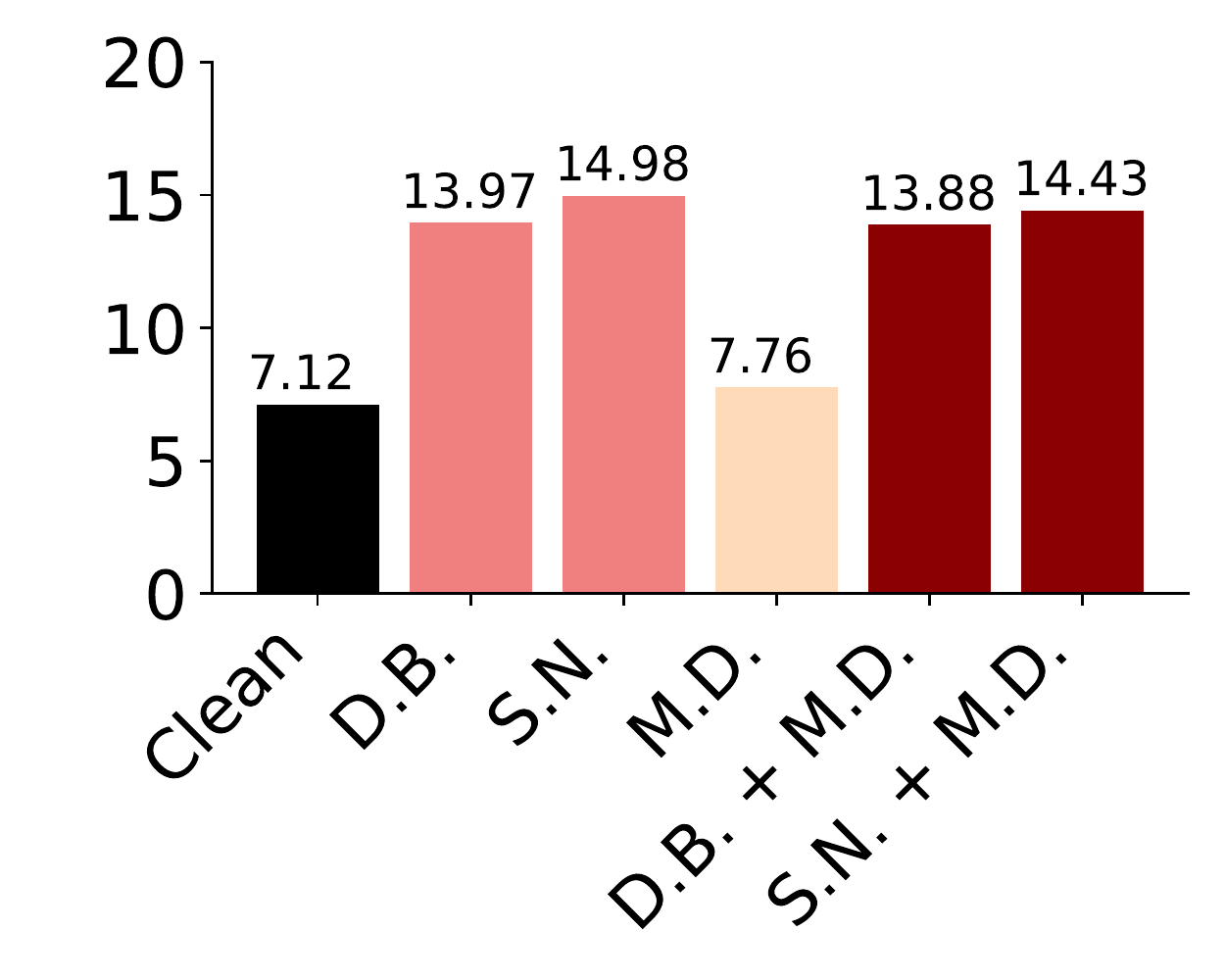}&
\includegraphics[width=\osrWidth]{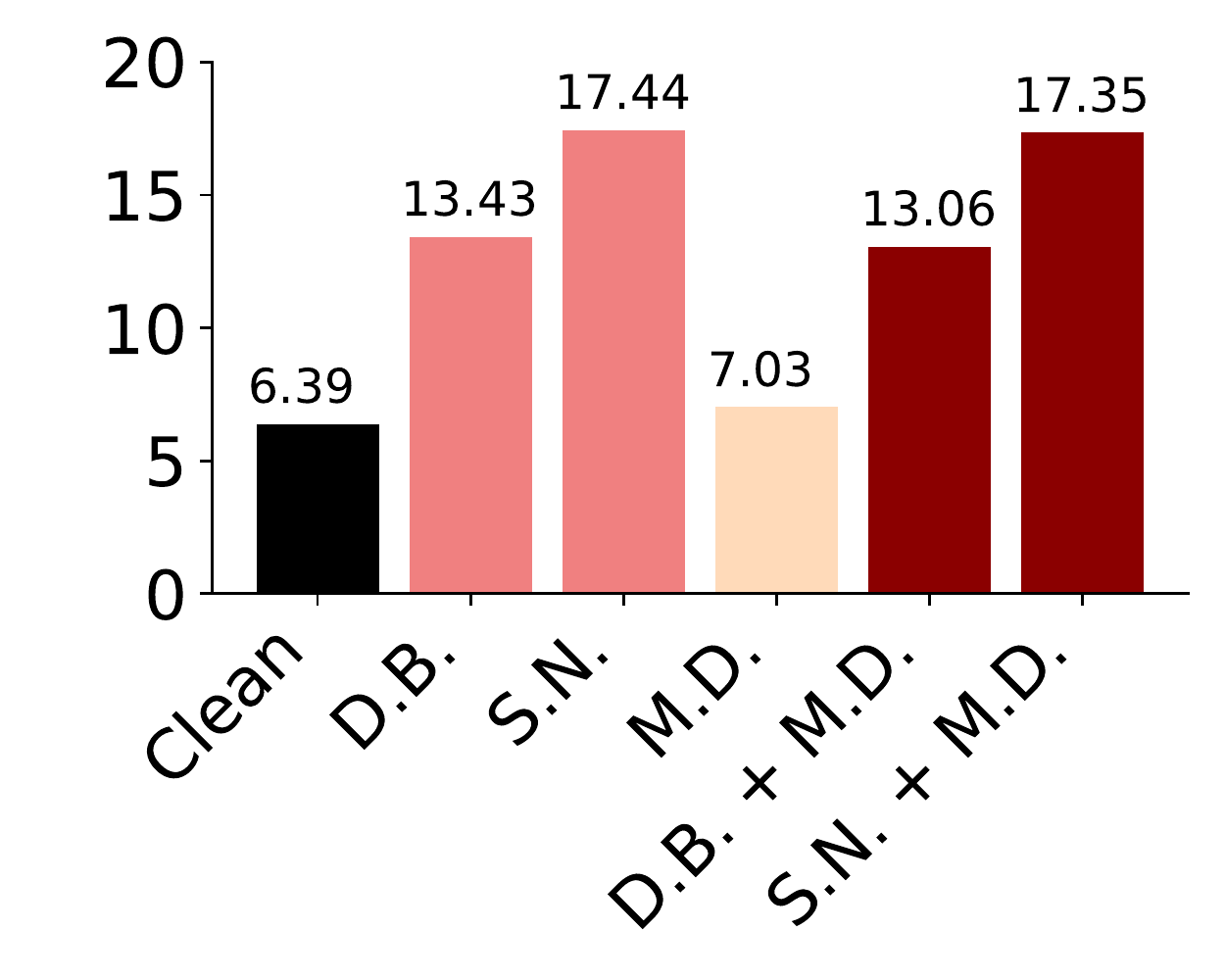}\\
\end{tabular}
\caption{\textbf{Effect of Degraded Stopping Mechanism.} To understand the extent to which
a degraded stopping mechanism under corruptions affects \onav \texttt{RGB} agent performance, we look
the difference between the agent's success rate (SR) compared to the setting where
the agent is equipped with an oracle stopping mechanism. SR$_{\text{Or}}$
denotes success rate when an \ed action is forcefully called in an episode whenever the goal is
in range. We consider one clean and five corrupt settings: Defocus Blur (D.B.), Speckle Noise (S.N.), Motion Drift (M.D.), Defocus Blur + Motion Drift, and Speckle Noise + Motion Drift. \protect\inlinegraphics{figures/clean.pdf} is clean, 
\protect\inlinegraphics{figures/visual.pdf} is \vcr~ corruptions, \protect\inlinegraphics{figures/dynamics.pdf} is \dcr~ corruptions and \protect\inlinegraphics{figures/visdyn.pdf} is \vdcr~ corruptions. Severities for S.N. and 
D.B. are set to $5$ (worst).}
\label{fig:oracle_sr_diff}
\vspace{-10pt}
\end{figure}

In Sec.~\ref{sec:behavior_analysis} of the main paper, we try to understand the idiosyncrasies
exhibited by the navigation agents under corruptions. Specifically, we look at the
number of collisions as observed through the number of failed actions in \rthor, the
closest the agent arrives to the target in an episode and Stop-Fail (Pos) and Stop-Fail
(Neg). Since for both \pnav and \onav, success depends on a notion of
``intentionality''~\cite{batra2020objectnav} -- the agent calls an \ed action when it
believes it has reached the goal --
we use both Stop-Fail (Pos) and Stop-Fail (Neg) to assess how corruptions impact this
``stopping'' mechanism of the agents. Stop-Fail (Pos) measures the fraction of times
the agent calls an \ed action when the goal is not in range\footnote{The goal in range criterion for \pnav checks whether the target is within the threshold
 distance. For \onav, this includes a visibility criterion
 in addition to taking distance into account.}, out of the number of times the agent
 calls an \ed action. Stop-Fail (Neg) measures the fraction of times the agent fails to
 invoke an \ed action when the goal is in range, out of the number of steps the goal
 is in range in an episode. Both are averaged across evaluation episodes. In addition to the above aspects, we also measure the average distance to the goal
 at episode termination. Here we report these measures for \pnav and \onav agents trained with
 \texttt{RGB} and \texttt{\texttt{RGB}-D} sensors in Fig.~\ref{fig:supp_analysis_figure}
 (\texttt{\texttt{RGB}-D} variants in addition to the \texttt{RGB} agents in Fig.4 of the main paper).
 
 We find that across \texttt{RGB} and \texttt{\texttt{RGB}-D} variants,
 (1) agents tend to collide more often under corruptions (Fig.~\ref{fig:supp_analysis_figure}, col 1),
 (2) agents generally end up farther from the target at episode
 termination under corruptions (Fig.~\ref{fig:supp_analysis_figure}, col 2) and (3) agents tend to be farther from the target under corruptions even
 in terms of minimum distance over an episode (Fig.~\ref{fig:supp_analysis_figure}, col 3). We further
 note that the effect of corruptions on the agent's
 stopping mechanism is more pronounced for \onav
 as opposed to \pnav (Fig.~\ref{fig:supp_analysis_figure}, cols 4 \& 5).
 
 \begin{table}[t!]
\footnotesize
\centering
\setlength{\tabcolsep}{3pt}
\resizebox{\columnwidth}{!}{
\begin{tabulary}{\columnwidth}{L  CCC CCC}
 \toprule
                & \textbf{\texttt{SR}}~$\uparrow$ & \textbf{\texttt{SPL}}~$\uparrow$ & Len. & \textbf{\texttt{SR}}~$\uparrow$ & \textbf{\texttt{SPL}}~$\uparrow$ & Len. \\
\midrule
 \textbf{Noise Free} & \multicolumn{3}{c}{\texttt{RGB}} & \multicolumn{3}{c}{\texttt{RGBD}}\\
  \midrule

\texttt{(1)} Clean  & 88.90 & 70.70 & 240.897 & 92.50 & 78.20 & 185.255\\
\texttt{(2)} Low-Light. & 63.30 & 32.90 & 566.286 & 91.90& 75.10& 207.994\\
\texttt{(3)} Spatter & 47.50 & 18.40 & 728.074 & 94.80 & 78.50 & 181.732\\
\midrule
 \textbf{HC Conditions} & \multicolumn{3}{c}{\texttt{RGB}} & \multicolumn{3}{c}{\texttt{RGBD}}\\
  \midrule
  \texttt{(4)} HC RGB Noise & N/A & N/A & N/A & 65.90 & 49.50 & 104.167\\
\texttt{(5)} Low-Light. & N/A & N/A & N/A & 60.50 & 45.50 & 107.932\\
\texttt{(6)} Spatter & N/A & N/A & N/A & 41.60 & 32.10 & 110.630\\
\bottomrule
\end{tabulary}
}
\caption{\textbf{OccAnt~\cite{ramakrishnan2020occant} 
results on Gibson~\cite{xia2018gibson}
val.}
Rows 1-3 are when \vcr\ corruptions are introduced over
clean settings under noise-free conditions, based on the checkpoint
used to report results in the publication.
Rows 4-6 are when RGB noise under Habitat Challenge (HC) conditions
is replaced with the \vcr\ corruptions, based on the HC submission
checkpoint.
Len indicates episode length.
N/A implies checkpoint not available.
Severity for Low-Lighting and Spatter is set to 5 (worst).
}
\label{tab:gibson_res}
\end{table}
 
 To further understand the
 extent to which a worse stopping mechanism impacts the agent's performance, in Fig.~\ref{fig:oracle_sr_diff}, we compare
 the agents' success rate (\texttt{SR}) with
a setting where the agent is equipped with an oracle stopping mechanism (forcefully call
\ed when goal is in range). For both \onav \texttt{RGB} and \texttt{\texttt{RGB}-D}, we find
that the presence of \vcr\ and \vdcr\ corruptions affects success significantly
compared to the clean settings (Fig.~\ref{fig:oracle_sr_diff}, black bars).

\subsection{Degradation Results}
\label{sec:deg_res}

\par \noindent
\textbf{Habitat Challenge Results.} As stated in Sec.5 of the main paper,
here we investigate the degree to which more sophisticated \pnav agents,
composed of map-based architectures, are susceptible to \vcr\
corruptions. Specifically, we evaluate the performance of the winning
entry of Habitat Challenge (HC) 2020~\cite{habitat-challenge} --
Occupancy Anticipation~\cite{ramakrishnan2020occant} on the
Gibson~\cite{xia2018gibson} validation scenes (see Table.~\ref{tab:gibson_res}). We evaluate the performance
of OccAnt (for \texttt{RGB} and \texttt{\texttt{RGB}-D}; based on provided checkpoints) when
\vcr\ corruptions are introduced (1) over clean settings under noise-free
conditions (rows 1-3 in Table.~\ref{tab:gibson_res}) and (2) by replacing
the \texttt{RGB} noise under Habitat Challenge (HC) conditions (rows 4-6 in
Table.~\ref{tab:gibson_res}). Under noise free conditions, we note that
degradation in performance from the clean settings is more pronounced
for the \texttt{RGB} agents as opposed to the \texttt{\texttt{RGB}-D} variants. Under HC 
conditions, we note that the \texttt{\texttt{RGB}-D} variants suffer significant degradation
in performance when \texttt{RGB} noise is replaced with \vcr\ corruptions.

\begin{table*}
    \setlength{\tabcolsep}{5pt}
    \centering
\resizebox{\textwidth}{!}{
\begin{tabular}{l c c c cc c cc c cc c cc c}
\toprule
    & & & && \multicolumn{5}{c}{\textbf{\pnav}} && \multicolumn{5}{c}{\textbf{\onav}} \\
    & & & && \multicolumn{2}{c}{\texttt{RGB}} && \multicolumn{2}{c}{\texttt{RGB-D}}  &&
             \multicolumn{2}{c}{\texttt{RGB}} && \multicolumn{2}{c}{\texttt{RGB-D}} \\
    \cmidrule{6-7} \cmidrule{9-10}
    \cmidrule{12-13} \cmidrule{15-16}
    \texttt{\#} & \textbf{Corruption}~$\downarrow$ &  V & D &&
        \textbf{\texttt{SR}}~$\uparrow$ &  \textbf{\texttt{SPL}}~$\uparrow$ &&   \textbf{\texttt{SR}}~$\uparrow$ &  \textbf{\texttt{SPL}}~$\uparrow$ &&
        \textbf{\texttt{SR}}~$\uparrow$ &  \textbf{\texttt{SPL}}~$\uparrow$ &&   \textbf{\texttt{SR}}~$\uparrow$ &  \textbf{\texttt{SPL}}~$\uparrow$ \\

   \midrule
   \band \texttt{1} & Clean &  &  && 98.97\tiny{$\pm$0.18} & 83.45\tiny{$\pm$0.27} && 99.24\tiny{$\pm$0.15} & 85.00\tiny{$\pm$0.25} && 31.78\tiny{$\pm$0.81} & 14.50\tiny{$\pm$0.47} && 35.43\tiny{$\pm$0.83} & 17.57\tiny{$\pm$0.52}\\
   \midrule
    \texttt{2} & Low Lighting (S3) &  \checkmark &  && 97.45\tiny{$\pm$0.27} & 80.53\tiny{$\pm$0.33} && 99.09\tiny{$\pm$0.17} & 84.91\tiny{$\pm$0.26} && 21.55\tiny{$\pm$0.72} &  8.91\tiny{$\pm$0.38} && 27.55\tiny{$\pm$0.78} & 13.08\tiny{$\pm$0.47}\\
    
    \texttt{3} & Low Lighting (S5) &  \checkmark &  && 93.42\tiny{$\pm$0.43} & 74.88\tiny{$\pm$0.43} && 99.27\tiny{$\pm$0.15} & 85.04\tiny{$\pm$0.25} && 11.69\tiny{$\pm$0.56} &  4.90\tiny{$\pm$0.30} && 23.26\tiny{$\pm$0.74} & 10.61\tiny{$\pm$0.43}\\
    
    \texttt{4} & Motion Blur (S3) &  \checkmark &  && 98.82\tiny{$\pm$0.19} & 80.64\tiny{$\pm$0.29} && 98.91\tiny{$\pm$0.18} & 84.62\tiny{$\pm$0.26} && 18.57\tiny{$\pm$0.68} &  8.18\tiny{$\pm$0.37} && 24.32\tiny{$\pm$0.75} & 11.52\tiny{$\pm$0.44}\\
    
    \texttt{5} & Motion Blur (S5) &  \checkmark &  && 96.15\tiny{$\pm$0.34} & 73.24\tiny{$\pm$0.38} && 99.06\tiny{$\pm$0.17} & 85.01\tiny{$\pm$0.25} && 10.50\tiny{$\pm$0.93} &  4.71\tiny{$\pm$0.51} && 18.26\tiny{$\pm$0.83} &  7.87\tiny{$\pm$0.45}\\
    
    \texttt{6} & Camera Crack &  \checkmark &  && 81.56\tiny{$\pm$0.68} & 63.48\tiny{$\pm$0.59} && 96.00\tiny{$\pm$0.34} & 81.48\tiny{$\pm$0.36} && 7.06\tiny{$\pm$0.45} &  3.54\tiny{$\pm$0.28} && 27.28\tiny{$\pm$0.78} & 13.49\tiny{$\pm$0.48}\\
    
    \texttt{7} & Defocus Blur (S3) &  \checkmark &  && 94.63\tiny{$\pm$0.39} & 73.28\tiny{$\pm$0.41} && 98.79\tiny{$\pm$0.19} & 84.47\tiny{$\pm$0.26} && 15.59\tiny{$\pm$0.63} &  6.99\tiny{$\pm$0.36} && 22.07\tiny{$\pm$0.72} &  9.75\tiny{$\pm$0.41}\\
    
    \texttt{8} & Defocus Blur (S5) &  \checkmark &  && 75.83\tiny{$\pm$0.75} & 53.48\tiny{$\pm$0.59} && 99.03\tiny{$\pm$0.17} & 85.44\tiny{$\pm$0.25} && 4.20\tiny{$\pm$0.35} &  1.81\tiny{$\pm$0.20} && 17.66\tiny{$\pm$0.67} &  7.47\tiny{$\pm$0.36}\\
    
    \texttt{9} & Speckle Noise (S3) &  \checkmark &  && 89.23\tiny{$\pm$0.54} & 68.18\tiny{$\pm$0.53} && 98.85\tiny{$\pm$0.19} & 84.58\tiny{$\pm$0.27} && 14.92\tiny{$\pm$0.62} &  6.54\tiny{$\pm$0.34} && 24.05\tiny{$\pm$0.75} & 10.34\tiny{$\pm$0.42}\\
    
    \texttt{10} & Speckle Noise (S5) &  \checkmark &  && 66.70\tiny{$\pm$0.82} & 47.92\tiny{$\pm$0.67} && 98.97\tiny{$\pm$0.18} & 84.79\tiny{$\pm$0.26} && 8.68\tiny{$\pm$0.49} &  3.90\tiny{$\pm$0.28} && 17.69\tiny{$\pm$0.67} &  7.42\tiny{$\pm$0.36}\\
    
    \texttt{11} & Lower-FOV &  \checkmark &  && 43.25\tiny{$\pm$0.86} & 32.39\tiny{$\pm$0.68} && 89.44\tiny{$\pm$0.54} & 73.92\tiny{$\pm$0.50} && 10.17\tiny{$\pm$0.53}&2.50\tiny{$\pm$0.19} && 10.02\tiny{$\pm$0.52} &  4.89\tiny{$\pm$0.31}\\
    
    \texttt{12} & Spatter (S3) &  \checkmark &  && 38.40\tiny{$\pm$0.85} & 25.53\tiny{$\pm$0.59} && 98.64\tiny{$\pm$0.20} & 84.00\tiny{$\pm$0.28} && 7.06\tiny{$\pm$0.45} &  3.81\tiny{$\pm$0.29} && 23.20\tiny{$\pm$0.74} &  9.71\tiny{$\pm$0.40}\\
    
    \texttt{13} & Spatter (S5) &  \checkmark &  && 34.64\tiny{$\pm$0.83} & 25.55\tiny{$\pm$0.64} && 99.30\tiny{$\pm$0.14} & 84.68\tiny{$\pm$0.25} && 7.79\tiny{$\pm$0.47} &  2.71\tiny{$\pm$0.22} && 21.43\tiny{$\pm$0.72} &  9.98\tiny{$\pm$0.42}\\

   \midrule 
    \texttt{14} & Motion Bias (S) &   & \checkmark && 95.69\tiny{$\pm$0.35} & 77.05\tiny{$\pm$0.38} && 96.60\tiny{$\pm$0.32} & 79.22\tiny{$\pm$0.35} && 33.21\tiny{$\pm$0.82} & 14.88\tiny{$\pm$0.47} && 34.98\tiny{$\pm$0.83} & 16.70\tiny{$\pm$0.51}\\
    
    \texttt{15} & Motion Drift &   & \checkmark && 95.94\tiny{$\pm$0.34} & 76.32\tiny{$\pm$0.35} && 93.57\tiny{$\pm$0.43} & 75.09\tiny{$\pm$0.40} && 28.55\tiny{$\pm$0.79} & 13.30\tiny{$\pm$0.46} && 34.37\tiny{$\pm$0.83} & 16.43\tiny{$\pm$0.50}\\
    
    \texttt{16} & Motion Bias (C) &   & \checkmark && 92.27\tiny{$\pm$0.47} & 77.48\tiny{$\pm$0.46} && 93.11\tiny{$\pm$0.44} & 79.04\tiny{$\pm$0.45} && 30.47\tiny{$\pm$0.80} & 13.20\tiny{$\pm$0.45} && 32.72\tiny{$\pm$0.82} & 15.67\tiny{$\pm$0.50}\\
    
    \texttt{17} & PyRobot~\cite{murali2019pyrobot} (ILQR) Mul. = 1.0 &   & \checkmark && 95.18\tiny{$\pm$0.37} & 67.45\tiny{$\pm$0.37} && 96.18\tiny{$\pm$0.33} & 69.48\tiny{$\pm$0.35} && 32.54\tiny{$\pm$0.82} & 11.65\tiny{$\pm$0.39} && 36.86\tiny{$\pm$0.84} & 14.24\tiny{$\pm$0.44}\\
    
    \texttt{18} & Motor Failure &   & \checkmark && 20.84\tiny{$\pm$0.71} & 17.91\tiny{$\pm$0.62} && 21.41\tiny{$\pm$0.71} & 18.39\tiny{$\pm$0.62} && 4.60\tiny{$\pm$0.37} &  2.88\tiny{$\pm$0.26} && 6.06\tiny{$\pm$0.42} &  3.65\tiny{$\pm$0.28}\\

   \midrule 
    \texttt{19} & Defocus Blur (S3) + Motion Bias (S) &  \checkmark & \checkmark && 92.72\tiny{$\pm$0.45} & 68.61\tiny{$\pm$0.43} && 97.45\tiny{$\pm$0.27} & 79.70\tiny{$\pm$0.32} && 14.40\tiny{$\pm$0.61} &  6.15\tiny{$\pm$0.33} && 22.40\tiny{$\pm$0.73} &  9.20\tiny{$\pm$0.39}\\
    
    \texttt{20} & Defocus Blur (S5) + Motion Bias (S) &  \checkmark & \checkmark && 75.80\tiny{$\pm$0.75} & 50.76\tiny{$\pm$0.58} && 97.00\tiny{$\pm$0.30} & 79.81\tiny{$\pm$0.33} && 5.66\tiny{$\pm$0.40} &  2.34\tiny{$\pm$0.22} && 17.53\tiny{$\pm$0.66} &  7.07\tiny{$\pm$0.35}\\
    
    \texttt{21} & Speckle Noise (S3) + Motion Bias (S) &  \checkmark & \checkmark && 86.62\tiny{$\pm$0.59} & 63.20\tiny{$\pm$0.54} && 96.85\tiny{$\pm$0.30} & 79.23\tiny{$\pm$0.34} && 14.92\tiny{$\pm$0.62} &  6.31\tiny{$\pm$0.34} && 24.72\tiny{$\pm$0.75} & 10.04\tiny{$\pm$0.41}\\
    
    \texttt{22} & Speckle Noise (S5) + Motion Bias (S) &  \checkmark & \checkmark && 64.36\tiny{$\pm$0.83} & 44.38\tiny{$\pm$0.66} && 96.78\tiny{$\pm$0.31} & 79.49\tiny{$\pm$0.34} && 8.95\tiny{$\pm$0.50} &  3.85\tiny{$\pm$0.27} && 18.39\tiny{$\pm$0.68} &  7.49\tiny{$\pm$0.36}\\
    
    \texttt{23} & Spatter (S3) + Motion Bias (S) &  \checkmark & \checkmark && 37.25\tiny{$\pm$0.84} & 23.83\tiny{$\pm$0.57} && 96.60\tiny{$\pm$0.32} & 78.62\tiny{$\pm$0.35} && 7.18\tiny{$\pm$0.45} &  3.60\tiny{$\pm$0.28} && 24.44\tiny{$\pm$0.75} &  9.80\tiny{$\pm$0.40}\\
    
    \texttt{24} & Spatter (S5) + Motion Bias (S) &  \checkmark & \checkmark && 33.85\tiny{$\pm$0.82} & 23.98\tiny{$\pm$0.61} && 95.94\tiny{$\pm$0.34} & 78.64\tiny{$\pm$0.36} && 7.64\tiny{$\pm$0.46} &  2.93\tiny{$\pm$0.23} && 20.91\tiny{$\pm$0.71} &  9.39\tiny{$\pm$0.40}\\

    \midrule
    \texttt{25} & Defocus Blur (S3) + Motion Drift &  \checkmark & \checkmark && 89.72\tiny{$\pm$0.53} & 65.84\tiny{$\pm$0.47} && 94.84\tiny{$\pm$0.39} & 75.97\tiny{$\pm$0.37} && 14.16\tiny{$\pm$0.61} &  6.26\tiny{$\pm$0.34} && 23.56\tiny{$\pm$0.74} & 10.65\tiny{$\pm$0.43}\\
    
    \texttt{26} & Defocus Blur (S5) + Motion Drift &  \checkmark & \checkmark && 73.92\tiny{$\pm$0.76} & 50.84\tiny{$\pm$0.59} && 94.72\tiny{$\pm$0.39} & 76.21\tiny{$\pm$0.37} && 4.57\tiny{$\pm$0.36} &  2.10\tiny{$\pm$0.21} && 17.26\tiny{$\pm$0.66} &  7.04\tiny{$\pm$0.35}\\
    
    \texttt{27} & Speckle Noise (S3) + Motion Drift &  \checkmark & \checkmark && 86.65\tiny{$\pm$0.59} & 62.44\tiny{$\pm$0.53} && 93.99\tiny{$\pm$0.41} & 75.02\tiny{$\pm$0.39} && 13.46\tiny{$\pm$0.60} &  5.95\tiny{$\pm$0.33} && 23.01\tiny{$\pm$0.73} &  9.96\tiny{$\pm$0.41}\\
    
    \texttt{28} & Speckle Noise (S5) + Motion Drift &  \checkmark & \checkmark && 63.18\tiny{$\pm$0.84} & 43.29\tiny{$\pm$0.65} && 94.51\tiny{$\pm$0.40} & 75.34\tiny{$\pm$0.38} && 7.49\tiny{$\pm$0.80} &  3.63\tiny{$\pm$0.46} && 18.93\tiny{$\pm$0.68} &  7.85\tiny{$\pm$0.36}\\
    
    \texttt{29} & Spatter (S3) + Motion Drift &  \checkmark & \checkmark && 37.70\tiny{$\pm$0.84} & 24.27\tiny{$\pm$0.57} && 94.57\tiny{$\pm$0.39} & 75.34\tiny{$\pm$0.38} && 7.15\tiny{$\pm$0.45} &  3.59\tiny{$\pm$0.27} && 23.44\tiny{$\pm$0.74} &  9.72\tiny{$\pm$0.40}\\
    
    \texttt{30} & Spatter (S5) + Motion Drift &  \checkmark & \checkmark && 33.36\tiny{$\pm$0.82} & 23.59\tiny{$\pm$0.60} && 95.03\tiny{$\pm$0.38} & 75.84\tiny{$\pm$0.37} && 7.21\tiny{$\pm$0.55} & 2.77\tiny{$\pm$0.28} && 18.69\tiny{$\pm$0.68} &  8.37\tiny{$\pm$0.38}\\

    \midrule
    \texttt{31} & Defocus Blur (S3) + PyRobot~\cite{murali2019pyrobot} (ILQR) Mul. = 1.0 &  \checkmark & \checkmark && 93.99\tiny{$\pm$0.41} & 58.88\tiny{$\pm$0.40} && 97.66\tiny{$\pm$0.26} & 70.54\tiny{$\pm$0.32} && 16.13\tiny{$\pm$0.64} &  5.22\tiny{$\pm$0.28} && 22.68\tiny{$\pm$0.73} &  7.33\tiny{$\pm$0.32}\\
    
    \texttt{32} & Defocus Blur (S5) + PyRobot~\cite{murali2019pyrobot} (ILQR) Mul. = 1.0 &  \checkmark & \checkmark && 79.34\tiny{$\pm$0.71} & 42.29\tiny{$\pm$0.49} && 97.24\tiny{$\pm$0.29} & 70.35\tiny{$\pm$0.33} && 5.81\tiny{$\pm$0.41} &  1.04\tiny{$\pm$0.11} && 18.48\tiny{$\pm$0.68} &  5.86\tiny{$\pm$0.29}\\
    
    \texttt{33} & Speckle Noise (S3) + PyRobot~\cite{murali2019pyrobot} (ILQR) Mul. = 1.0 &  \checkmark & \checkmark && 88.38\tiny{$\pm$0.56} & 54.60\tiny{$\pm$0.49} && 96.12\tiny{$\pm$0.34} & 68.67\tiny{$\pm$0.35} && 14.95\tiny{$\pm$0.62} &  4.71\tiny{$\pm$0.26} && 24.11\tiny{$\pm$0.75} &  7.51\tiny{$\pm$0.32}\\
    
    \texttt{34} & Speckle Noise (S5) + PyRobot~\cite{murali2019pyrobot} (ILQR) Mul. = 1.0 &  \checkmark & \checkmark && 67.12\tiny{$\pm$0.82} & 37.77\tiny{$\pm$0.57} && 96.36\tiny{$\pm$0.33} & 69.44\tiny{$\pm$0.34} && 8.89\tiny{$\pm$0.50} &  2.66\tiny{$\pm$0.20} && 18.72\tiny{$\pm$0.68} &  5.73\tiny{$\pm$0.29}\\
    
    \texttt{35} & Spatter (S3) + PyRobot~\cite{murali2019pyrobot} (ILQR) Mul. = 1.0 &  \checkmark & \checkmark && 40.70\tiny{$\pm$0.86} & 18.26\tiny{$\pm$0.45} && 96.09\tiny{$\pm$0.34} & 68.25\tiny{$\pm$0.36} && 8.31\tiny{$\pm$0.48} &  1.76\tiny{$\pm$0.16} && 23.17\tiny{$\pm$0.74} &  7.76\tiny{$\pm$0.33}\\
    
    \texttt{36} & Spatter (S5) + PyRobot~\cite{murali2019pyrobot} (ILQR) Mul. = 1.0 &  \checkmark & \checkmark && 36.37\tiny{$\pm$0.84} & 19.70\tiny{$\pm$0.51} && 96.03\tiny{$\pm$0.34} & 68.98\tiny{$\pm$0.36} && 8.58\tiny{$\pm$0.49} &  2.09\tiny{$\pm$0.17} && 20.85\tiny{$\pm$0.71} &  7.41\tiny{$\pm$0.33}\\
    
    \bottomrule 
\end{tabular}
}
\caption{\textbf{\textsc{PointNav} and \textsc{ObjectNav} Performance.} Degradation in task
  performance of pretrained \pnav (trained for $\sim75$M frames) and \onav (trained for $\sim300$M frames) agents when evaluated
  under \vcr\ and \dcr\ corruptions present in \rnav. 
  \pnav agents have additional access to a GPS-Compass sensor.
  For visual corruptions with controllable
  severity levels, we report results with severity set to 5 and 3. Performance
  is measured across tasks of varying difficulties (easy, medium and hard).
  Reported results are mean and standard error across 3 evaluation runs with different seeds.
  Rows are sorted based on SPL values for RGB \pnav agents. Success and SPL values
  are reported as percentages.
   (V = Visual, D = Dynamics)
}
	\label{tab:pnav_onav_base_results_seed}
	\vspace{-17pt}
\end{table*}

\par \noindent
\textbf{More Degradation Results.} In Table.~\ref{tab:pnav_onav_base_results_seed},
we report the degradation in performance (relative to clean settings) of
\pnav and \onav agents when operating under \vcr, \dcr\ and \vdcr\ corruptions. We report
mean and standard error values across $3$ evaluation runs under actuation noise (wherever
applicable). For \vcr\ corruptions with controllable severity levels -- Motion Blur,
Low-Lighting, Defocus Blur, Speckle Noise and Spatter -- we report results
with severities set to $3$ and $5$ (identified by S3 and S5; excluded from the
main paper due to space constraints) -- for both \vcr\ and \vdcr\ settings.
We note that unlike the \texttt{\texttt{RGB}-D} variants, for \pnav \texttt{RGB} agents,
performance 
drops more as severity levels increase
(increasing degradation from severity $3\rightarrow$5).
For \onav, we find that for both \texttt{RGB} and \texttt{\texttt{RGB}-D} variants, performance decreases
as with increasing severity of corruptions ($3\rightarrow5$).

\begin{table*}[ht!]
\footnotesize
\centering
\setlength{\tabcolsep}{7pt}
\resizebox{\linewidth}{!}{
\begin{tabulary}{\linewidth}{L  CC  CC  CC}
 \toprule
 \textbf{Increasing Episode Difficulty $\rightarrow$} & \multicolumn{2}{c}{\textbf{Easy}} & \multicolumn{2}{c}{\textbf{Medium}} & \multicolumn{2}{c}{\textbf{Hard}}\\
\textbf{Corruption $\downarrow$} 
  &\texttt{\textbf{SR}}$\uparrow$ & \texttt{\textbf{SPL}}$\uparrow$ &  \texttt{\textbf{SR}}$\uparrow$ & \texttt{\textbf{SPL}}$\uparrow$ &  \texttt{\textbf{SR}}$\uparrow$ & \texttt{\textbf{SPL}}$\uparrow$\\
  \midrule
  \textbf{\pnav-RGB} & & & & & & \\
 \midrule
 \band \texttt{1} Clean&  99.64\tiny{$\pm$0.18} & 82.80\tiny{$\pm$0.38} & 99.36\tiny{$\pm$0.24} & 84.21\tiny{$\pm$0.47} & 97.91\tiny{$\pm$0.43} & 83.34\tiny{$\pm$0.54} \\
 \midrule
 \texttt{2} Low Lighting &  99.36\tiny{$\pm$0.24} & 80.59\tiny{$\pm$0.45} & 95.54\tiny{$\pm$0.62} & 75.83\tiny{$\pm$0.70} & 85.34\tiny{$\pm$1.07} & 68.22\tiny{$\pm$0.94} \\
 \texttt{3} Camera Crack &  94.10\tiny{$\pm$0.71} & 75.81\tiny{$\pm$0.70} & 80.05\tiny{$\pm$1.21} & 62.15\tiny{$\pm$1.06} & 70.49\tiny{$\pm$1.38} & 52.44\tiny{$\pm$1.14} \\
 \texttt{4} Spatter &  74.93\tiny{$\pm$1.31} & 57.85\tiny{$\pm$1.08} & 18.67\tiny{$\pm$1.18} & 12.03\tiny{$\pm$0.80} & 10.20\tiny{$\pm$0.91} &  6.69\tiny{$\pm$0.63} \\
 \midrule
 \texttt{5} Speckle Noise + Motion Bias (S) & 86.74\tiny{$\pm$1.02} & 60.86\tiny{$\pm$0.96} & 61.11\tiny{$\pm$1.47} & 41.30\tiny{$\pm$1.15} & 45.17\tiny{$\pm$1.50} & 30.93\tiny{$\pm$1.11} \\
 \texttt{6} Spatter + Motion Bias (S) &  72.48\tiny{$\pm$1.35} & 53.25\tiny{$\pm$1.08} & 18.85\tiny{$\pm$1.18} & 11.99\tiny{$\pm$0.79} & 10.11\tiny{$\pm$0.91} &  6.63\tiny{$\pm$0.62} \\
 \midrule
 \texttt{7} Speckle Noise + Motion Drift &  88.74\tiny{$\pm$0.95} & 63.57\tiny{$\pm$0.89} & 59.47\tiny{$\pm$1.48} & 38.75\tiny{$\pm$1.11} & 41.26\tiny{$\pm$1.49} & 27.50\tiny{$\pm$1.07} \\
 \texttt{8} Spatter + Motion Drift &  73.21\tiny{$\pm$1.34} & 54.16\tiny{$\pm$1.06} & 17.12\tiny{$\pm$1.14} & 10.50\tiny{$\pm$0.74} & 9.65\tiny{$\pm$0.89} &  6.04\tiny{$\pm$0.58} \\
 
 \midrule
  \textbf{\pnav-RGBD} & & & & & & \\
 \midrule
 \band \texttt{9} Clean&  99.55\tiny{$\pm$0.20} & 82.36\tiny{$\pm$0.41} & 99.45\tiny{$\pm$0.22} & 85.38\tiny{$\pm$0.47} & 98.72\tiny{$\pm$0.34} & 87.27\tiny{$\pm$0.40} \\
 \midrule
 \texttt{10} Low Lighting &  99.55\tiny{$\pm$0.20} & 82.25\tiny{$\pm$0.42} & 99.36\tiny{$\pm$0.24} & 86.15\tiny{$\pm$0.43} & 98.91\tiny{$\pm$0.31} & 86.73\tiny{$\pm$0.41} \\
 \texttt{11} Camera Crack &  99.27\tiny{$\pm$0.26} & 81.79\tiny{$\pm$0.43} & 97.18\tiny{$\pm$0.50} & 83.19\tiny{$\pm$0.59} & 91.53\tiny{$\pm$0.84} & 79.45\tiny{$\pm$0.80} \\
 \texttt{12} Spatter &  99.82\tiny{$\pm$0.13} & 82.40\tiny{$\pm$0.40} & 99.09\tiny{$\pm$0.29} & 84.69\tiny{$\pm$0.48} & 99.00\tiny{$\pm$0.30} & 86.96\tiny{$\pm$0.41} \\
 \midrule
 \texttt{13} Speckle Noise + Motion Bias (S) &  96.28\tiny{$\pm$0.57} & 75.59\tiny{$\pm$0.62} & 97.27\tiny{$\pm$0.49} & 80.77\tiny{$\pm$0.56} & 96.81\tiny{$\pm$0.53} & 82.11\tiny{$\pm$0.55} \\
 \texttt{14} Spatter + Motion Bias (S) &  96.46\tiny{$\pm$0.56} & 76.02\tiny{$\pm$0.62} & 94.99\tiny{$\pm$0.66} & 78.61\tiny{$\pm$0.67} & 96.36\tiny{$\pm$0.57} & 81.29\tiny{$\pm$0.59} \\
 \midrule
 \texttt{15} Speckle Noise + Motion Drift &  99.27\tiny{$\pm$0.26} & 77.85\tiny{$\pm$0.41} & 96.17\tiny{$\pm$0.58} & 76.77\tiny{$\pm$0.61} & 88.07\tiny{$\pm$0.98} & 71.39\tiny{$\pm$0.86} \\
 \texttt{16} Spatter + Motion Drift &  99.18\tiny{$\pm$0.27} & 77.24\tiny{$\pm$0.44} & 97.36\tiny{$\pm$0.48} & 78.42\tiny{$\pm$0.53} & 88.52\tiny{$\pm$0.96} & 71.87\tiny{$\pm$0.85}\\

\bottomrule
\end{tabulary}
}
\caption{\textbf{Breakdown of \pnav Performance Degradation by Episode Difficulty.} 
  Degradation in task
  performance of pre-trained \pnav RGB and RGB-D agents (trained for $\sim$75M frames) 
  for episodes of varying difficulties (based on shortest path lengths)
  when evaluated
  under \vcr\ and \dcr\ corruptions present in \rnav. For visual corruptions with controllable
  severity levels, 
  severity is set to 5 (worst).
  Reported results are mean and standard error across 3 evaluation runs under noisy actuations (wherever applicable). Success and SPL values
  are reported as percentages.
  }
  \label{tab:pnav_ep_diff_Res}
\end{table*}

\begin{table*}[ht!]
\footnotesize
\centering
\setlength{\tabcolsep}{7pt}
\resizebox{\linewidth}{!}{
\begin{tabulary}{\linewidth}{L  CC  CC  CC}
 \toprule
 \textbf{Increasing Episode Difficulty $\rightarrow$} & \multicolumn{2}{c}{\textbf{Easy}} & \multicolumn{2}{c}{\textbf{Medium}} & \multicolumn{2}{c}{\textbf{Hard}}\\
\textbf{Corruption $\downarrow$} 
  &\texttt{\textbf{SR}}$\uparrow$ & \texttt{\textbf{SPL}}$\uparrow$ &  \texttt{\textbf{SR}}$\uparrow$ & \texttt{\textbf{SPL}}$\uparrow$ &  \texttt{\textbf{SR}}$\uparrow$ & \texttt{\textbf{SPL}}$\uparrow$\\
  \midrule
  \textbf{\onav-RGB} & & & & & & \\
 \midrule
 \band \texttt{1} Clean&  40.50\tiny{$\pm$1.94} & 12.43\tiny{$\pm$1.04} & 33.48\tiny{$\pm$1.29} & 15.51\tiny{$\pm$0.73} & 25.75\tiny{$\pm$1.21} & 14.49\tiny{$\pm$0.75} \\
 \midrule
 \texttt{2} Low Lighting &  22.59\tiny{$\pm$1.65} &  8.50\tiny{$\pm$0.96} & 13.23\tiny{$\pm$0.93} &  5.60\tiny{$\pm$0.46} & 4.75\tiny{$\pm$0.59} &  2.40\tiny{$\pm$0.33} \\
 \texttt{3} Camera Crack &  21.65\tiny{$\pm$1.63} & 10.10\tiny{$\pm$1.05} & 5.38\tiny{$\pm$0.62} &  2.72\tiny{$\pm$0.34} & 1.61\tiny{$\pm$0.35} &  1.15\tiny{$\pm$0.26} \\
 \texttt{4} Spatter &  21.18\tiny{$\pm$1.61} &  6.39\tiny{$\pm$0.78} & 6.65\tiny{$\pm$0.68} &  2.66\tiny{$\pm$0.32} & 2.38\tiny{$\pm$0.42} &  0.95\tiny{$\pm$0.18} \\
 \midrule
 \texttt{5} Speckle Noise + Motion Bias (S) &  20.56\tiny{$\pm$1.60} &  7.01\tiny{$\pm$0.85} & 9.79\tiny{$\pm$0.81} &  4.89\tiny{$\pm$0.47} & 2.38\tiny{$\pm$0.42} &  1.23\tiny{$\pm$0.23} \\
 \texttt{6} Spatter + Motion Bias (S) &  20.56\tiny{$\pm$1.60} &  6.71\tiny{$\pm$0.83} & 7.32\tiny{$\pm$0.71} &  3.35\tiny{$\pm$0.38} & 1.61\tiny{$\pm$0.35} &  0.65\tiny{$\pm$0.15} \\
 \midrule
 \texttt{7} Speckle Noise + Motion Drift &  18.69\tiny{$\pm$2.67} &  8.55\tiny{$\pm$1.59} & 7.62\tiny{$\pm$1.26} &  3.86\tiny{$\pm$0.74} & 1.84\tiny{$\pm$0.64} &  0.97\tiny{$\pm$0.36} \\
 \texttt{8} Spatter + Motion Drift & 21.50\tiny{$\pm$1.99}  & 7.08\tiny{$\pm$1.01} & 6.84\tiny{$\pm$0.85} & 3.18\tiny{$\pm$0.45} & 0.57\tiny{$\pm$0.26} & 0.23\tiny{$\pm$0.11} \\
 
 \midrule
  \textbf{\onav-RGBD} & & & & & & \\
 \midrule
 \band \texttt{9} Clean&  46.73\tiny{$\pm$1.97} & 15.22\tiny{$\pm$1.15} & 35.72\tiny{$\pm$1.31} & 18.09\tiny{$\pm$0.80} & 29.58\tiny{$\pm$1.26} & 18.18\tiny{$\pm$0.86} \\
 \midrule
 \texttt{10} Low Lighting &  28.82\tiny{$\pm$1.79} & 10.55\tiny{$\pm$1.04} & 25.41\tiny{$\pm$1.19} & 11.56\tiny{$\pm$0.68} & 18.31\tiny{$\pm$1.07} &  9.68\tiny{$\pm$0.65} \\
 \texttt{11} Camera Crack &  35.51\tiny{$\pm$1.89} & 11.53\tiny{$\pm$1.03} & 28.03\tiny{$\pm$1.23} & 13.90\tiny{$\pm$0.73} & 22.45\tiny{$\pm$1.16} & 14.03\tiny{$\pm$0.79} \\
 \texttt{12} Spatter &  29.75\tiny{$\pm$1.81} &  9.79\tiny{$\pm$0.99} & 18.76\tiny{$\pm$1.07} &  9.06\tiny{$\pm$0.62} & 20.08\tiny{$\pm$1.11} & 11.00\tiny{$\pm$0.67} \\
 \midrule
 \texttt{13} Speckle Noise + Motion Bias (S) &  22.12\tiny{$\pm$1.64} &  5.93\tiny{$\pm$0.80} & 18.54\tiny{$\pm$1.06} &  8.29\tiny{$\pm$0.58} & 16.40\tiny{$\pm$1.03} &  7.43\tiny{$\pm$0.54} \\
 \texttt{14} Spatter + Motion Bias (S) &  27.26\tiny{$\pm$1.76} &  8.66\tiny{$\pm$0.92} & 19.81\tiny{$\pm$1.09} &  8.81\tiny{$\pm$0.60} & 18.93\tiny{$\pm$1.08} & 10.35\tiny{$\pm$0.66} \\
 \midrule
 \texttt{15} Speckle Noise + Motion Drift &  22.74\tiny{$\pm$1.66} &  6.35\tiny{$\pm$0.84} & 19.13\tiny{$\pm$1.08} &  7.86\tiny{$\pm$0.56} & 16.86\tiny{$\pm$1.04} &  8.56\tiny{$\pm$0.59} \\
 \texttt{16} Spatter + Motion Drift &  25.08\tiny{$\pm$1.71} &  8.16\tiny{$\pm$0.89} & 17.79\tiny{$\pm$1.05} &  8.16\tiny{$\pm$0.57} & 16.48\tiny{$\pm$1.03} &  8.69\tiny{$\pm$0.60} \\

\bottomrule
\end{tabulary}
}
\caption{\textbf{Breakdown of \onav Performance Degradation by Episode Difficulty.} 
  Degradation in task
  performance of pre-trained \onav RGB and RGB-D agents (trained for $\sim$300M frames) 
  for episodes of varying difficulties (based on shortest path lengths)
  when evaluated
  under \vcr\ and \dcr\ corruptions present in \rnav. For visual corruptions with controllable
  severity levels, 
  severity is set to 5 (worst).
  Reported results are mean and standard error across 3 evaluation runs under noisy actuations (wherever applicable). Success and SPL values
  are reported as percentages.
  }
  \label{tab:onav_ep_diff_Res}
\end{table*}

\par \noindent
\textbf{Performance Breakdown by Episode Difficulty.} In Tables.~\ref{tab:pnav_ep_diff_Res} and~\ref{tab:onav_ep_diff_Res} we break down performance
of \pnav and \onav agents by difficulty of evaluation episodes (based on shortest path lengths). We report results for a subset of \vcr, \dcr\ and \vdcr\ corruptions (mean
across $3$ evaluation runs under noisy actuations, wherever applicable).
For \pnav \texttt{RGB} agents, we find that while performance is comparable
across easy, medium and hard episodes under clean settings, under corruptions,
navigation performance decreases significantly 
with increase in episode difficulty --
indicating that under corruptions,
\pnav-\texttt{RGB} agents are more successful at reaching goal locations
closer to the spawn location. However, this is not the case
for \pnav \texttt{\texttt{RGB}-D} agents, 
where the drop in performance with increasing episode difficulty
is much less pronounced.
For \onav-\texttt{RGB} agents, we observe that performance (in terms of \texttt{SR} and \texttt{SPL}) drops as episodes become more difficult. For \onav-\texttt{\texttt{RGB}-D} agents,
although we find comparable \texttt{SPL} across episode difficulties in some
cases, the trends are mostly the same -- decreasing performance (in terms of \texttt{SR} and \texttt{SPL}) with increasing episode difficulty.

\end{document}